%% file: main.tex
\documentclass[10pt,twocolumn,letterpaper]{article}

\usepackage{cvpr}              % To produce the CAMERA-READY version
\input{preamble}
\usepackage{multirow}
\usepackage{afterpage} % 导言区添加此宏包
\definecolor{kellygreen}{rgb}{0.3, 0.73, 0.09}
\definecolor{alizarin}{rgb}{0.82, 0.1, 0.26}
\newcommand{\cmark}{{\color{kellygreen} \ding{51}}}
\newcommand{\xmark}{{\color{alizarin} \ding{55}}}
\definecolor{cvprblue}{rgb}{0.21,0.49,0.74}
\usepackage[pagebackref,breaklinks,colorlinks,allcolors=cvprblue]
{hyperref}
\usepackage{xcolor} % 放在导言区
\usepackage{pifont} % provide \ding command
\definecolor{kellygreen}{rgb}{0.3, 0.73, 0.09}
\definecolor{alizarin}{rgb}{0.82, 0.1, 0.26}

\definecolor{darkgreen}{RGB}{43, 195, 68} % 自定义一个暗绿色
\usepackage[accsupp]{axessibility}  % Improves PDF readability for those with disabilities.
\usepackage{colortbl}  % 用于给表格单元格上色
\definecolor{wkred}{RGB}{255, 190, 190}
\definecolor{wkblue}{RGB}{210, 230, 250}

\def\paperID{7944} % *** Enter the Paper ID here
\def\confName{CVPR}
\def\confYear{2025}
\newcommand{\myparagraph}[1]{\vspace{-1.0em}\paragraph{#1}}

\title{\textbf{\LARGE M}\textbf{V-\LARGE M}ATH: Evaluating Multimodal Math Reasoning in Multi-Visual Contexts}

\author{\textbf{Peijie Wang$^{1,2}$, Zhong-Zhi Li$^{1,2}$, Fei Yin$^{1,2}$, Xin Yang$^{1,2}$, Dekang Ran$^{1,2}$, Cheng-Lin Liu$^{1,2}$\thanks{Corresponding author}}\\
MAIS, Institute of Automation of Chinese Academy of Sciences$^{1}$\\
School of Artificial Intelligence, University of Chinese Academy of Sciences$^{2}$\\
\{wangpeijie2023, lizhongzhi2022, randekang2025\}@ia.ac.cn\\
\{fyin, liucl\}@nlpr.ia.ac.cn
}

\begin{document}
\maketitle
\input{sec/0_abstract}    
\input{sec/1_intro}
\input{sec/2_relatedwork}

\input{sec/3_MI-MATH}

\input{sec/4_EXPRIMENTS}

\input{sec/5_Analysis}
\input{sec/6_CONCLUSION}

\input{sec/7_Acknowledgement}
\clearpage
{
    \small
    \bibliographystyle{ieeenat_fullname}
    \bibliography{main}
}

\input{appendix/X_suppl}

\end{document}

%% file: preamble.tex
\usepackage[dvipsnames]{xcolor}

\usepackage{marvosym}  % Package for symbols
\usepackage{graphicx}
\usepackage{caption}
\usepackage{stfloats} % add this to your preamble
\usepackage{cuted}
\usepackage{multirow}
\usepackage{tabularx}

\usepackage{tocloft}
\usepackage{etoc}

\usepackage{titletoc}

\newcommand\DoToC{%
  \startcontents
  \printcontents{}{1}{\noindent \textbf{\Large{Table of Contents in Appendix}}\vskip3pt\vskip5pt}
  \vskip3pt\vskip5pt
}

\newcommand{\listappendixname}{List of Appendices}
\newlistof{appendices}{apc}{\listappendixname}
\newlistof{appfigures}{afg}{List of Case Study Figures}

\usepackage[normalem]{ulem}
\useunder{\uline}{\ul}{}

\extrafloats{100}

\usepackage{multicol}
\usepackage{aliascnt}
\usepackage{adjustbox}

%% file: sec/0_abstract.tex
\begin{abstract}
Multimodal Large Language Models (MLLMs) have shown promising capabilities in mathematical reasoning within visual contexts across various datasets. However, most existing multimodal math benchmarks are limited to single-visual contexts, which diverges from the multi-visual scenarios commonly encountered in real-world mathematical applications. To address this gap, we introduce MV-MATH: a meticulously curated dataset of 2,009 high-quality mathematical problems. Each problem integrates multiple images interleaved with text, derived from authentic K-12 scenarios, and enriched with detailed annotations. MV-MATH includes multiple-choice, free-form, and multi-step questions, covering 11 subject areas across 3 difficulty levels, and serves as a comprehensive and rigorous benchmark for assessing MLLMs’ mathematical reasoning in multi-visual contexts. Through extensive experimentation, we observe that MLLMs encounter substantial challenges in multi-visual math tasks, with a considerable performance gap relative to human capabilities on MV-MATH. Furthermore, we analyze the performance and error patterns of various models, providing insights into MLLMs' mathematical reasoning capabilities within multi-visual settings. The data and code: \url{https://eternal8080.github.io/MV-MATH.github.io/}.
\end{abstract}

%% file: sec/1_intro.tex
\section{Introduction}
\label{sec:intro}

\begin{figure*}[!t]
    \centering
    \includegraphics[width=0.9\linewidth]{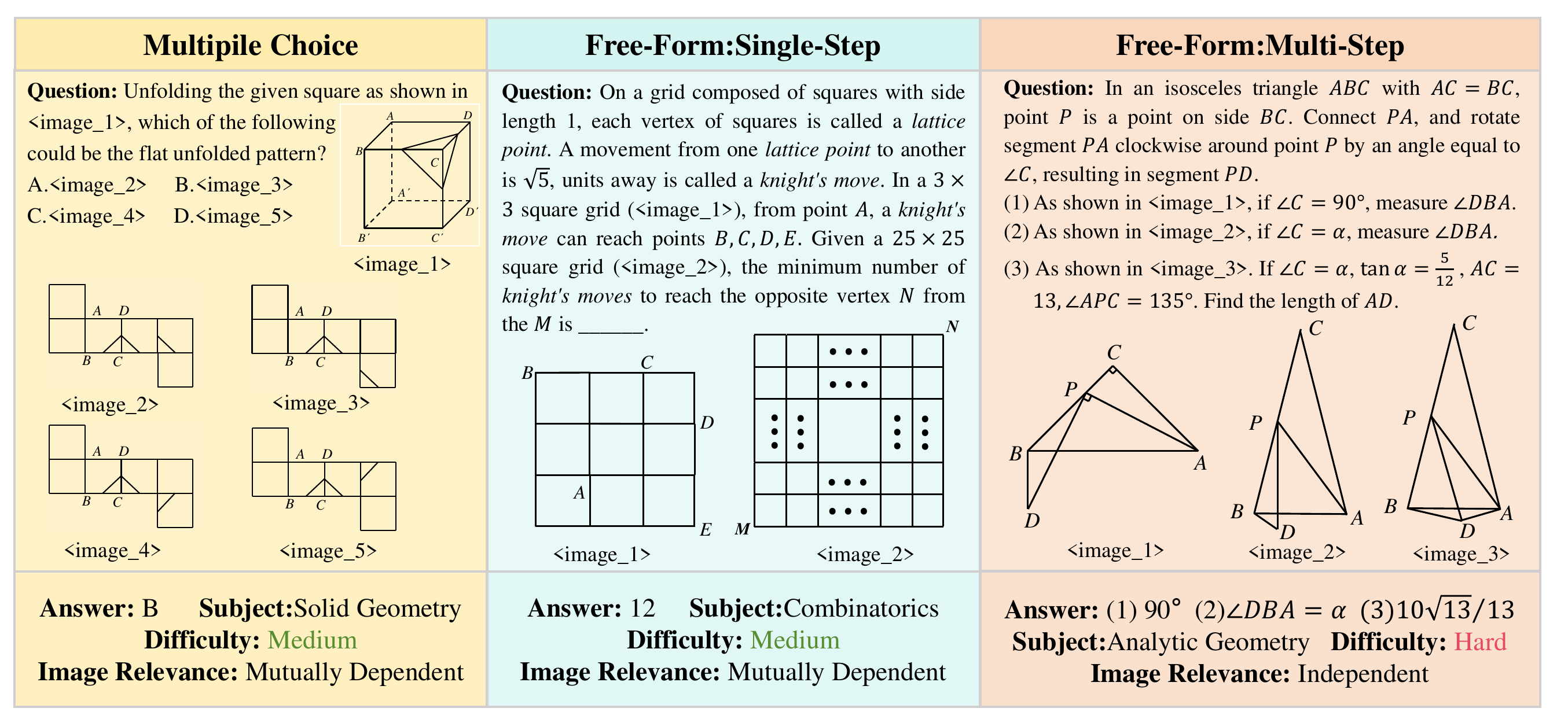}
    \vspace{-3mm}
    \caption{Sampled MV-MATH examples from each question type. Each sample contains a multi-visual contexts. The specifics of image relevance are discussed in detail in~\cref{subsec: dataset_overview}}
    \label{fig: dataset_examples}
    \vspace{-5mm}
\end{figure*}
{\small \textit{"The more you see, the more you know." — Leonardo da Vinci}}\vspace{0.3em}
Large Language Models (LLMs) have demonstrated exceptional performance across a wide range of tasks including natural language understanding, question answering and code generation~\cite{llms1,llms2,llms3, Geoqa,qa1,xu2025prompting}. Building on LLMs, MLLMs excel at various multimodal tasks by learning specialized features from different modalities, demonstrating a strong ability to understand different modalities, and have achieved remarkable results in various visual-language tasks such as image captioning, visual question answering, and visual reasoning~\cite{mllms1,mllms2,mllms3,mllms4,mllms5,wang2023large}. 

As MLLMs advance in various tasks, their mathematical reasoning abilities have garnered significant attention. Progressing from early text-based benchmarks like GSM8K~\cite{GSM8K} and MATH~\cite{MATH} to more recent multimodal benchmarks that incorporate visual elements, such as MathVista~\cite{mathvista}, MMMU~\cite{MMMU}, and GeoEval~\cite{geoeval}, MLLMs have demonstrated exceptional proficiency in mathematical reasoning, with the best-performing model on MathVista even surpassing human performance~\cite{qa2}. Most existing multimodal mathematics evaluation datasets, such as MathVista, MathVision~\cite{mathvision}, and MathVerse~\cite{mathverse}, are limited to single-visual contexts, requiring models to reason based on only a single image input. This raises an important question: are single-visual scenarios sufficient to fully capture the reasoning capabilities of MLLMs? Evidently, they are not.

Several multi-visual datasets in general domains have been introduced ~\cite{blink,muirbench,demon,mmtbench,Mementos}, which provide valuable resources. However, research specifically focused on mathematical evaluation within multi-visual scenarios remains limited. MathVerse-mv~\cite{llava-next-interleave} extends the original MathVerse dataset by adapting problems and manually adding supplementary images to transform MathVerse into a multi-visual dataset. However, this approach to data construction limits the diversity of visual information and introduces potential biases. Additionally, MathVerse-mv is relatively small, containing only 788 samples. Another dataset, CMM-Math~\cite{cmm-math}, focuses on Chinese contexts and includes 765 multi-image samples in its test set, but some images are of suboptimal quality, affecting the clarity of certain samples. Both MathVerse-mv and CMM-Math lack fine-grained categorizations and a diverse range of question types.

To overcome the limitations of existing multi-image mathematical reasoning datasets, which are constrained in both quantity and diversity, and to enable a more comprehensive evaluation of MLLMs in multi-visual contexts, we present a novel dataset: MV-MATH. This dataset consists of 2,009 high-quality mathematics problems, each presented in interleaved multi-visual contexts with more than two images, covering a variety of grade 12 scenarios. Every problem has been meticulously curated from authentic K-12 sources. To ensure the quality and reliability of MV-MATH, each sample has undergone cross-validation by at least two annotators, ensuring the accuracy and precision of the problems, answers, and images. The dataset includes 1,109 multiple-choice questions and 900 free-form questions, of which 100 are multi-step problems, adding complexity and presenting a greater challenge for the models. The problems are categorized into three difficulty levels and span 11 distinct mathematical subjects, each annotated with image relevance, allowing for finer-grained analysis.

Extensive experiments are conducted on MV-MATH to comprehensively evaluate model performance in multi-visual mathematical contexts. On our MV-MATH dataset, Claude~\cite{claude3} achieves the highest performance with a score of 33.9\%. Notably, the open-source model LLaVA-OneVision~\cite{llava-onevision} achieves a competitive performance with a score of 26.2\%, surpassing GPT-4v~\cite{GPT-4V} and ranking just below Qwen-vl-max~\cite{qwenvl}. Nevertheless, the performance of all these models remains well below human-level capability. Through analysis of these results, we offer insights into the strengths and limitations of open-source models. In conclusion, the contributions of our study are as follows:
\begin{itemize}
    \item We propose the MV-MATH benchmark, comprising 2,009 interleaved multi-visual mathematics problems derived from real K-12 scenarios, covering multiple-choice, free-form, and multi-step formats. All problems are classified across 3 difficulty levels and annotated with 11 fine-grained subjects.

    \item We evaluate 25 MLLMs on MV-MATH, revealing a substantial gap between current model performance and human-level capabilities in multi-visual mathematical tasks. Through an in-depth study of LLaVA-OneVision, we provide insights for enhancing model capabilities in multi-image mathematical reasoning.

    \item Our further investigation on MV-MATH reveals that MLLMs perform less effectively on image-dependent tasks compared to image-independent ones. Additionally, sequential image input outperforms the merged approach, and chain-of-thought (CoT) prompting does not consistently yield improvements.
    
    \item With the fine-grained categorization, we perform a comprehensive error analysis of current MLLMs, offering in-depth insights to inform future advancements.
\end{itemize}

\begin{figure*}[!t]
    \centering
    \includegraphics[width=0.9\linewidth]{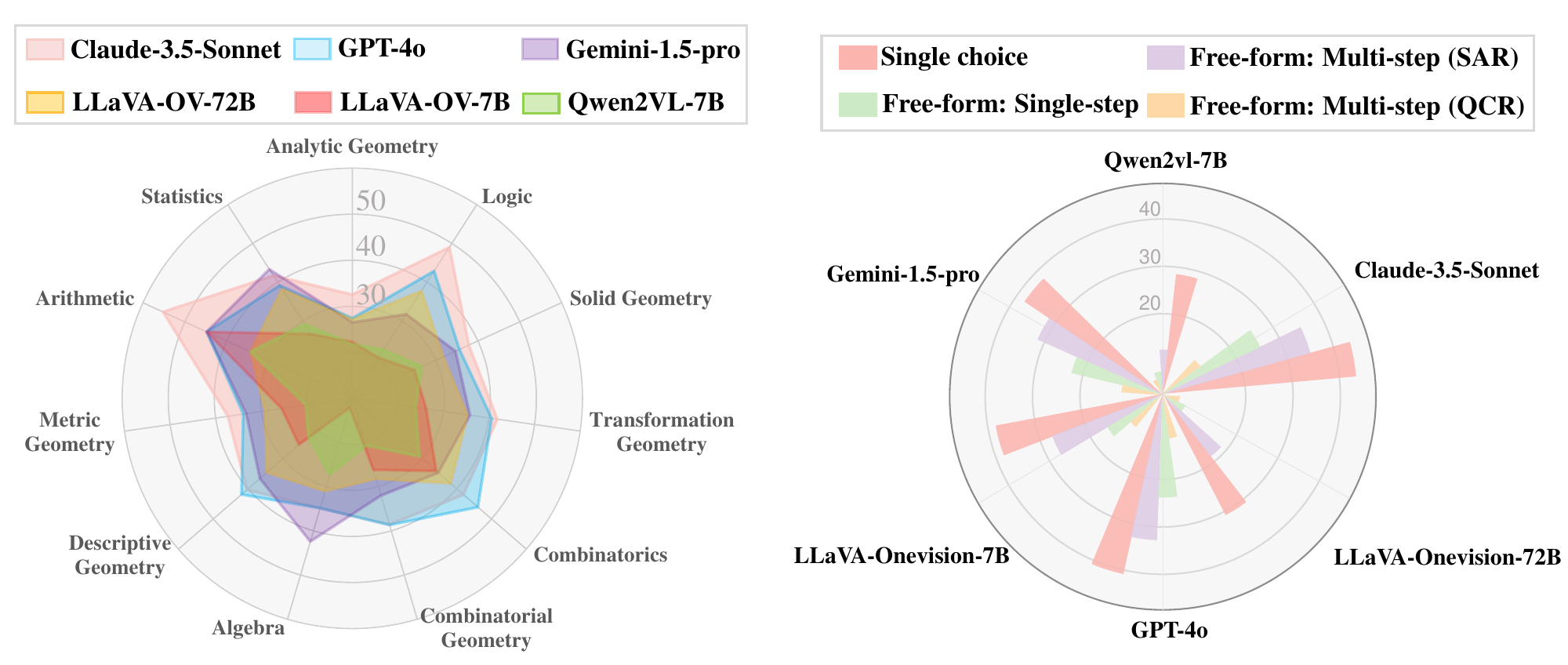}
    \vspace{-3mm}
    \caption{Performance comparison of six Multimodal Large Language Models(MLLMs) on our proposed MV-MATH dataset across 11 subjects(left) and 3 question types(right). SAR: Step Accuracy Rate, QCR: Question Completeness Rate.}
    \label{fig: lidar}
    \vspace{-5mm}
\end{figure*}

%% file: sec/2_relatedwork.tex
\section{Related Work}
\label{sec:formatting}
%-------------------------------------------------------------------------
\subsection{Benchmark for Mathematical Reasoning}

\paragraph{Pure Text Benchmarks} 
MathQA~\cite{MATHQA} is a large-scale benchmark focused on math word problems, designed to evaluate models' capabilities in solving complex arithmetic and algebraic tasks through natural language. GSM8K~\cite{GSM8K}, developed by OpenAI, assesses LLMs' mathematical reasoning with 8,500 grade-school-level problems requiring 2 to 8 steps, covering arithmetic, fractions, and basic algebra. MATH~\cite{MATH} offers a more challenging dataset of 12,500 high school competition problems, surpassing GSM8K in difficulty. MathBench~\cite{MathBench} includes 3,709 problems, ranging from basic arithmetic to college-level questions and spanning multiple difficulty levels. SuperCLUE-Math~\cite{SuperCLUE-Math}, a Chinese benchmark, evaluates multi-step reasoning with over 2,000 problems, each requiring multiple steps and accompanied by natural language solutions.
\vspace{-5mm}
\paragraph{Multimodal Benchmarks}
With the advancement of MLLMs, several high-quality benchmarks have been developed to evaluate mathematical problem-solving abilities in visual contexts. MathVista~\cite{mathvista} is the first multimodal dataset aimed at assessing MLLMs’ mathematical reasoning, specifically focusing on visual math QA tasks. MATH-Vision~\cite{mathvision} targets multimodal mathematical understanding, featuring questions primarily sourced from mathematics competitions and requiring deep reasoning. MathVerse~\cite{mathverse} tests whether current MLLMs can effectively leverage diagram information in mathematical reasoning tasks, offering a collection of 2,612 visual diagrams. GeoEval~\cite{geoeval} focuses on geometry, examining the capabilities of large models in this field through 2,000 geometric samples. CMMath~\cite{cmmath} primarily evaluates mathematical reasoning abilities in the context of Chinese K-12 education.

All of the above datasets are restricted to single-visual contexts, relying on a single image input per problem. In contrast, our MV-MATH dataset emphasizes multi-visual mathematical reasoning by incorporating multiple images per problem. Addressing multi-image problems requires models to develop a more comprehensive understanding of relationships between mathematical visual elements, aligning more closely with real-world mathematical scenarios.

\subsection{Multimodal Foundation Models}
The progress in visual-language alignment and rapid evolution of LLMs have endowed these models with visual capabilities, with notable achievements including CLIP~\cite{CLIP} and BLIP~\cite{BLIP}. CLIP aligns image-text pairs via contrastive learning, while BLIP combines contrastive and generative learning to align and generate image-text pairs. Inspired by these, MiniGPT-4~\cite{minigpt4}, LLaMA-Adapter~\cite{llama-adapter}, and LLaVA~\cite{llava} extend text-only models into multimodal ones through vision-language alignment and instruction tuning. Subsequently, large multimodal models emerged, such as the closed-source GPT-4V~\cite{GPT-4V}, Gemini~\cite{gemini}, Claude~\cite{claude3}, GPT-4o~\cite{GPT4o} and open-source DeepSeek-VL~\cite{deepseekvl}, Qwen-VL~\cite{qwenvl}, InternLMXComposer-VL~\cite{InternLMXComposer-VL}, LLaVA-NeXT~\cite{llavanext}, and LLaVA-OneVision~\cite{llava-onevision}, which perform well on MMMU~\cite{MMMU} and MathVista~\cite{mathvista}. In mathematics, G-LLaVA~\cite{G-LLaVA} and Math-LLaVA~\cite{math-llava} enhance performance on mathematical tasks using large instruction datasets Geo170K~\cite{G-LLaVA} and MathV360K~\cite{math-llava}. MAVIS~\cite{MAVIS} improves visual perception with three-stage fine-tuning, while MultiMath~\cite{multimath} refines visual-language alignment through a four-stage training process.

%% file: sec/3_MI-MATH.tex
\section{The \textbf{M}V-\textbf{M}ATH Benchmark}
\label{sec:dataset}
\subsection{Overview}
\label{subsec: dataset_overview}
We introduce the MV-MATH benchmark, a meticulously annotated dataset designed to evaluate the mathematical reasoning capabilities of MLLMs in multi-visual contexts. Each sample in MV-MATH consists of interleaved multi-image and text, closely reflecting the multimodal distribution found in real-world scenarios. This interleaved format imposes greater demands on modal fusion and comprehension in MLLMs, providing a more realistic assessment of their performance~\cite{mantis,muirbench}

The MV-MATH dataset comprises 2,009 multi-image questions, with some questions containing up to 8 images. It includes three types: multiple-choice, free-form and multi-step questions. MV-MATH is organized into 11 subjects over 3 difficulty levels, including \textit{Analytic Geometry, Algebra, Metric Geometry, Combinatorics, Transformation Geometry, Logic, Solid Geometry, Arithmetic, Combinatorial Geometry, Descriptive Geometry and Statistics}, covering a range of scenarios from the K-12 mathematics curriculum. Based on image relevance, we categorize MV-MATH into two subsets: a mutually dependent set (\textbf{MD}), where images are interrelated and understanding one image necessitates information from another (left and middle of~\Cref{fig: dataset_examples}); and an independent set (\textbf{ID}), where images are unrelated and can be interpreted independently without reference to other images (right of~\Cref{fig: dataset_examples}). To facilitate analysis, we provide a test-mini version of MV-MATH containing 200 samples, preserving similar distribution of difficulty levels and question types as the full MV-MATH dataset. Detailed statistics and coverage of MV-MATH are presented in~\Cref{tab:dataset_statistics}.

\begin{table}[!t]
\centering
\small
    \begin{tabular}{lc}
        \toprule
         \textbf{Statistics} & \multicolumn{1}{c}{\textbf{Number}} \\ \midrule
        Total Questions                  & 2009                      \\
         \hspace{1em}*multiple-choice questions      & 1109                      \\
         \hspace{1em}*Free-form questions            & 900                       \\ 
         \hspace{1.5em}-one-step questions            & 800                       \\
         \hspace{1.5em}-multi-step questions          & 100                       \\  
         Questions in the testmini set & 200    \\   \midrule
         
        Difficulties (Easy: Medium: Hard)  & 27\%:48\%:25\%   \\
       \midrule
        Total images                 & 6061              \\
        \hspace{1em}* question with 2 images    & 979 (48.73\%)        \\
        \hspace{1em}* question with 3 images    & 312 (15.53\%)          \\
        \hspace{1em}* question with 4 images    & 453 (22.55\%)        \\
        \hspace{1em}* question with 5 images and above & 265 (13.39\%)        \\
        \midrule
         Image Relevance & \\
         \hspace{1em}* Mutually Dependent questions & 1412 \\
         \hspace{1em}* Independent questions & 597 \\
         \midrule
        Minimum question length          & 14                     \\
        Maximum question length          & 383                      \\
        Average question length          & 80.17                    \\ 

        \bottomrule
    \end{tabular}
    \vspace{-2mm}
    \caption{Key statistics of \textbf{M}V-\textbf{M}ATH.}
    \vspace{-5mm}
    \label{tab:dataset_statistics}
\end{table}

\subsection{Data Construction}
\noindent\textbf{Data collection.} 
We initially collect over 300,000 mathematics problems spanning grade 12 scenarios, comprising multiple-choice and free-form questions, including plain text, single-image, and multi-image questions, all stored in PDF format. Using the Mathpix API\footnote{\url{https://mathpix.com/convert}}, we extract text content—such as questions, answers, and analyses—as well as images, and organized this data in JSON format. Based on image counts, we identified a preliminary set of 49,538 multi-image problems.

\noindent\textbf{Data filtering.}
To ensure the high quality of the multi-image mathematical data, we employ a three-stage data screening strategy. In the first stage, we verify the alignment between text and images within each question, this is a crucial step since Mathpix does not always yield the correct number of images. This initial filtering retains 35,562 samples. In the second stage, we inspect the remaining samples for missing text fields or semantic inaccuracies, then categorize them into two subsets: 10,110 multiple-choice and 8,933 free-form questions for further screening. In the third stage, we manually filter out samples with low-quality images (e.g., blurry images or images containing text content). This process ultimately yields 1,109 multiple-choice questions, 800 one-step free-form questions, and 100 multi-step free-form questions. Each step was cross-verified by at least two graduate students to ensure the reliability of the screening outcomes.

\noindent\textbf{Data labeling.}
Using the above samples, we conduct difficulty grading, subject categorization, and image relevance classification. For each problem, difficulty is categorized by analyzing the lengths of the question and analysis, assigning weights of $0.4$ and $0.6$ respectively. Based on the weighted lengths, MV-MATH is divided into three levels: easy, medium, and hard. For subject categorization and image relevance, we apply a majority voting approach, leveraging classification results from GPT-4o, Claude-3.5-Sonnet, and Qwen-VL-Max to determine each problem's subject and its degree of image dependence. We then manually validate the assigned difficulty levels, subjects, and image relevance, with two annotators independently reviewing each problem and making adjustments to correct misclassifications, ensuring the accuracy of the final result.

\begin{figure}[!t]
    \centering
    \includegraphics[width=1.0\linewidth]{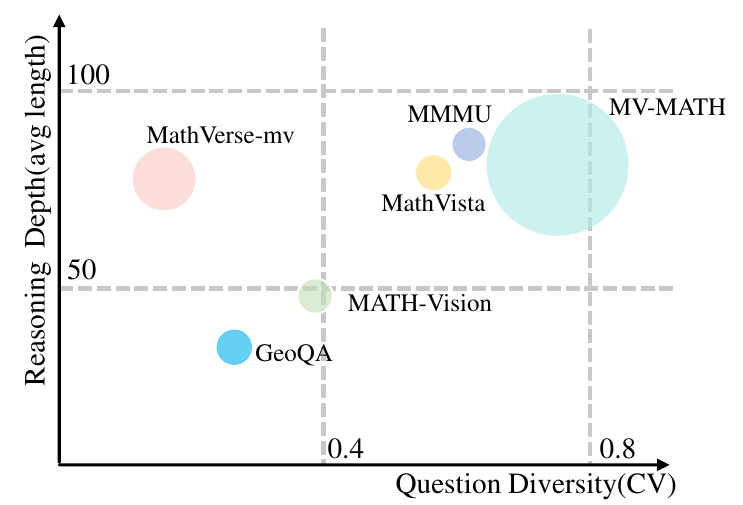}
    \vspace{-7mm}
    \caption{Comparison of MV-MATH with other existing benchmarks. The size of each circle represents the image distribution per question. The definition and calculation method of CV are provided in~\Cref{subsec: cv}}
    \vspace{-5mm}
\label{fig:comparison}
\end{figure}

\subsection{Comparison with Existing Benchmarks}
\label{subsec: cv}
Most existing multimodal mathematical reasoning datasets are limited to single-visual contexts, where the model interprets only a single image without the need for information interaction between images. To further distinguish the difference between MV-MATH and other existing ones, we elaborate the benchmark details in~\Cref{fig:comparison}. Here, we primarily compare our MV-MATH with MathVerse-mv~\cite{llava-next-interleave} and CMM-MATH~\cite{mathvista}.
\myparagraph{Comparing with MathVerse-mv}
MathVerse-mv~\cite{llava-next-interleave} is derived by reformatting original problems in MathVerse through question rephrasing and image combination, resulting in 788 generated problems. However, all questions in MathVerse-mv are multiple-choice and primarily focus on a single task type: \textit{'select the correct paired math diagram from given images.'} This dataset lacks differentiation in difficulty levels and subject categories. Such artificial rephrasing and image augmentation introduce distribution bias, limiting its representation of authentic multi-image mathematical scenarios. In comparison, our MV-MATH dataset provides 2,009 high-quality samples from real K-12 contexts, including both multiple-choice and free-form questions, each classified by fine-grained difficulty and subject. Sample lengths in MathVerse-mv range from 62 to 246, with an average of 81.91, whereas MV-MATH samples range from 14 to 383, with an average of 80.17. To assess the diversity in sample length distribution, we calculate the coefficient of variation (CV), with a higher CV indicating greater distribution richness. The CVs for MathVerse-mv and MV-MATH are 0.19 and 0.74 respectively, highlighting the broader distribution diversity of MV-MATH. The CV calculation is as follows:
\begin{equation}
\text{CV} = \frac{\sigma}{\mu} \times 100\%
\end{equation}
where \(\sigma\) denotes the standard deviation of sample lengths, and \(\mu\) denotes the mean sample length.

\begin{table}[t]
\small
    \centering
    \resizebox{\linewidth}{!}{%
\begin{tabular}{@{}lcccc@{}}
\toprule
\multirow{2}{*}{Models} & Easy & Medium & Hard & Overall \\
 & (546) & (968) & (495) & (2009) \\ \midrule
Claude-3.5-sonnet~\cite{claude3} & 35.7 & 37.5 & 26.6 & 33.9 \\
GPT-4o~\cite{GPT4o} & 40.3 & 32.7 & 22.9 & 32.1\\
Gemini-1.5-pro~\cite{gemini} & 34.1 & 30.0 & 20.3 & 29.0 \\
Qwen-vl-max~\cite{qwenvl} & 32.7 & 27.7 & 20.1 & 26.9 \\
GPT-4V~\cite{GPT-4V} & 31.6 & 24.7 & 17.1 & 24.5 \\  \midrule
LLaVA-OneVision-Chat-72B~\cite{llava-onevision} & 34.6 & 26.0 & 19.2 &26.2 \\
LLaVA-OneVision-Chat-7B~\cite{llava-onevision} & 20.6 & 21.2 & 14.8 &19.0\\
Qwen2VL-Instruct-7B~\cite{Qwen2VL} & 18.8 & 17.1 & 13.9 & 16.5 \\
LLaVA-NeXT-Interleave-7B~\cite{llava-next-interleave} & 15.3 & 16.1 & 10.9 &14.7  \\
Deepseek-VL-Chat-7B~\cite{deepseekvl} & 17.1 & 14.1 & 12.5 & 14.5 \\
\bottomrule
\end{tabular}
}
    \vspace{-3mm}
    \caption{Result decomposition across question difficulty levels.}
    \label{tab:result_diffculty}
    \vspace{-5mm}
\end{table}

\myparagraph{Comparing with CMM-Math}
CMM-Math~\cite{cmm-math} is a dataset designed for Chinese-language scenarios, with the multi-image portion containing 765 samples. Some images in CMM-Math are of lower quality, due to Mathpix’s filtering inconsistencies, some images contain excessive text. In contrast, we performed manual checks on each image in our samples to ensure that every sample in MV-MATH is of reliable quality. Additionally, MV-MATH primarily focuses on English-language scenarios. To investigate the impact of language on multi-image mathematical reasoning performance, we also provide a translated Chinese version of \textbf{M}V-\textbf{M}ATH.

%% file: sec/4_EXPRIMENTS.tex
\section{Experiments}
\label{sec:expriment}

\begin{table}[!t]
\small
    \centering
    \resizebox{\linewidth}{!}{%
\begin{tabular}{@{}lccc@{}}
\toprule
\textbf{{Models}} & \textbf{{Original}} & \textbf{{CoT}} & \textbf{{CoT \&2-shot}}  \\  \midrule
Claude-3.5-sonnet~\cite{claude3} & 29.2 & 32.6\textcolor{red}{\scalebox{0.8}{(+3.4)}} & \textbf{33.9}\textcolor{red}{\scalebox{0.8}{(+1.3)}}\\
GPT-4o~\cite{GPT4o} & 31.8 & 30.9\textcolor{darkgreen}{\scalebox{0.8}{(-0.9)}}
 & \textbf{32.1}\textcolor{red}{\scalebox{0.8}{(+1.2)}}
  \\
Gemini-1.5-pro~\cite{gemini} & \textbf{29.8} & 28.3\textcolor{darkgreen}{\scalebox{0.8}{(-1.5)}} & 29.1\textcolor{red}{\scalebox{0.8}{(+0.8)}}  \\
Qwen-vl-max~\cite{qwenvl} & 28.4 & \textbf{29.1}\textcolor{red}{\scalebox{0.8}{(+0.7)}} & 26.9\textcolor{darkgreen}{\scalebox{0.8}{(-2.2)}} \\
GPT-4V~\cite{GPT-4V} & 23.4 & \textbf{25.1}\textcolor{red}{\scalebox{0.8}{(+1.7)}} & 24.5\textcolor{darkgreen}{\scalebox{0.8}{(-0.6)}}  \\  \midrule
LLaVA-OneVision-Chat-72B~\cite{llava-onevision} & \textbf{27.3} & 26.7\textcolor{darkgreen}{\scalebox{0.8}{(-0.6)}} & 26.2\textcolor{darkgreen}{\scalebox{0.8}{(-0.5)}}  \\
LLaVA-OneVision-Chat-7B~\cite{llava-onevision} & \textbf{20.5} & 20.1\textcolor{darkgreen}{\scalebox{0.8}{(-0.4)}} & 19.1\textcolor{darkgreen}{\scalebox{0.8}{(-1.0)}}  \\
LLaVA-NeXT-Interleave-7B~\cite{llava-next-interleave} & \textbf{17.9} & 16.3\textcolor{darkgreen}{\scalebox{0.8}{(-1.6)}} & 14.7\textcolor{darkgreen}{\scalebox{0.8}{(-1.6)}}  \\
Qwen2VL-Instruct-7B~\cite{Qwen2VL} & \textbf{19.7} & 16.7\textcolor{darkgreen}{\scalebox{0.8}{(-2.0)}} & 16.5\textcolor{darkgreen}{\scalebox{0.8}{(-0.2)}}  \\
Deepseek-VL-Chat-7B~\cite{deepseekvl} & \textbf{17.2} & 15.7\textcolor{darkgreen}{\scalebox{0.8}{(-1.5)}} & 14.5\textcolor{darkgreen}{\scalebox{0.8}{(-1.2)}}  \\
\bottomrule

\end{tabular}
}
    \vspace{-3mm}
    \caption{Model Performance Evaluation for Original, CoT, and CoT with 2-shot configurations.}
    \label{tab:CoT&shot}
    \vspace{-5mm}
\end{table}

\begin{table*}[htbp]
\centering
\resizebox{\textwidth}{!}
{%
\begin{tabular}{l|c|ccccccccccc}
\toprule
Model & Overall & AG  & Algebra & MG & Combinatorics & TG & Logic  & SG & Arithmetic & CG & DG & Statistics  \\
\midrule

\multicolumn{13}{c}{LLMs(Text-only, CoT with 2-shot)}\\
\midrule
Deepseek-Chat\cite{deepseekchat} & 16.1 & 14.7 & 10.1 & 16.3 & 18.7 & 25.3 & 21.8 & 18.3 & 6.2 & 19.0 & 8.2 & 13.2 \\
\midrule

\multicolumn{13}{c}{LLMs(Text + Image Caption, CoT with 2-shot)}\\
\midrule
Deepseek-Chat\cite{deepseekchat} & 16.9 & 14.1 & 8.0 & 15.3 & 13.4 & 13.2 & 12.7 & 9.4 & 14.4 & 12.8 & 18.0 & 17.3 \\
\midrule

\multicolumn{13}{c}{Open-source MLLMs (Text + Image, CoT with 2-shot)}\\
\midrule
Math-LLaVA-13B\cite{math-llava} & 3.0 & 1.6 & 6.9 & 4.7 & 4.8 & 2.9 & 0.0 & 3.2 & 18.7 & 6.6 & 2.1 & 5.8 \\
LLaVA-v1.5-13B\cite{llava} & 5.0 & 4.8 & 6.8 & 4.1 & 4.8 & 8.7 & 9.0 & 3.5 & 12.5 & 5.1 & 5.0 & 11.7 \\
LLaVA-v1.5-7B\cite{llava} & 10.3 & 9.3 & 11.7 & 11.2 & 9.7 & 12.8 & 13.6 & 10.2 & 0.0 & 7.7 & 23.7 & 11.7 \\
VILA-13B\cite{VILA} & 12.0 & 11.5 & 11.0 & 11.0 & 12.1 & 14.4 & 18.1 & 13.2 & 37.5 & 10.6 & 20.8 & 5.8 \\
InternLM-XComposer2.5-VL-7B\cite{InternLMXComposer-VL} & 13.1 & 12.2 & 12.6 & 13.2 & 24.3 & 20.6 & 36.3 & 9.4 & 18.7 & 11.1 & 23.7 & 17.6 \\
InternVL-Chat-8B\cite{internvl} & 14.4 & 14.1 & 20.4 & 17.5 & 19.5 & 19.6 & 27.2 & 13.0 & 31.2 & 9.9 & 20.1 & 23.5 \\
Llama-3.2-Vision-Instruct-11B\cite{llama} & 14.4 & 15.0 & 15.4 & 16.2 & 23.1 & 15.6 & 18.1 & 11.9 & 31.2 & 14.3 & 25.1 & 17.6 \\
Deepseek-VL-7B\cite{deepseekvl} & 14.5 & 14.8 & 20.2 & 10.8 & 17.0 & 19.8 & 9.0 & 15.1 & 18.7 & 10.9 & 26.6 & 29.4 \\
LLaVA-NeXT-Interleave-7B\cite{llava-next-interleave} & 14.7 & 14.0 & 15.5 & 15.2 & 17.0 & 18.2 & 18.1 & 16.3 & 6.2 & 14.1 & 24.4 & 23.5 \\
Mantis-Idefics2-8B\cite{mantis} & 15.5 & 13.3 & 17.7 & 19.2 & 14.6 & 20.9 & 22.7 & 12.3 & 6.2 & 16.0 & 26.6 & 11.7 \\
Mantis-siglip-8B\cite{mantis} & 15.8 & 17.9 & 17.7 & 17.9 & 14.6 & 20.4 & 22.7 & 12.1 & 18.7 & 10.8 & 32.3 & 17.6 \\
Qwen2VL-Instruct-7B\cite{Qwen2VL} & 16.5 & 14.2 & 18.6 & 14.8 & 17.0 & 21.9 & 22.7 & 17.2 & 31.2 & 16.1 & 25.1 & 23.5 \\
LLaVA-OneVision-SI-7B\cite{llava-onevision} & 17.2 & 16.1 & 19.5 & 13.2 & 16.0 & 19.5 & 12.6 & 15.0 & 36.5 & 13.2 & 31.3 & 13.6 \\
LLaVA-OneVision-SFT-7B\cite{llava-onevision} & 18.8 & 18.2 & 20.3 & 22.3 & 17.3 & 20.1 & 9.0 & 15.8 & 43.1 & 15.8 & 27.3 & 23.5 \\
LLaVA-OneVision-Chat-7B\cite{llava-onevision} & 19.1 & 19.6 & 20.4 & 21.4 & 14.6 & 18.8 & 4.5 & 20.4 & 43.7 & 16.7 & 28.9 & 29.4 \\
LLaVA-OneVision-SI-72B\cite{llava-onevision} & 25.0 & 24.7 & 24.3 & 27.6 & 27.0 & 25.3 & \colorbox{wkblue}{37.9} & 24.4 & 37.1 & 20.4 & 31.2 & 23.5 \\
LLaVA-OneVision-SFT-72B\cite{llava-onevision} & 25.9 & 24.2 & 31.3 & 21.1 & 23.1 & 28.9 & 31.8 & 32.8 & 18.7 & 21.5 & 39.5 & 29.4 \\
LLaVA-OneVision-Chat-72B\cite{llava-onevision} & 26.2 & 25.1 & 32.4 & 23.9 & 35.3 & 28.1 & 27.2 & 31.6 & 31.2 & 22.6 & 35.9 & 35.2 \\
\midrule

\multicolumn{13}{c}{Closed-source MLLMs (Text + Image, CoT with 2-shot)} \\
\midrule
Qwen-vl-plus\cite{qwenvl} & 19.7 & 17.9 & 24.1 & 22.0 & 16.0 & 19.9 & 24.8 & 15.9 & 15.2 & 18.7 & 31.4 & 29.4  \\
GPT-4V\cite{GPT-4V} & 24.5 & 18.7 & 31.6 & 32.4 & 25.6 & 26.3 & 36.3 & 26.8 & \colorbox{wkblue}{43.7} & 19.3 & 33.8 & 35.2 \\
Qwen-vl-max\cite{qwenvl} & 26.9 & 27.6 & 32.1 & 24.7 & 36.5 & 29.6 & 31.8 & 30.9 & 37.5 & \colorbox{wkblue}{23.7} & 32.3 & 23.5 \\
Gemini-1.5-Pro\cite{gemini} & 29.1 & \colorbox{wkblue}{29.9} & 32.9 & 28.3 & 28.0 & 30.5 & \colorbox{wkred}{40.5} & 33.9 & 42.7 & 21.7 & 30.6 & 35.2 \\
GPT-4o\cite{GPT4o} & \colorbox{wkblue}{32.1} & 28.7 & \colorbox{wkblue}{36.7} & \colorbox{wkred}{34.4} & \colorbox{wkblue}{39.4} & \colorbox{wkblue}{30.6} & 29.8 & \colorbox{wkred}{38.2} & 41.7 & 20.8 & \colorbox{wkred}{44.3} & \colorbox{wkred}{47.0} \\
Claude-3.5\cite{claude3} & \colorbox{wkred}{33.9} & \colorbox{wkred}{32.7} & \colorbox{wkred}{38.1} & \colorbox{wkblue}{34.3} & \colorbox{wkred}{46.7} & \colorbox{wkred}{33.3} & 29.8 & \colorbox{wkblue}{36.3} & \colorbox{wkred}{54.2} & \colorbox{wkred}{27.0} & \colorbox{wkblue}{38.2} & \colorbox{wkblue}{41.1} \\
\midrule

\multicolumn{13}{c}{Human Performance} \\
\midrule
Human (testmini) & 76.5 & 70.2 & 74.9 & 91.2 & 83.1 & 67.4 & 80.2 & 62.5 & 83.2 & 71.7 & 68.5 & 85.6 \\

\bottomrule
\end{tabular}
}
\vspace{-3mm}
\caption{Comparison of model performances across various mathematical subjects. AG: Analytic Geometry, MG: Metric Geometry, TG: Transformation Geometry, SG: Solid Geometry, CG: Combinatorial Geometry, DG: Descriptive Geometry. The \colorbox{wkred}{first} and \colorbox{wkblue}{second} highest accuracy of LMMs are marked in {red} and {blue}, respectively.}
\label{tab:main_model_performance}
\vspace{-5mm}
\end{table*}

We conduct extensive experiments on MV-MATH to evaluate a diverse range of models, including 18 open-source and 7 API-based models, assessing both LMMs and MLLMs. Our evaluation encompasses three conditions: text-only, text with image captions, and text with images. Results indicate that models perform better when images are input sequentially rather than merged, with mutually dependent image tasks proving more challenging. Experimental outcomes show that even the top-performing model, Claude 3.5-Connect, falls significantly short of human-level performance, underscoring considerable room for improvement in mathematical multi-visual reasoning capabilities.

Our evaluation is conducted under three settings: original answering, Chain of Thought (CoT), and CoT with 2-shot. To establish a human performance baseline, we use the testmini set and recruit 50 high school students to independently complete the questions. For multiple-choice, single-step, and multi-step free-form questions, we carefully design specific prompts to ensure that models generated responses in the correct format. Answer extraction and evaluation are performed using the Deepseek API\footnote{\url{https://www.deepseek.com/}}, with prompts specifically crafted for each question type to optimize model evaluation accuracy, more detailed information can be found in the Appendix.

\subsection{Main Results}
\noindent \textbf{Challenging Nature of MV-MATH:} The data presented in ~\Cref{tab:main_model_performance} illustrates the inherent challenges of MV-MATH. Notably, the overall accuracy of the leading model, Claude-3.5-Sonnet, is only 33.9\%, which falls significantly short of the human level accuracy of 76.5\%. Among the open-source models, only LLaVA-OneVision-72B (LLaVA-OV-72B) achieves an accuracy above 20\% on MV-MATH, while all others fall below this threshold. Notably, Math-LLaVA, a specialized mathematical MLLM, demonstrates limited performance on the MV-MATH dataset, with an accuracy of only 3.0\%. Our analysis suggests that this underperformance stems from the distribution of its training data. Specifically, Math-LLaVA trained on the Math360K dataset do not effectively generalize to the MV-MATH dataset, underscoring a significant limitation in its generalization capabilities. This finding underscores a challenge for current mathematical MLLMs: the need to achieve robust performance on both in-distribution and out-of-distribution data. Future advancements should prioritize strengthening generalization capabilities over memorizing training data.

\noindent \textbf{Disparity of Closed-source and Open-source Models:} 
Most open-source models exhibit a substantial performance gap compared to closed-source counterparts. Notably, Claude-3.5-Sonnet achieves the highest accuracy on the MV-MATH dataset at 33.9\%, while LLaVA-OV-Chat-72B attains a commendable 26.2\%, thereby narrowing the performance disparity between open-source and closed-source models. Models with smaller parameter sizes, such as the 7B and 13B variants, generally attain accuracy levels between 10\% and 20\%, with the top-performing LLaVA-OV-Chat-7B reaching 19.0\%.

\begin{figure*}[!t]
    \centering
    \begin{subfigure}[b]{0.49\linewidth}
        \centering
        \includegraphics[width=\linewidth]{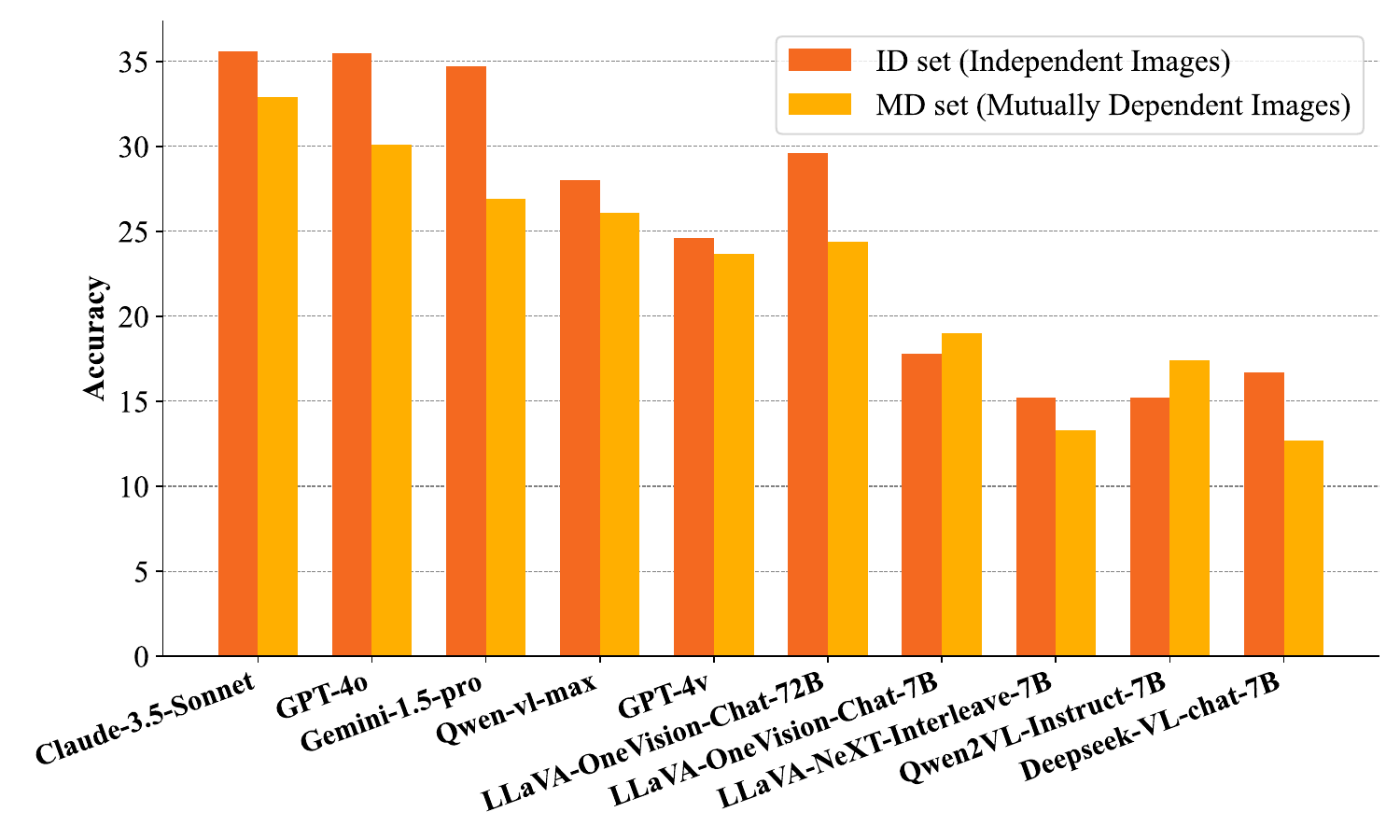}
        \caption{Performance Evaluation of Models on ID and MD Sets.}
        \label{fig:image_relavance}
    \end{subfigure}
    \hfill
    \begin{subfigure}[b]{0.49\linewidth}
        \centering
        \includegraphics[width=\linewidth]{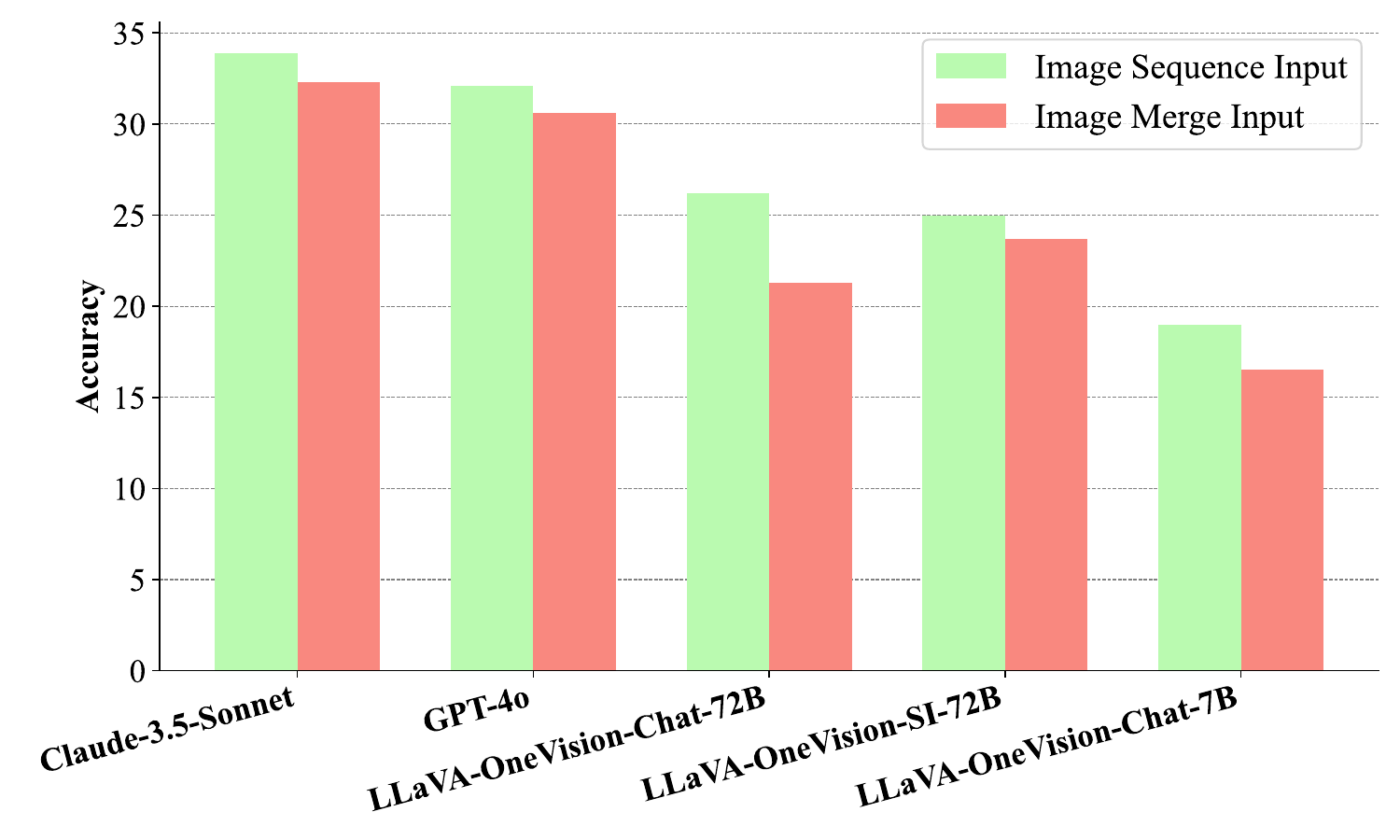}
        \caption{Comparison of Accuracy with Merge and Sequence Image Inputs.}
        \label{fig:merge_sequence}
    \end{subfigure}
    \vspace{-3mm}
    \caption{Evaluation of Model Performance across Different Image Relevance (ID vs. MD) and Input Methods (Merged vs. Sequential).}
    \label{fig:combined_figure}
    \vspace{-5mm}
\end{figure*}

\noindent \textbf{Comparison among Different Subjects:} The dataset encompasses 11 distinct mathematical subjects, ranging from algebra to geometry, each characterized by unique features. Model performance tends to be weaker in subjects requiring a higher level of image understanding, such as combinatorial geometry, where even the most advanced model, Claude-3.5-Sonnet, achieves an accuracy of only 27.0\%. In contrast, Claude’s accuracy in arithmetic reaches 54.2\%. This disparity indicates that current MLLMs lack the necessary capabilities to effectively process complex images and comprehend relationships between multiple intricate images. Significant improvements are still needed in fine-grained visual tasks, particularly in geometric perception and understanding.

\noindent \textbf{Failure of CoT/few-shot:}
As shown in~\Cref{tab:CoT&shot}, we observe that CoT and few-shot do not always improve model performance on MV-MATH. While Claude performs 3.4\% better with CoT and 1.3\% better with 2-shot based on CoT, GPT4o performs best under CoT with 2-shot setting, while Qwen-vl-max and gpt-4v perform best under CoT setting. For Gemini-1.5-pro and all open-source models, the original versions outperform CoT and CoT with 2-shot prompts. Notably, for all open-source models, performance declines steadily with the introduction of CoT and 2-shot prompting. We conduct further analysis on different types of questions in Appendix, and we find that original prompts perform better for multiple-choice questions, while CoT tends to produce better results for free-form questions. Adding 2-shot on top of CoT tends to degrade model performance on MV-MATH. Similar results are observed in Math-Vision~\cite{mathvision}.

\subsection{Analysis of Question type and Difficulty}
\noindent \textbf{Different Question Types.} We compare the performance of various models on three types of questions. As shown in~\Cref{fig: lidar}(right), the accuracy of the model on multiple-choice questions is significantly higher than on free-form questions. We design two evaluation metrics for multi-step questions: Step Accuracy Rate (\textbf{SAR}) and Question Completeness Rate (\textbf{QCR}). SAR refers to the proportion of correctly answered steps out of the total steps, while QCR is the proportion of questions for which all steps were answered correctly. The model with the best performance on the QCR metric is GPT-4o, achieving only 6\%, while its corresponding SAR is 32.6\%. This reflects the model's insufficient ability to perform complex reasoning on open and multi-step questions, and also reveals that the multiple-choice question format may not truly reflect the model's actual reasoning and problem-solving ability, as they usually rely on prompts to identify options.

\noindent \textbf{Different Difficulty Levels.} ~\Cref{tab:result_diffculty} compares the performance of selected models across three difficulty levels. GPT-4o demonstrates significantly higher proficiency with a success rate of 40.3\% in the "Easy" category. For the "Medium" category, Claude achieves the best performance at 37.5\%. The performance disparity further decreases in the "Hard" category, where the leading Claude model reaches only 26.6\% accuracy. As the difficulty level increases, the performance gap between models becomes smaller. Of particular note, Claude is the only model whose performance on medium difficulty exceeds that on low difficulty, whereas the performance of other models progressively declines as the difficulty level increases.

\subsection{Impact of Image Relevance and Input Methods}
\label{subsec: Image Relevance}
\noindent \textbf{Image Relevance, Mutually Dependent \textit{vs.} Independent:} To facilitate a more detailed analysis of MLLMs on mathematical multi-visual tasks, we conduct experiments on mutually dependent sets (MD) and independent sets (ID) (as detailed in ~\Cref{subsec: dataset_overview})in MV-MATH. While the ID set includes multiple images, they are not interrelated. In contrast, the MD set comprises mutually dependent images, demanding a higher level of cross-image understanding. As shown in ~\Cref{fig:image_relavance}, with the exception of Qwen2VL-7B and LLaVA-OV-Chat-7B, all models exhibit lower performance on the MD set compared to the ID set, with Gemini-1.5-pro demonstrating the largest performance gap between the two, reaching 7.8\%. This observation suggests that most models face challenges in effectively handling interdependent image tasks in mathematical scenarios, highlighting the potential limitations of handling cross-image interdependencies in mathematical multi-visual contexts.

\noindent \textbf{Input Methods, Merge \textit{vs.} Sequential:} To investigate the impact of image input methods on model performance, we apply both merge and sequence input approaches on MV-MATH , with results presented in~\Cref{fig:merge_sequence}. The findings demonstrate that the sequence input method outperforms the merge input method across all tested models, indicating that preserving the positional and sequential information of images is crucial for effective multi-image mathematical reasoning. This superior performance of sequence input underscores the importance of structured visual information in enhancing model capabilities for interpreting and processing complex mathematical scenarios. An interesting observation is that, while the LLaVA-OV-SI-72B model underperforms compared to the LLaVA-OV-Chat-72B model on sequence input method, it outperforms the latter on merge input method. This shift may be attributed to the initial single-image training of LLaVA-OV-SI, which enhances its ability to retain distinct visual features within individual images when processed in a merged format.

%% file: sec/5_Analysis.tex
\section{Error Analysis and Model Insights}
\label{sec:analysis}

\begin{figure}[!t]
    \centering
    \vspace{-10mm}
    \includegraphics[width=0.8\linewidth]{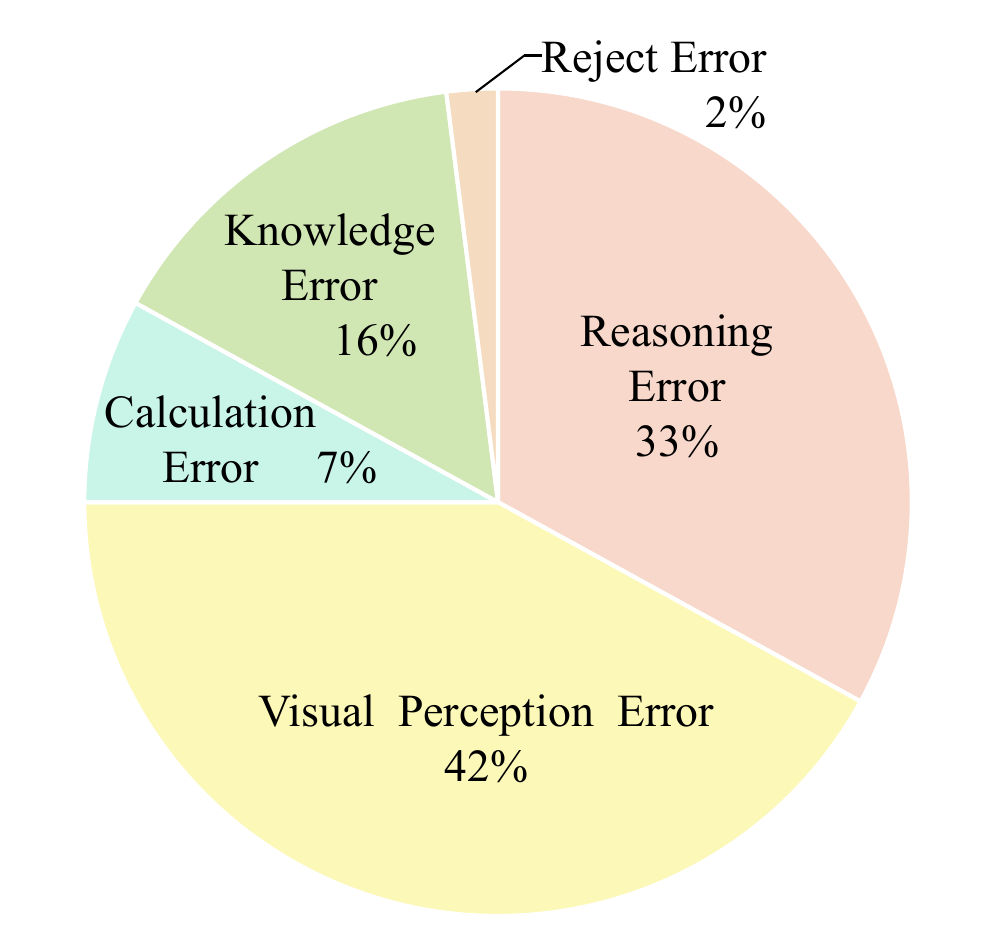}
    \caption{Error distribution over 100 annotated Claude errors.}
    \vspace{-5mm}
\label{fig:error_distribution}
\end{figure}

\paragraph{Error Analysis}We conduct a detailed error analysis of Claude-3.5-Sonnet. We randomly sample 100 errors, examine their distribution in detail, and provide corresponding qualitative examples. Based on this analysis, we classify the errors into five types, with the specific categories and their distribution shown in~\Cref{fig:error_distribution}.

\begin{figure}[t]
    \centering
    \includegraphics[width=1.00\linewidth]{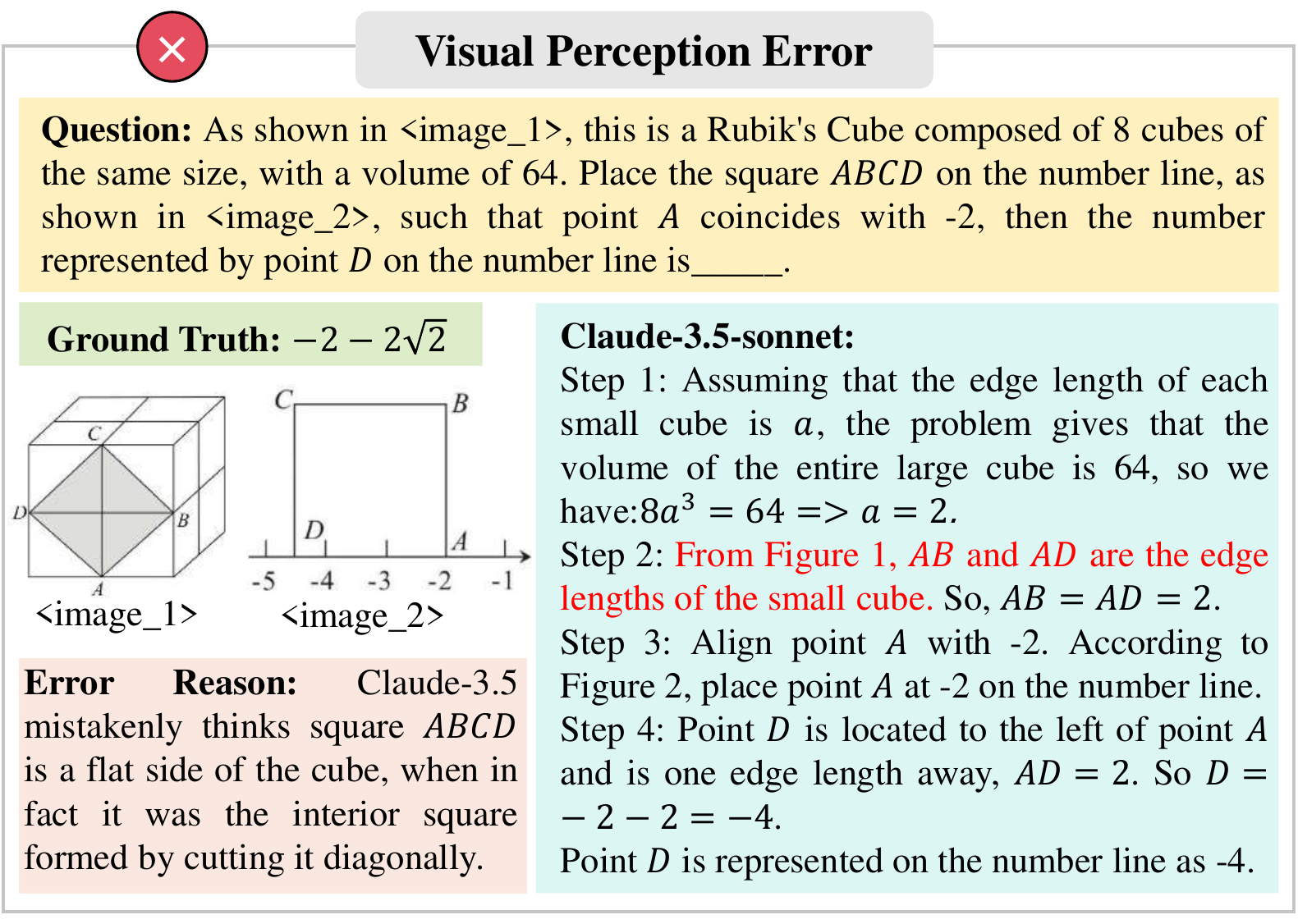}
    \vspace{-7mm}
    \caption{A basic visual perception error, with the error highlighted in red. More examples can be found in the appendix.}
    \label{fig:error_example}
    \vspace{-3mm}
\end{figure}

Visual perception error is the most common type, accounting for 42\%, which is higher than that of most single-image datasets such as MATH-Vision and MMMU~\cite{mathvision,MMMU}. This suggests that perceiving multi-visual contexts in the MV-MATH dataset presents a greater challenge compared to single-visual contexts. An example of a visual perception error in MV-MATH is shown in ~\Cref{fig:error_example}. Multi-visual perception requires MLLMs to not only accurately perceive and understand each image individually but also to comprehend the relationships between them. Reasoning errors also constitute a significant proportion. Even when the model correctly interprets text and images, it often fails to apply logical and mathematical reasoning effectively, leading to incorrect answers due to flaws in reasoning. Knowledge errors, stemming from the model's lack of relevant expertise, represent the third most common error type on MV-MATH, accounting for 7\% of errors. Calculation errors and reject errors make up 7\% and 2\%, respectively. Notably, we observe that all calculation errors are associated with symbolic computation rather than numerical computation.
\vspace{-5mm}
\paragraph{Model Insights}We analyze LLaVA-OV, the best-performing open-source model on MV-MATH, to gain insights for enhancing model capabilities in multi-visual mathematical contexts. LLaVA-OV’s strong performance is attributed to its staged training strategy and innovative architecture, which progressively increase the complexity of visual signals, thereby enhancing generalization across multi-image and multimodal tasks. As shown in ~\Cref{tab:main_model_performance}, LLaVA-OV-SI consistently underperforms compared to LLaVA-OV-SFT, while LLaVA-OV-Chat achieves the highest accuracy across both 7B and 72B model sizes. LLaVA-OV-SI is trained with a single-image stage; LLaVA-OV-SFT incorporates additional multi-image training to enhance image-text comprehension; and LLaVA-OV-Chat, building on LLaVA-OV-SFT, leverages Direct Preference Optimization (DPO) and human feedback to further improve generalization and reasoning in multi-visual contexts. The combination of multi-image training, DPO, and human feedback significantly advances multi-image mathematical reasoning, strengthening both generalization and inference capabilities in multi-visual scenarios.

%% file: sec/6_CONCLUSION.tex
\section{Conclusion}
\label{sec:conclusion}
The introduction of the MV-MATH dataset represents a major advancement in evaluating MLLMs’ multi-visual mathematical reasoning capabilities, addressing a critical gap in the assessment of multi-image mathematical tasks. MV-MATH encompasses a diverse and challenging array of multi-visual mathematical content. Using this dataset, we conduct a systematic and comprehensive evaluation of numerous open-source and closed-source models, establishing an extensive benchmark. This benchmark not only reveals a substantial performance gap between human capabilities and current MLLMs in multi-visual mathematical reasoning but also highlights the impact of image relavance, image input methods and question types on multi-visual math performance. Additionally, insights derived from the LLaVA-OV model offer guidance for enhancing performance in multi-image mathematical tasks, providing valuable direction for future advancements in this field.

%% file: sec/7_Acknowledgement.tex
\section*{Acknowledgment}
This work has been supported by the National Natural Science Foundation of China (NSFC) Grant 62436009.

%% file: appendix/X_suppl.tex
\clearpage
\setcounter{page}{1}
\maketitlesupplementary

\setcounter{section}{0}
\renewcommand{\thesection}{\Alph{section}}  % Use letters for appendix sections
\DoToC

\clearpage
\input{appendix/X_1_dataset}
\input{appendix/X_2_evaluation}
\input{appendix/X_3_mainresult}
\input{appendix/X_4_cot}
\input{appendix/X_5_imagerelevance}
\input{appendix/X_6_datalabeling}

\input{appendix/X_7_comparison}
\input{appendix/X_8_casestudy}

%% file: appendix/X_1_dataset.tex
\section{More Details about MV-MATH}
In this chapter, we will introduce MV-MATH in more detail.
\subsection{Question Distribution}
\begin{figure}[!ht]
    \centering
\includegraphics[width=1.0\linewidth]{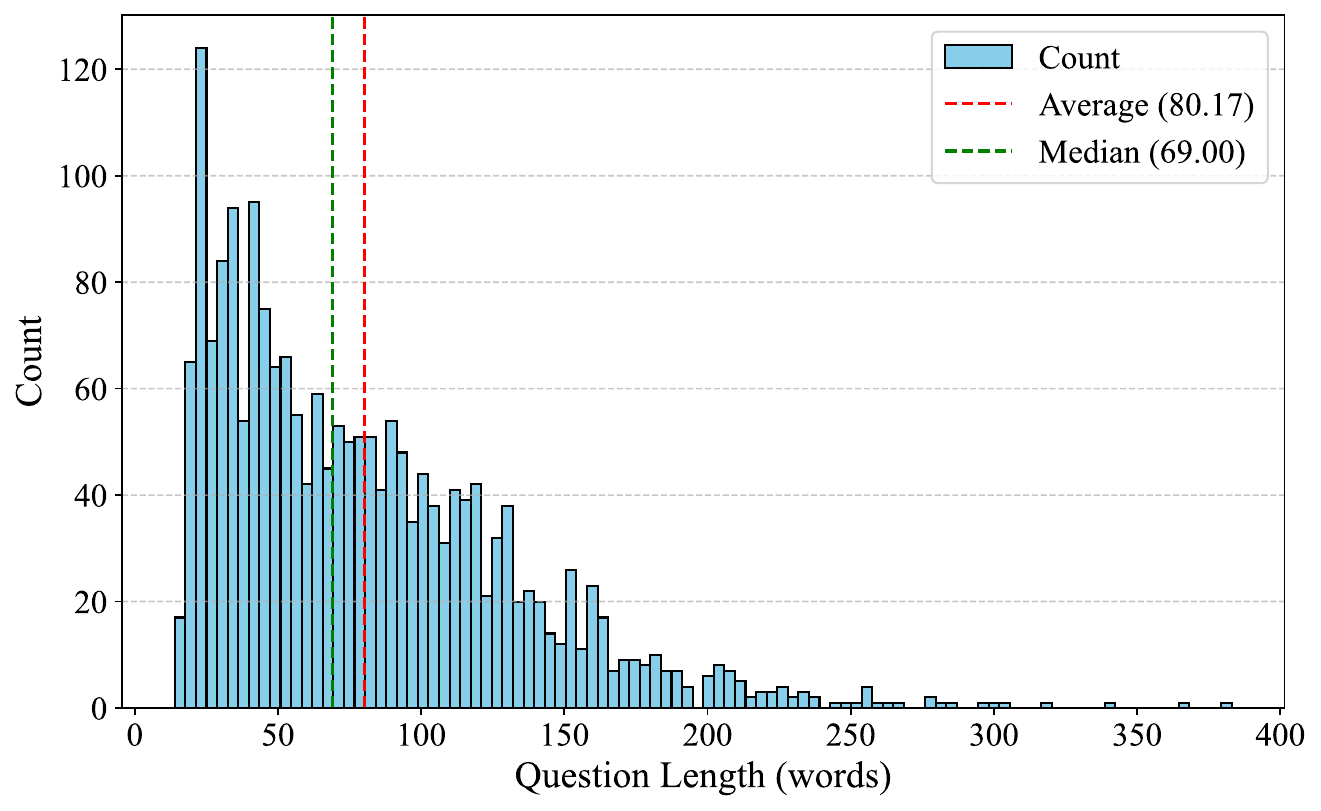}
    \caption{The distribution of the number of words per
question in MV-MATH.}
\label{fig:question_length_distribution}
\vspace{-5mm}
\end{figure}

All questions in MV-MATH are presented in English. As shown in~\Cref{tab:dataset_statistics}, the longest question in MV-MATH spans 383 words, while the shortest contains 14 words, with an average length of 80.17 words. \Cref{fig:question_length_distribution} further illustrates the distribution of text lengths, highlighting the diversity of MV-MATH.

\subsection{Image Distribution}
\begin{figure}[!ht]
    \centering
\includegraphics[width=1.0\linewidth]{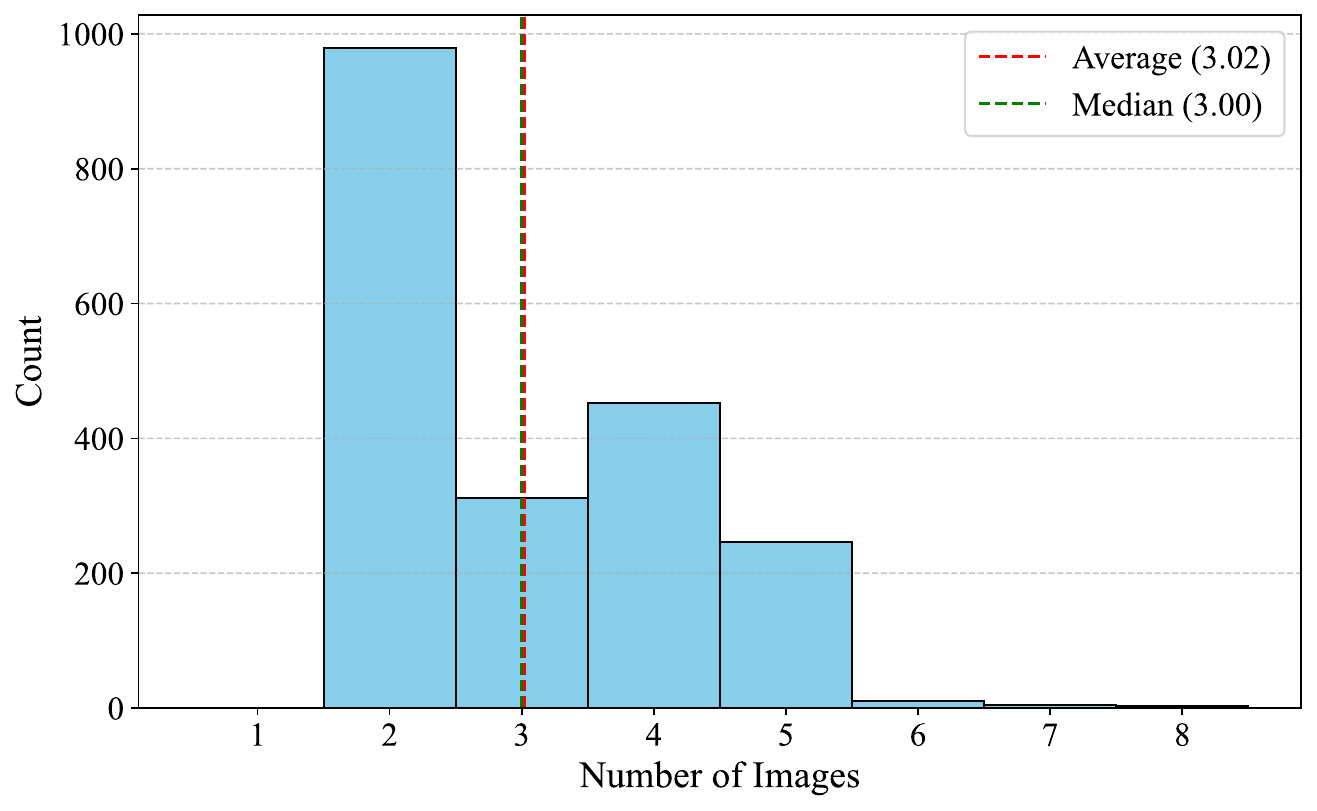}
    \caption{The distribution of the number of images per
question in MV-MATH.}
\label{fig:image_length_distribution}
\vspace{-5mm}
\end{figure}

\Cref{fig:image_length_distribution} illustrates the distribution of the number of images associated with each question in the MV-MATH dataset, highlighting its multimodal diversity. The dataset features an average of 3.02 images per question, with a median of 3, demonstrating a balanced and realistic allocation of visual resources. Most questions include 2,3 or 4 images, reflecting the dataset's emphasis on providing sufficient visual context for reasoning tasks. The presence of questions with 5 or more images showcases the dataset's capability to handle complex, multi-visual scenarios. This diversity ensures MV-MATH is suitable for evaluating the ability of models to integrate and reason across multiple interconnected visual elements, further enhancing its value as a benchmark for multimodal mathematical reasoning.

\subsection{Division of Difficulty Levels}
Since MV-MATH is a meticulously annotated dataset containing both answers and an analysis field, it offers a unique advantage for difficulty classification based on the lengths of the question and analysis. The difficulty classification is conducted in two steps.

\textbf{Step 1:} A weighted average is calculated using the lengths of the \textit{question} and \textit{analysis} fields, assigning a weight of 0.4 to the \textit{question} length and 0.6 to the \textit{analysis} length. This weighting reflects the assumption that the solution process (captured in the \textit{analysis} field) is more indicative of a question's difficulty. The weighted length distribution is shown in Figure 8. Using the K-means clustering algorithm, we cluster the weighted lengths into three difficulty levels: easy ($0–150$), medium ($150–500$), and hard ($>500$).

\textbf{Step 2:} Manual verification is performed on the clustering results from Step 1 by two graduate students. This process adjusts the difficulty classification to account for cases where the weighted length alone may not fully reflect complexity. For instance, a question with a short weighted length but involving intricate formula derivation is reclassified as harder, while a question with a long weighted length but straightforward reasoning is reclassified as easier.

\begin{figure}[!ht]
    \centering
\includegraphics[width=1.0\linewidth]{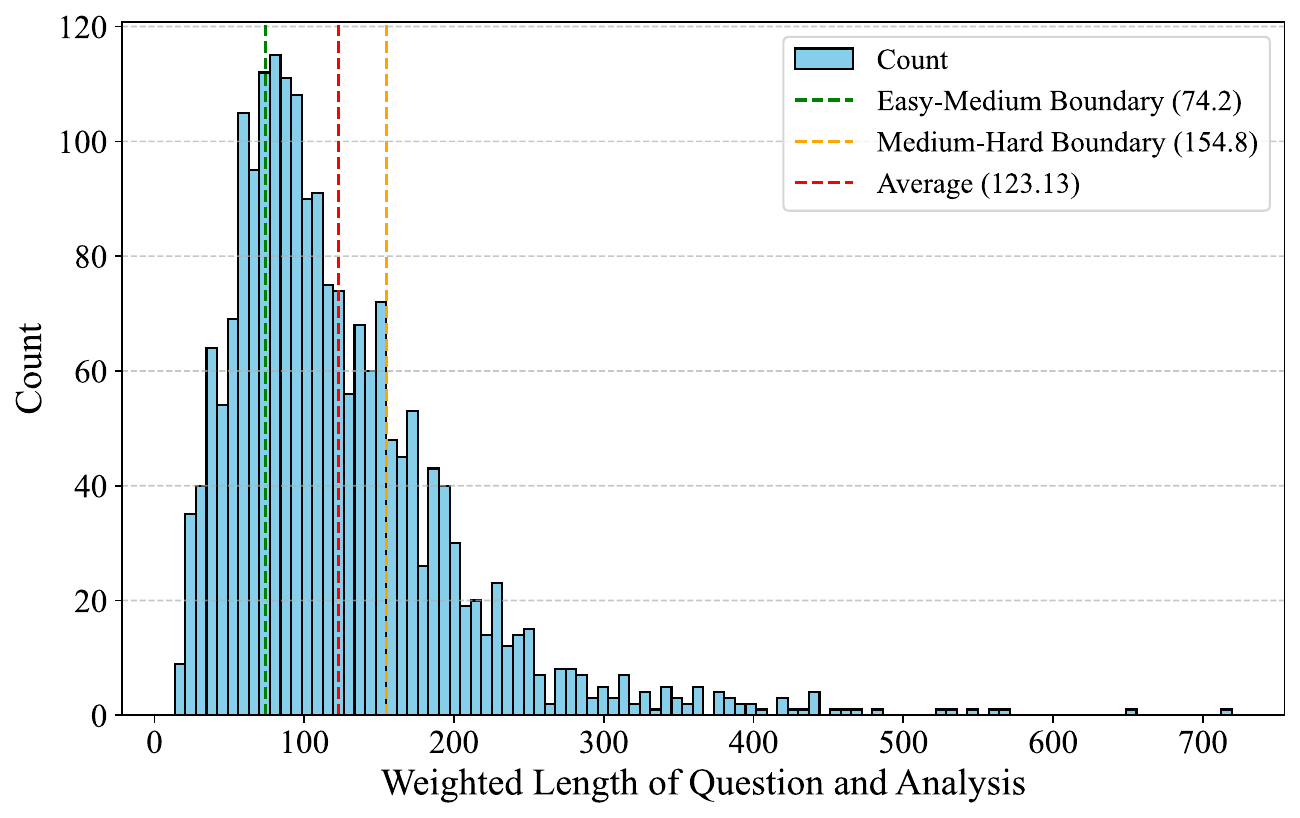}
    \caption{Weighted Length Distribution of Questions and Analysis for Difficulty Classification in MV-MATH}
\label{fig:weighted_length_distribution}
\vspace{-5mm}
\end{figure}

The detailed distribution of weighted lengths can be seen in~\Cref{fig:weighted_length_distribution}. After the two-step process described above, we classified the questions into 542 as \textit{easy}, 964 as \textit{medium}, and 503 as \textit{hard}.

\subsection{Image by Subjects}
\begin{figure*}[!htbp]
    \centering
\includegraphics[width=0.95\linewidth]{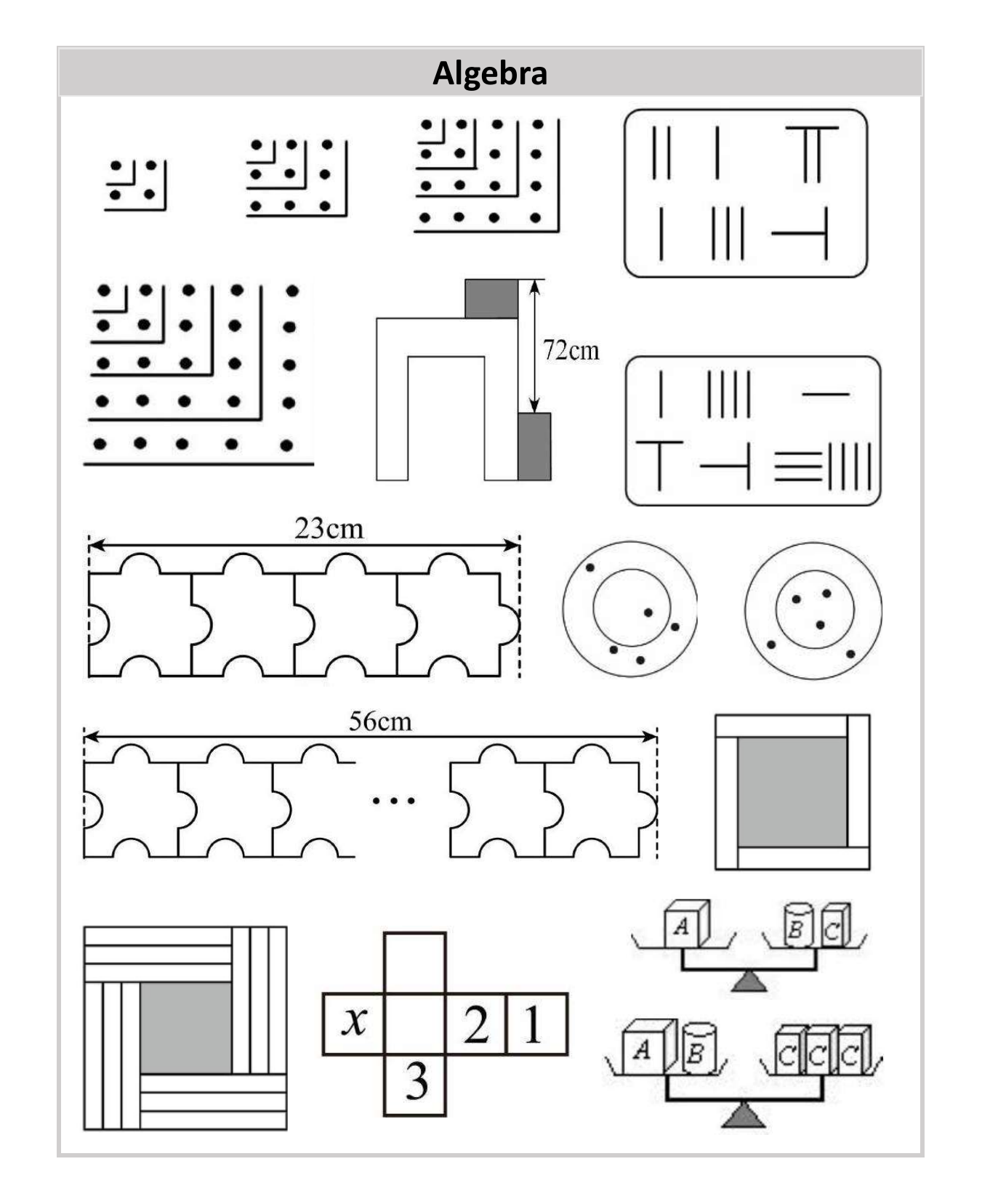}
    \caption{Some images from Analytic Algebra.}
    \label{fig: Algebra}
\end{figure*}

\begin{figure*}[!htbp]
    \centering
\includegraphics[width=0.95\linewidth]{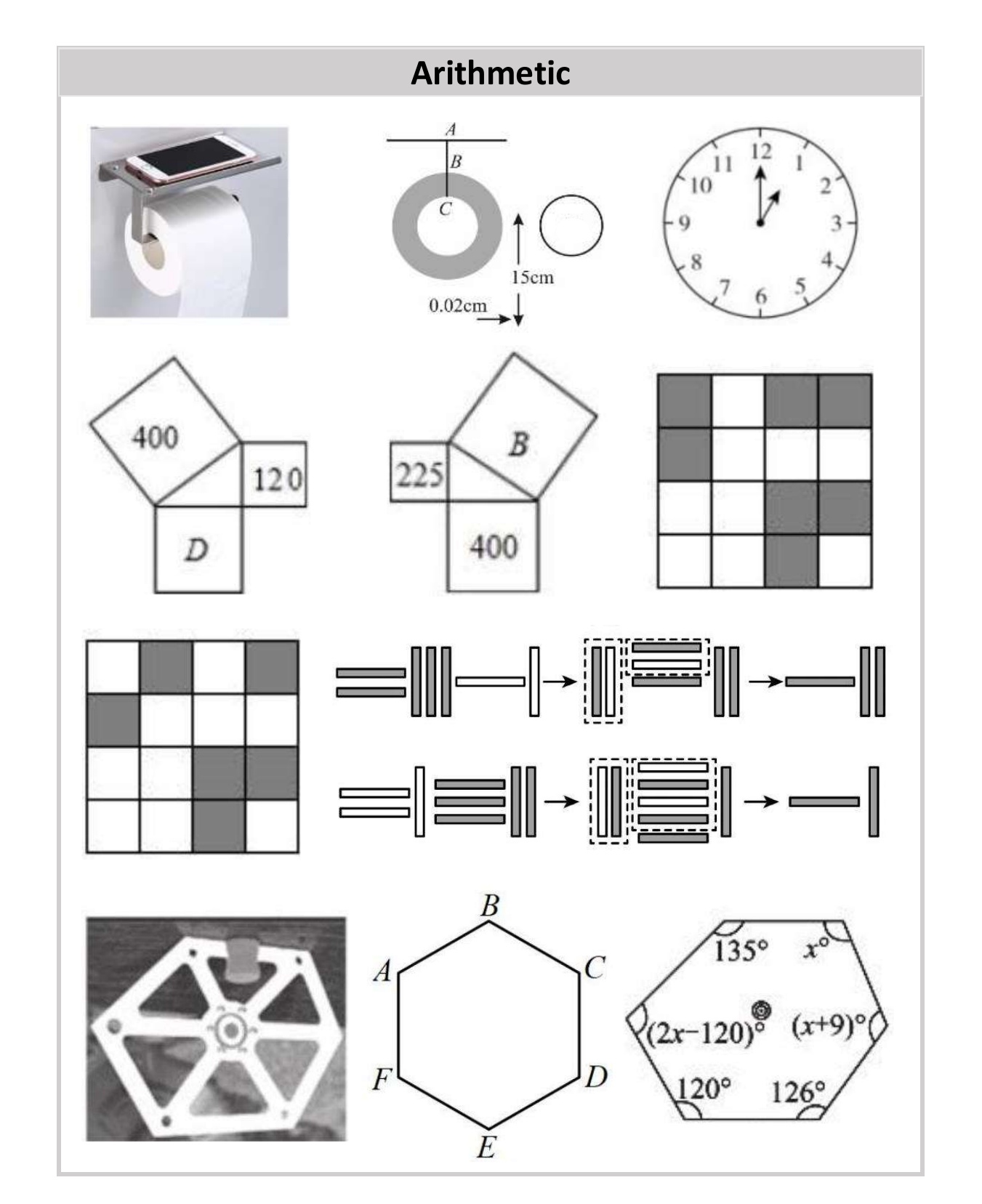}
    \caption{Some images from Arithmetic.}
    \label{fig: Arithmetic}
\end{figure*}

\begin{figure*}[!htbp]
    \centering
\includegraphics[width=0.95\linewidth]{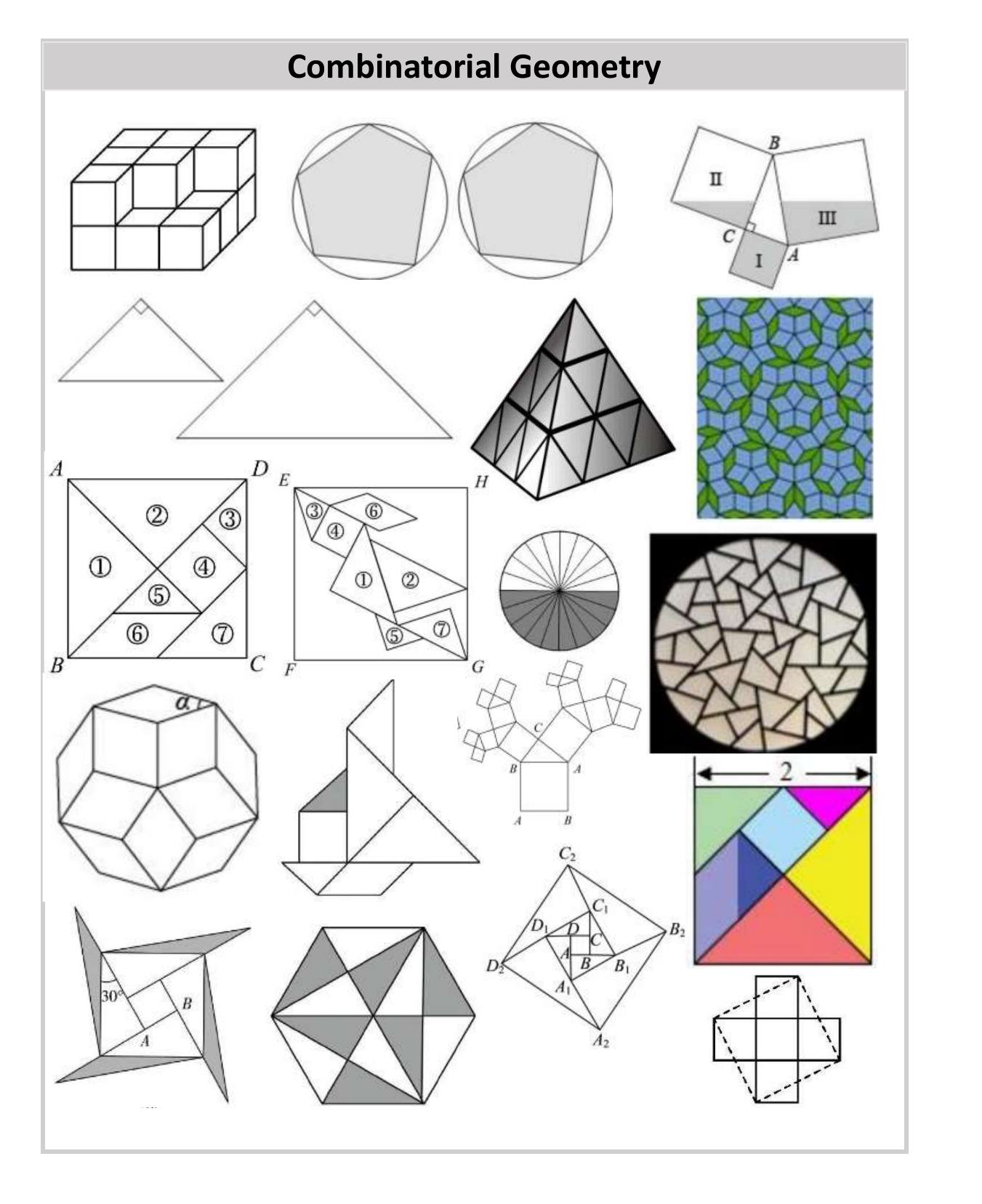}
    \caption{Some images from Combinatorial Geometry.}
    \label{fig: Combinatorial Geometry}
\end{figure*}

\begin{figure*}[!htbp]
    \centering
\includegraphics[width=0.95\linewidth]{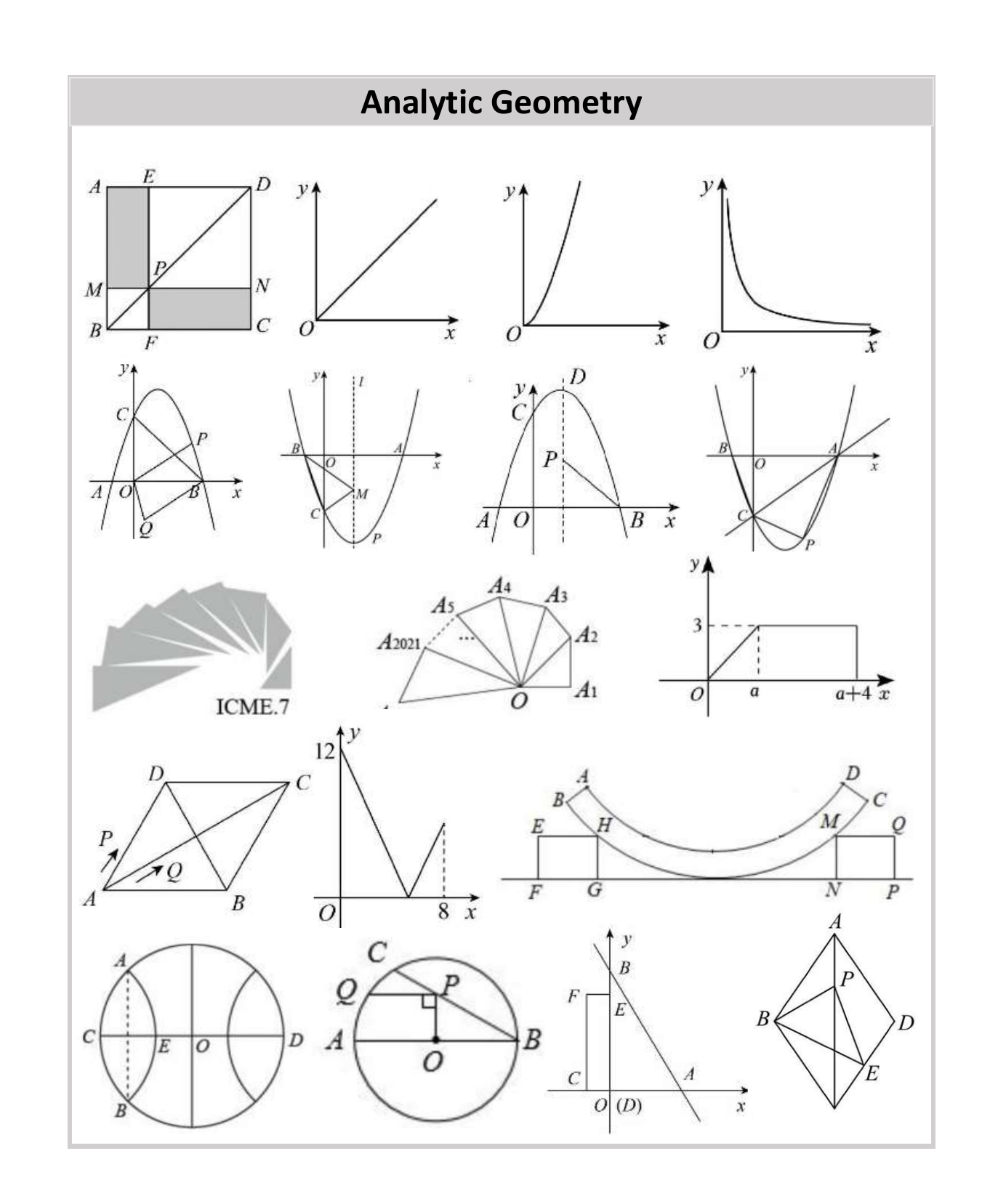}
    \caption{Some images from Analytic Geometry.}
    \label{fig: Analytic Geometry}
\end{figure*}

\begin{figure*}[!htbp]
    \centering
\includegraphics[width=0.95\linewidth]{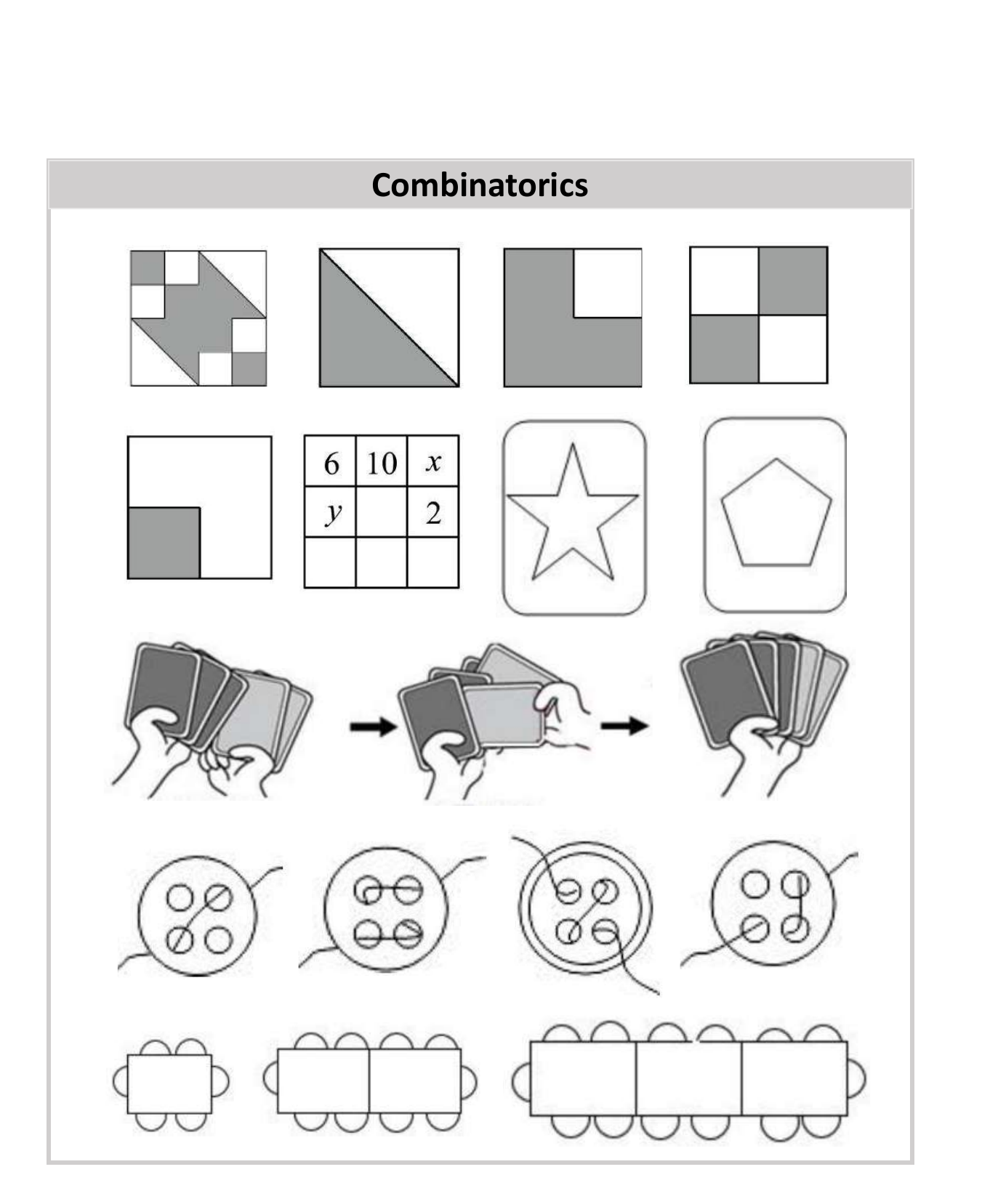}
    \caption{Some images from Combinatorics.}
    \label{fig: Combinatorics}
\end{figure*}

\begin{figure*}[!htbp]
    \centering
\includegraphics[width=0.95\linewidth]{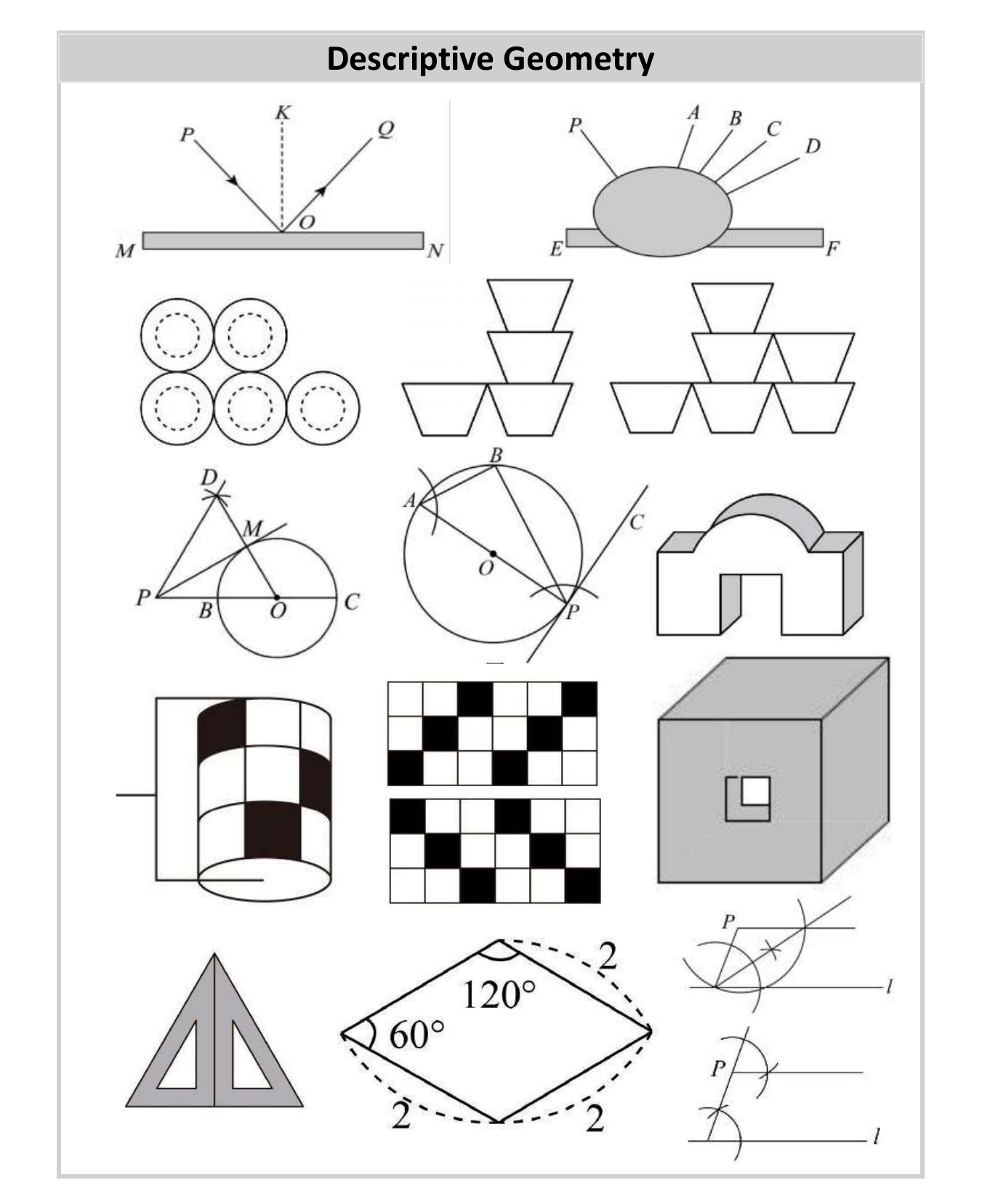}
    \caption{Some images from Descriptive Geometry.}
    \label{fig: Descriptive Geometry}
\end{figure*}

\begin{figure*}[!htbp]
    \centering
\includegraphics[width=0.95\linewidth]{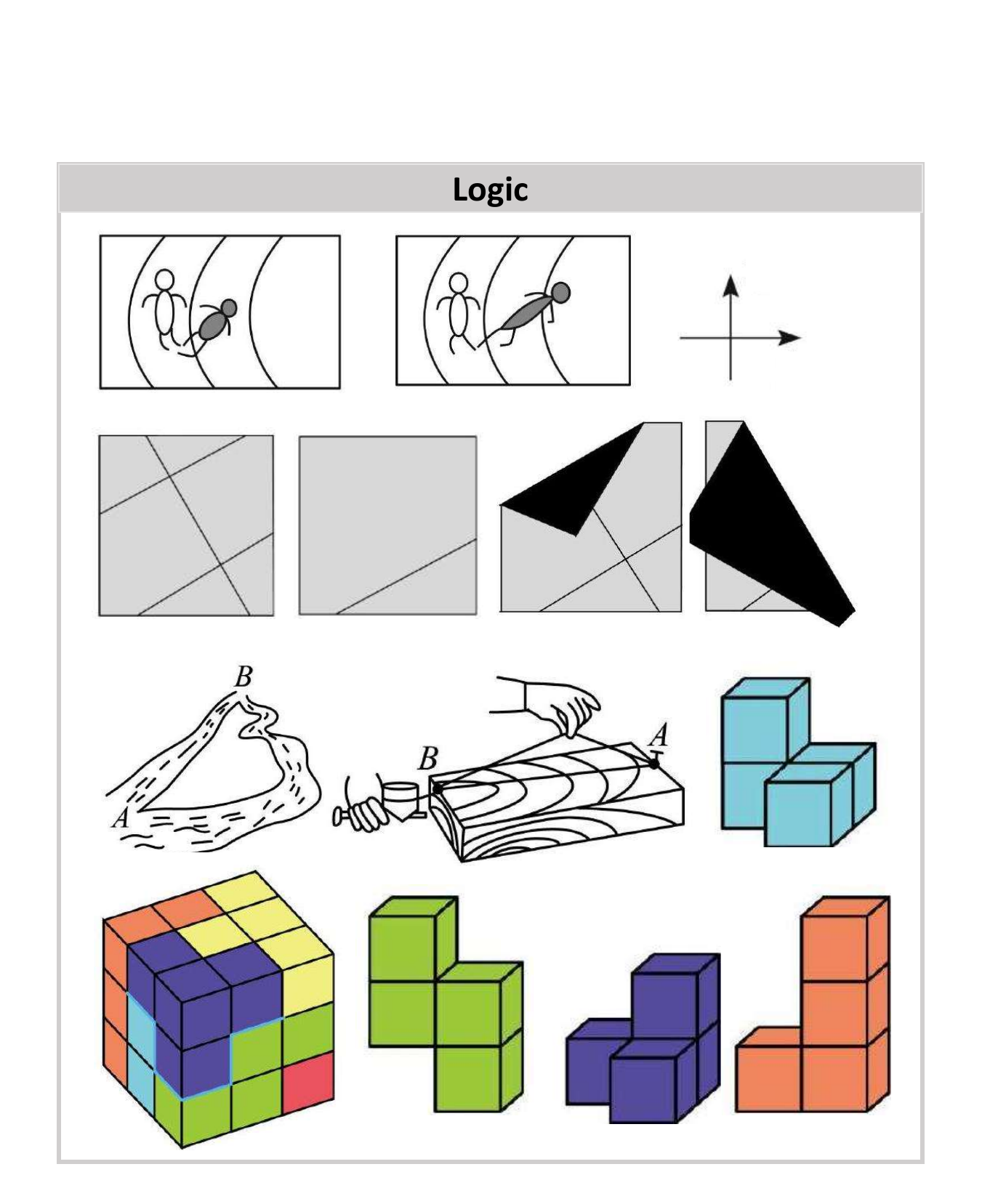}
    \caption{Some images from Logic.}
    \label{fig: Logic}
\end{figure*}

\begin{figure*}[!htbp]
    \centering
\includegraphics[width=0.95\linewidth]{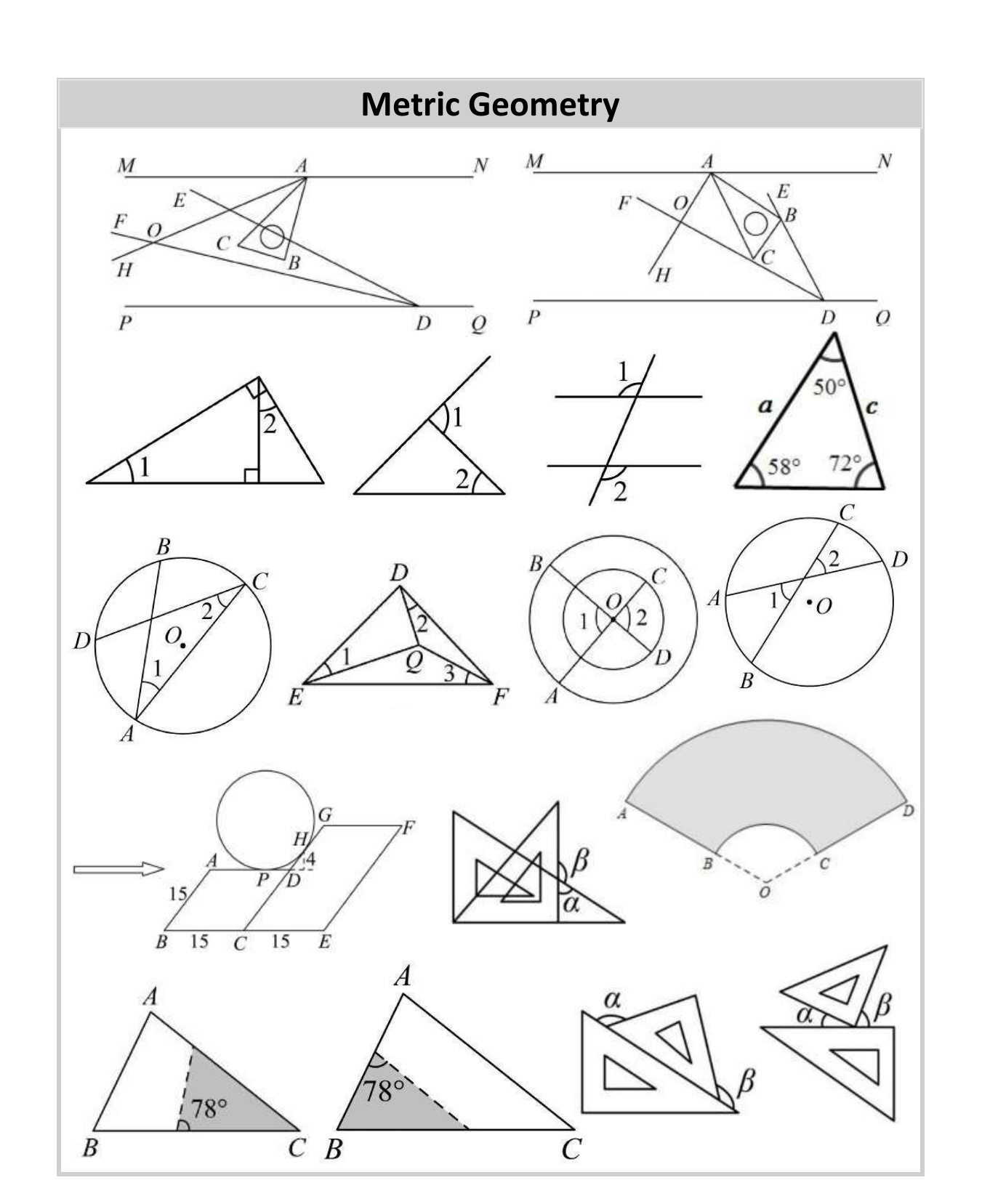}
    \caption{Some images from Metric Geometry.}
    \label{fig: Metric Geometry}
\end{figure*}

\begin{figure*}[!htbp]
    \centering
\includegraphics[width=0.95\linewidth]{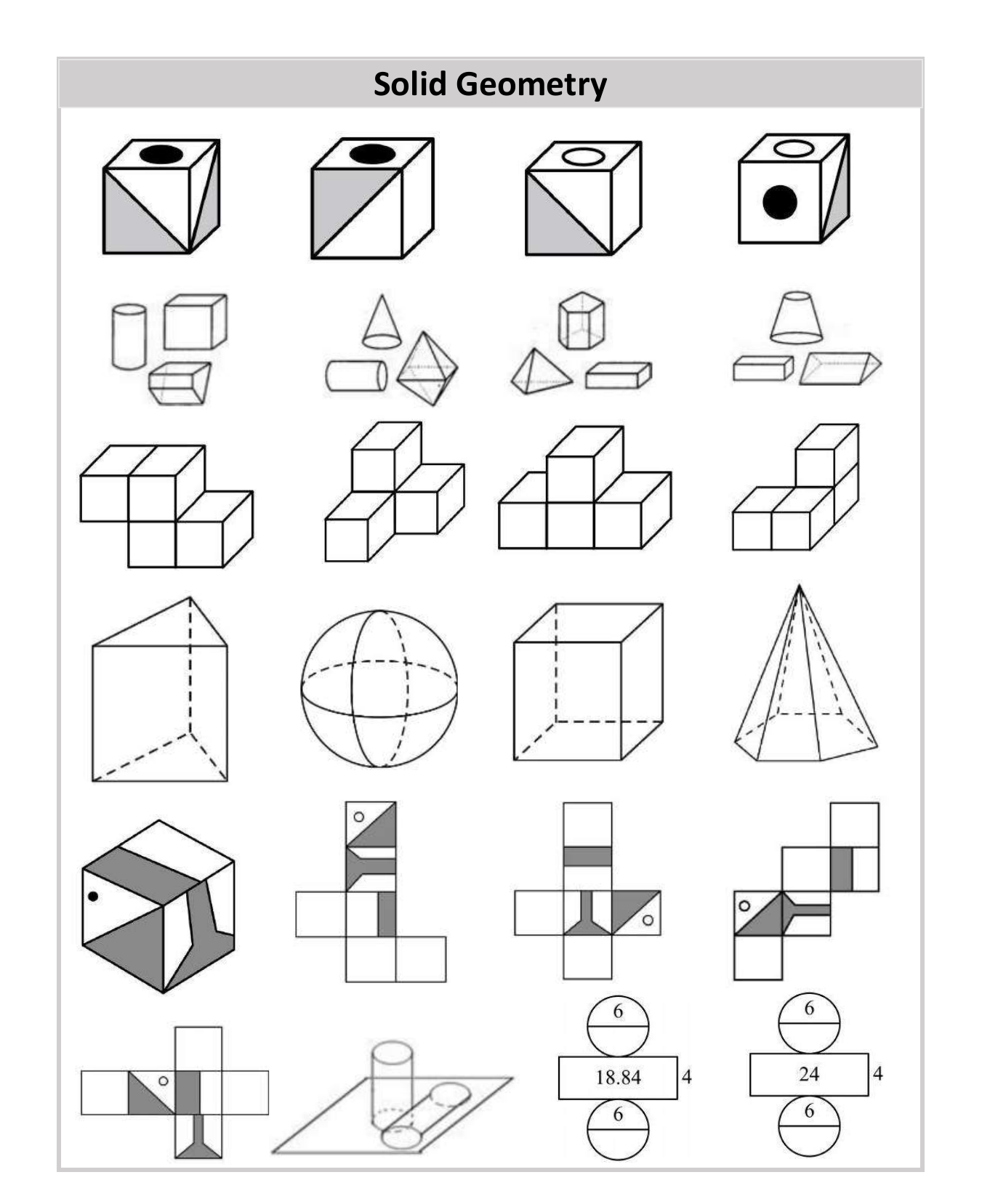}
    \caption{Some images from Solid Geometry.}
    \label{fig: Solid Geometry}
\end{figure*}

\begin{figure*}[!htbp]
    \centering
\includegraphics[width=0.95\linewidth]{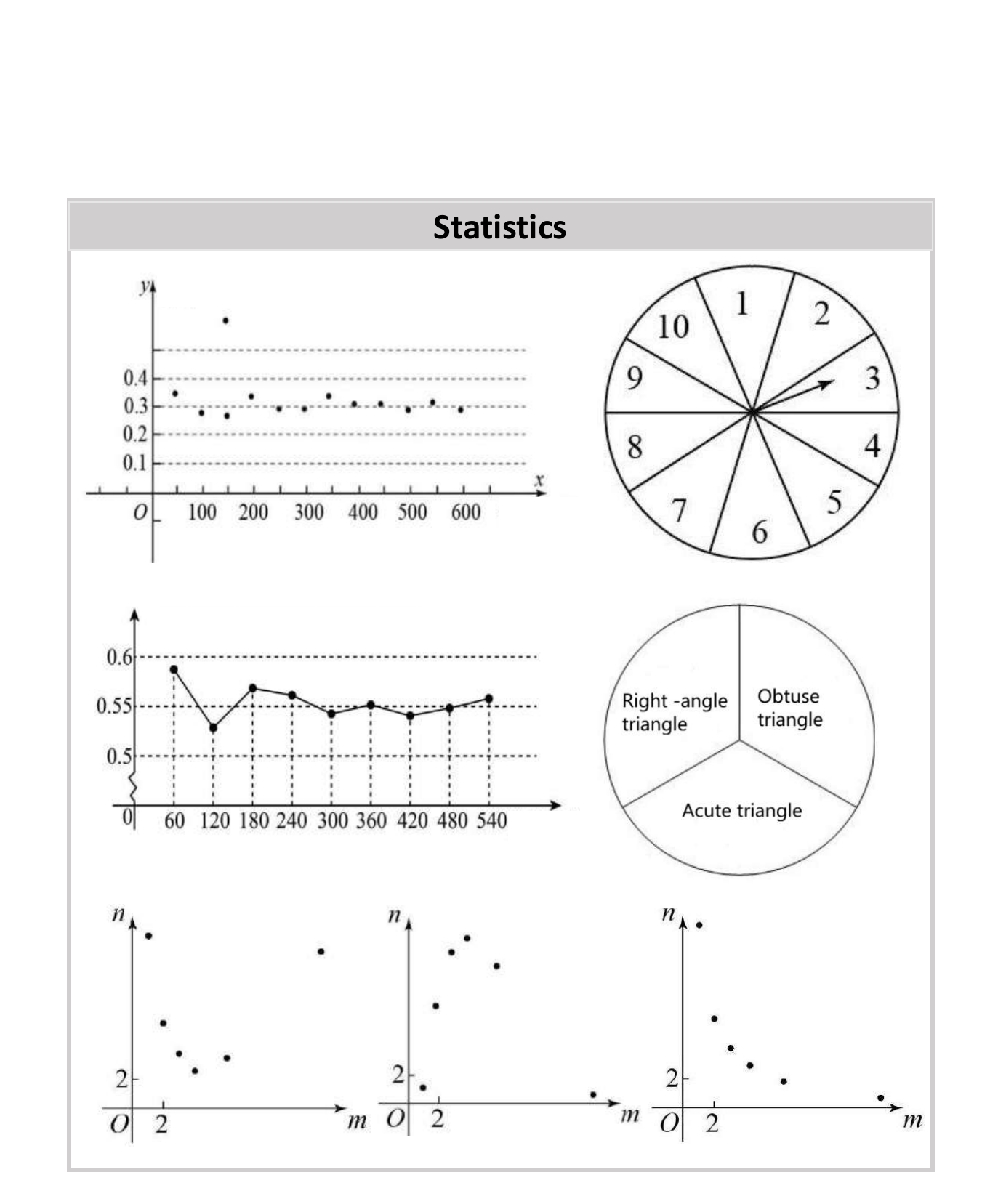}
    \caption{Some images from Statistics.}
    \label{fig: Statistics}
\end{figure*}

\begin{figure*}[!htbp]
    \centering
\includegraphics[width=0.95\linewidth]{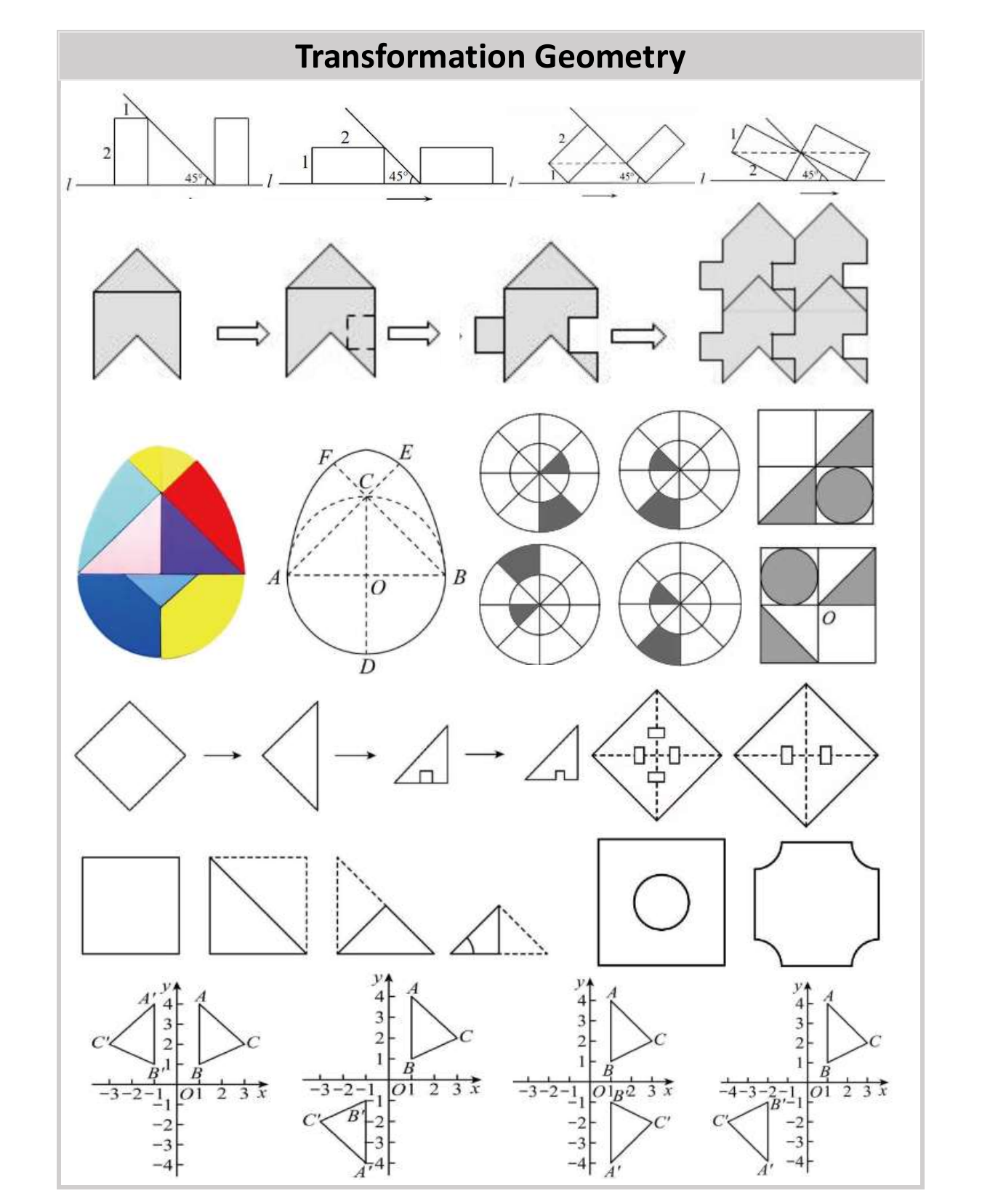}
    \caption{Some images from Transformation Geometry.}
    \label{fig: Transformation Geometry}
\end{figure*}

This section presents examples of images from various categories in the MV-MATH dataset. \Cref{fig: Algebra,fig: Arithmetic,fig: Combinatorial Geometry,fig: Analytic Geometry,fig: Combinatorics,fig: Descriptive Geometry,fig: Logic,fig: Metric Geometry,fig: Solid Geometry,fig: Statistics,fig: Transformation Geometry} correspond to images under the categories of \textit{Algebra}, \textit{Arithmetic}, \textit{Combinatorial Geometry}, \textit{Analytic Geometry}, \textit{Combinatorics}, \textit{Descriptive Geometry}, \textit{Logic}, \textit{Metric Geometry}, \textit{Solid Geometry}, \textit{Statistics}, and \textit{Transformation Geometry}, respectively.

The diversity of images across categories reflects the wide range of mathematical concepts covered in MV-MATH. For instance, images in the \textit{Statistics} category primarily include charts, graphs, and tables, providing rich visual representations of statistical data. In contrast, images under \textit{Arithmetic}, often targeted at younger students, feature elements with real-world objects or playful designs to aid comprehension. Similarly, the \textit{Combinatorics} and \textit{Combinatorial Geometry} categories include diagrams with intricate arrangements, requiring detailed reasoning. On the other hand, images in \textit{Analytic Geometry} and \textit{Metric Geometry} are more abstract and geometric, often involving coordinate systems, vectors, or precise measurements.

This variety of visual styles within and across categories highlights the versatility of MV-MATH in evaluating models' ability to interpret and reason over diverse visual contexts, making it a comprehensive benchmark for multi-visual mathematical reasoning tasks.

\subsection{Introduction of Subjects}
\textbf{Analytic Geometry.} Analytic Geometry integrates algebraic techniques with geometry through the use of a coordinate system. It provides a systematic method to describe geometric shapes using equations and to interpret these equations visually. This branch of mathematics enables a detailed study of geometric properties, such as distance, angles, and tangents.\vspace{2mm}\\
\textbf{Algebra.} Algebra is a fundamental area of mathematics that explores the use of symbols to represent numbers and quantities in formulas and equations. It encompasses a wide range of topics, from solving linear equations to studying abstract structures such as groups, rings, and fields, playing a crucial role in generalizing mathematical principles and solving problems where specific values are unknown.\vspace{2mm}\\
\textbf{Metric Geometry.} Metric Geometry focuses on the study of geometric figures based on distances and angles. It examines properties that remain invariant under transformations such as rotations, translations, and reflections. Metric Geometry serves as a foundation for applications in computer graphics, engineering, and physics.\vspace{2mm}\\
\textbf{Combinatorics.} Combinatorics is the branch of mathematics concerned with counting, arrangement, and combination of objects. It investigates problems related to discrete structures, such as permutations, combinations, and graph theory. Combinatorics is widely used in areas like cryptography, algorithm design, and network theory.\vspace{2mm}\\
\textbf{Transformation Geometry.} Transformation Geometry studies geometric transformations, such as translations, rotations, reflections, and dilations, to understand how shapes change while preserving certain properties. It provides insights into symmetry, congruence and has significant applications in computer vision.\vspace{2mm}\\
\textbf{Logic.} Logic in mathematics involves the study of principles of reasoning, including the formulation and analysis of valid arguments. It lays the groundwork for proof techniques and the development of formal systems. Mathematical logic is essential for understanding the structure of mathematical theories.\vspace{2mm}\\
\textbf{Solid Geometry.} Solid Geometry is the study of three-dimensional figures such as spheres, cubes, cones, and cylinders. It explores their properties, measurements, and spatial relationships, often involving volume, surface area, and intersections.\vspace{2mm}\\
\textbf{Arithmetic.} Arithmetic is the oldest and most fundamental branch of mathematics, dealing with numbers and basic operations such as addition, subtraction, multiplication, and division. It serves as the foundation for other mathematical disciplines and is essential for everyday problem-solving and quantitative reasoning.\vspace{2mm}\\
\textbf{Combinatorial Geometry.} Combinatorial Geometry combines concepts from combinatorics and geometry to study the arrangement and interaction of geometric objects. It addresses problems involving configurations of points, lines, and planes, often focusing on optimization and enumeration. This field has applications in graph theory, computational geometry, and optimization.\vspace{2mm}\\
\textbf{Descriptive Geometry.} Descriptive Geometry is a method for representing three-dimensional objects in two dimensions using projections. It involves techniques to visualize and solve spatial problems through precise drawings. This field is widely used in engineering and architecture for designing and visualizing complex structures.\vspace{2mm}\\
\textbf{Statistics.} Statistics is the study of data collection, analysis, interpretation, and presentation. It involves mathematical techniques to summarize and infer conclusions from data, often focusing on patterns, variability, and uncertainty. Statistics is indispensable in research, economics, and decision-making processes across various domains.

\clearpage

%% file: appendix/X_2_evaluation.tex
\clearpage

\section{Evaluation Details}
For open-source models, all experiments for 72B-scale models are conducted on H-800 GPUs, while the remaining models are conducted on A6000 GPUs.

\subsection{Prompt for Image Caption Generation}
We use Claude-3.5-Sonnet to generate image caption, the prompt is as follows: \textit{You are an expert in image description, here is a diagram of a math problem, you need to convert the image into text description in detail so that your description can be used to replace the diagram}.

\subsection{Prompt for Response Generation}
To ensure the model provides accurate responses, we design distinct CoT and 2-shot prompts tailored for multiple-choice, single-step, and multi-step free-form questions. The original prompt directly instructs the model to generate the final answer without intermediate reasoning. Detailed information can be found in Table~\Cref{tab:model_prompts,tab:single_step_prompts,tab:multi_step_prompts}.

\subsection{Prompt for Answer Evaluation}
Our evaluation is conducted using the Deepseek API. For the evaluation of multiple-choice, single-step, and multi-step free-form questions, different prompts are designed to ensure accuracy in answer extraction and assessment. We first use the Deepseek API to extract the model's answers and then it compares the extracted answers with the ground truth to determine the correctness of the answers. The specific prompts are shown in~\Cref{tab:deepseek_prompts} below.

\subsection{Model Details}  
All experiments are conducted using models configured with a temperature of 0.2 and a max\_new\_token limit of 2048 for text generation. Comprehensive details regarding the models utilized in the evaluation are presented in~\Cref{tab:model_sources}.

\clearpage
\begin{table*}[!b]
    \centering
    \begin{tabular}{p{0.8\textwidth}}
    \toprule
    \multicolumn{1}{c}{\textbf{Original}} \\
    \midrule
    You are an assistant for solving math problems. Your input consists of a math question and images, give your answer directly, without any intermediate steps. \\
    \midrule
    \multicolumn{1}{c}{\textbf{CoT (Chain of Thought)}} \\
    \midrule
    You are an assistant for solving math problems. Your input consists of a math question and images. Your task is to output the solution steps and the answer. The output format should be a step-by-step approach. Each question is multiple choice with one correct answer. Your final answer must be one of A, B, C, or D, and it should be placed within \{\}. For example: \{A\}, \{B\}, \{C\}, or \{D\}. \\
    \midrule
    \multicolumn{1}{c}{\textbf{CoT with 2-shot}} \\
    \midrule
    Example 1: \\
    Question: If a triangle has two sides of length 3 and 4, what is the length of the hypotenuse?  \\
    A.10  B.8  C.5  D.4 \\
    Answer: \\
    Step 1 (Mathematical theorem used: Pythagorean theorem): The Pythagorean theorem states that in a right triangle, the square of the hypotenuse is equal to the sum of the squares of the other two sides. The formula is: $c^2 = a^2 + b^2$, where $a$ and $b$ are the legs, and $c$ is the hypotenuse. \\
    Step 2 (Substitute the known values): Given $a = 3$ and $b = 4$. Substituting these values into the formula: $c^2 = 3^2 + 4^2 = 9 + 16 = 25$ \\
    Step 3 (Calculate the hypotenuse): Taking the square root gives: $c = \sqrt{25} = 5$ \\
    Answer: \{C\} \\

    Example 2: \\
    Question: In the right triangle ABC, AB is perpendicular to BC. It is known that AC=5 and AB=4. Find the area of the right triangle.  
    A.20  B.10  C.5  D.6 \\
    Answer: \\
    Step 1 (Mathematical theorem used: Pythagorean theorem): We first use the Pythagorean theorem to find the length of $BC$. The formula is: $AC^2 = AB^2 + BC^2$, where $AC$ is the hypotenuse, and $AB$ and $BC$ are the legs. \\
    Step 2 (Substitute the known values): Given $AC = 5$ and $AB = 4$. Substituting these values: $5^2 = 4^2 + BC^2 \implies 25 = 16 + BC^2$ \\
    Step 3 (Solve for $BC$): $BC^2 = 25 - 16 = 9 \implies BC = \sqrt{9} = 3$ \\
    Step 4 (Calculate the area): The area of the right triangle is given by $\frac{1}{2} \times AB \times BC$. Substituting the known values: $\text{Area} = \frac{1}{2} \times 4 \times 3 = 6$ \\
    Answer: \{D\} \\
    Your final answer must be one of A, B, C, or D, and it should be placed within \{\} \\
    \bottomrule
    \end{tabular}
    \caption{The prompts used for choice questions in the evaluation for response generation.}
    \label{tab:model_prompts}
\end{table*}
\clearpage

\clearpage
\begin{table*}
    \centering
    \begin{tabular}{p{0.80\textwidth}}
    \toprule
    \multicolumn{1}{c}{\textbf{Original Prompt}} \\
    \midrule
    You are an assistant for solving math problems. Your input consists of a math question and images. Give your answer directly, without any intermediate steps. \\
    \midrule
    \multicolumn{1}{c}{\textbf{CoT (Chain of Thought)}} \\
    \midrule
    You are an assistant for solving math problems. Your input consists of a math question and images. Your task is to output the solution steps and the answer. The output format should be a step-by-step approach. \\
    \midrule
    \multicolumn{1}{c}{\textbf{CoT with 2-shot}} \\
    \midrule
    Example 1: \\
    Question: If a triangle has two sides of length 3 and 4, what is the length of the hypotenuse? \\  
    Answer: \\  
    Step 1: (Mathematical theorem used: Pythagorean theorem): The Pythagorean theorem states that in a right triangle, the square of the hypotenuse is equal to the sum of the squares of the other two sides. The formula is: $c^2 = a^2 + b^2$, where $a$ and $b$ are the legs, and $c$ is the hypotenuse. \\  
    Step 2: (Substitute the known values): Given $a = 3$ and $b = 4$. Substituting these values into the formula: $c^2 = 3^2 + 4^2 = 9 + 16 = 25$. \\  
    Step 3: (Calculate the hypotenuse): Taking the square root gives: $c = \sqrt{25} = 5$. \\  
    Answer: {5} \\

    Example 2: \\
    Question: In the right triangle ABC, AB is perpendicular to BC. It is known that $AC = 5$ and $AB = 4$. Find the area of the right triangle. \\  
    Answer: \\  
    Step 1: (Mathematical theorem used: Pythagorean theorem): We first use the Pythagorean theorem to find the length of $BC$. The formula is: $AC^2 = AB^2 + BC^2$, where $AC$ is the hypotenuse, and $AB$ and $BC$ are the legs. \\  
    Step 2: (Substitute the known values): Given $AC = 5$ and $AB = 4$. Substituting these values: $5^2 = 4^2 + BC^2 \implies 25 = 16 + BC^2$. \\  
    Step 3: (Solve for $BC$): $BC^2 = 25 - 16 = 9 \implies BC = \sqrt{9} = 3$. \\  
    Step 4: (Calculate the area): The area of the right triangle is given by $\frac{1}{2} \times AB \times BC$. Substituting the known values: $\text{Area} = \frac{1}{2} \times 4 \times 3 = 6$. \\  
    Answer: {6} \\  

    Please reason step by step. Each step is placed on a new line, using the following format: Step X (Mathematical theorem/basis used): Detailed solution steps. Answer: \{\} \\
    \bottomrule
    \end{tabular}
    \caption{Prompts used for single-step free-form questions in the evaluation for response generation.}
    \label{tab:single_step_prompts}
\end{table*}
\clearpage

\clearpage
\begin{table*}
    \centering
    \begin{tabular}{p{0.80\textwidth}}
    \toprule
    \multicolumn{1}{c}{\textbf{Original Prompt}} \\
    \midrule
    You are an assistant for solving math problems. Your input consists of a math question and images. Each problem is a multi-step problem. Give your answer directly, without any intermediate steps. \\
    \midrule
    \multicolumn{1}{c}{\textbf{CoT (Chain of Thought)}} \\
    \midrule
    You are a math problem-solving assistant. Your input is a math problem and a picture of the problem. Each problem is a multi-step problem. Your task is to output the solution ideas and answers for each step. The output format is step-by-step. \\
    \midrule
    \multicolumn{1}{c}{\textbf{CoT with 2-shot Examples}} \\
    \midrule
    Example 1: \\
    Question: If a triangle has two sides of length 3 and 4, (1) what is the length of the hypotenuse? (2) what is the area of this triangle? \\
    Answer: \\
    (1) Step 1: (Mathematical theorem used: Pythagorean theorem): The Pythagorean theorem states that in a right triangle, the square of the hypotenuse is equal to the sum of the squares of the other two sides. The formula is: $c^2 = a^2 + b^2$, where $a$ and $b$ are the legs, and $c$ is the hypotenuse. \\
    Step 2: (Substitute the known values): Given $a = 3$ and $b = 4$. Substituting these values into the formula: $c^2 = 3^2 + 4^2 = 9 + 16 = 25$. \\
    Step 3: (Calculate the hypotenuse): Taking the square root gives: $c = \sqrt{25} = 5$. \\
    So the length of the hypotenuse is 5. \\
    (2) Step 1: The area of a right triangle is half the product of its two sides. \\
    Step 2: So the area of this triangle is $3 \times 4 / 2 = 6$. \\
    So the area of this triangle is 6. \\

    Example 2: \\
    Question: In the right triangle ABC, AB is perpendicular to BC. It is known that $AC = 5$ and $AB = 4$. (1) Find the area of the right triangle. (2) What is the height of the hypotenuse of this right triangle? \\
    Answer: \\
    (1) Step 1: (Mathematical theorem used: Pythagorean theorem): We first use the Pythagorean theorem to find the length of $BC$. The formula is: $AC^2 = AB^2 + BC^2$, where $AC$ is the hypotenuse, and $AB$ and $BC$ are the legs. \\
    Step 2: (Substitute the known values): Given $AC = 5$ and $AB = 4$. Substituting these values: $5^2 = 4^2 + BC^2 \implies 25 = 16 + BC^2$. \\
    Step 3: (Solve for $BC$): $BC^2 = 25 - 16 = 9 \implies BC = \sqrt{9} = 3$. \\
    Step 4: (Calculate the area): The area of the right triangle is given by $\frac{1}{2} \times AB \times BC$. Substituting the known values: $\text{Area} = \frac{1}{2} \times 4 \times 3 = 6$. \\
    So the area of the right triangle is 6. \\
    (2) Step 1: According to the equal area method, the area of a right triangle is equal to half the product of the two right-angled sides, which is also equal to half the product of the hypotenuse and the corresponding height. \\
    Step 2: According to the above principle and the conclusion of the first step, we can get $AB \times BC / 2 = AC \times h / 2$. \\
    Step 3: Substituting the values, we get $h = 3 \times 4 / 5 = 2.4$. \\
    So the height of the hypotenuse of this right triangle is 2.4. \\

    Please reason step by step. Each step is placed on a new line, using the following format: Step X (Mathematical theorem/basis used): Detailed solution steps. Answer:\{\} \\
    \bottomrule
    \end{tabular}
    \caption{Prompts used for multi-step free-form questions in the evaluation for response generation.}
    \label{tab:multi_step_prompts}
\end{table*}
\clearpage

\clearpage
\begin{table*}[!b]
    \centering
    \begin{tabular}{p{0.80\textwidth}}
    \toprule
    \multicolumn{1}{c}{\textbf{Multiple-Choice Prompt}} \\
    \midrule
    You are an assistant for evaluating math problems. Your task is to extract the model's answer to the given multiple-choice question and compare it with the ground truth. 

    Steps: \\
    1. Extract the model's answer. The answer must be one of A, B, C, or D. \\
    2. Compare the extracted answer with the ground truth. \\
    3. Indicate whether the model's answer is correct or incorrect. \\
    Output format: \\
    - Extracted Answer: \{A\}, \{B\}, \{C\}, or \{D\}. \\
    - Correctness: [true/false]. \\
    \midrule
    \multicolumn{1}{c}{\textbf{Single-Step Free-Form Prompt}} \\
    \midrule
    You are an assistant for evaluating math problems. Your task is to extract the model's answer to the given single-step free-form question and compare it with the ground truth. 

    Steps: \\
    1. Extract the model's final answer. \\
    2. Compare the extracted answer with the ground truth. \\
    3. Indicate whether the model's answer is correct or incorrect. \\
    Output format: \\
    - Extracted Answer: [Final Answer]. \\
    - Correctness: [true/false]. \\
    \midrule
    \multicolumn{1}{c}{\textbf{Multi-Step Free-Form Prompt}} \\
    \midrule
    You are an assistant for evaluating math problems. Your task is to extract the model's answers to each sub-question of a multi-step free-form problem and compare them with the ground truth. 

    Steps: \\
    1. Extract the final answers for each sub-question. \\
    2. Compare the extracted answers with the corresponding ground truth. \\
    3. Indicate whether each answer is correct or incorrect. \\
    Output format: \\
    - Sub-Question 1: Extracted Answer: [Answer]. Correctness: [true/false]. \\
    - Sub-Question 2: Extracted Answer: [Answer]. Correctness: [true/false]. \\
    \bottomrule
    \end{tabular}
    \caption{Prompts used for evaluating different types of math problems with the Deepseek API.}
    \label{tab:deepseek_prompts}
\end{table*}
\clearpage

\clearpage
\begin{table*}
    \small
    \centering
    \begin{tabular}{l|l|p{0.52\textwidth}}
    \toprule
    \textbf{Model}                                      & \textbf{Source} & \textbf{URL} \\
    \midrule
    Deepseek-chat             & Deepseek-chat & \url{https://api-docs.deepseek.com/} \\
    \midrule
    Math-LLaVA-13B                    & local checkpoint & \url{https://huggingface.co/Zhiqiang007/Math-LLaVA} \\
    \midrule
    LLaVA-v1.5-13B                    & local checkpoint & \url{https://huggingface.co/liuhaotian/llava-v1.5-13b} \\
    \midrule
    LLaVA-v1.5-7B                    & local checkpoint & \url{https://huggingface.co/liuhaotian/llava-v1.5-7b} \\
    \midrule
    \multirow{2}{*}{VILA-13B}                   & \multirow{2}{*}{local checkpoint} & \url{https://huggingface.co/Efficient-Large-Model/VILA-13b} \\
    \midrule
    \multirow{2}{*}{InternLM-XComposer2.5-VL-7B}                   & \multirow{2}{*}{local checkpoint} & \url{https://huggingface.co/internlm/internlm-xcomposer2d5-7b} \\
    \midrule
    InternVL-Chat-8B                  & local checkpoint & \url{https://huggingface.co/OpenGVLab/InternVL2-8B} \\
    \midrule
    \multirow{2}{*}{Llama-3.2-Vision-Instruct-11B}          & \multirow{2}{*}{local checkpoint} & \url{https://huggingface.co/meta-llama/Llama-3.2-11B-Vision-Instruct} \\
    \midrule
    \multirow{2}{*}{Deepseek-VL-7B}          & \multirow{2}{*}{local checkpoint} & \url{https://huggingface.co/deepseek-ai/deepseek-vl-7b-chat} \\
    \midrule
    \multirow{2}{*}{LLaVA-NeXT-Interleave-7B}           & \multirow{2}{*}{local checkpoint} & \url{https://huggingface.co/lmms-lab/llava-next-interleave-qwen-7b} \\
    \midrule
    \multirow{2}{*}{Mantis-Idefics2-8B}          & \multirow{2}{*}{local checkpoint} & \url{https://huggingface.co/TIGER-Lab/Mantis-8B-Idefics2} \\
    \midrule
    \multirow{2}{*}{Mantis-siglip-8B}          & \multirow{2}{*}{local checkpoint} & \url{https://huggingface.co/TIGER-Lab/Mantis-8B-siglip-llama3} \\
    \midrule
    Qwen2VL-Instruct-7B          & local checkpoint & \url{https://huggingface.co/Qwen/Qwen2-VL-7B-Instruct} \\
    \midrule
    \multirow{2}{*}{LLaVA-OneVision-SI-7B}          & \multirow{2}{*}{local checkpoint} & \url{https://huggingface.co/lmms-lab/llava-onevision-qwen2-7b-si} \\
    \midrule
    \multirow{2}{*}{LLaVA-OneVision-SFT-7B}          & \multirow{2}{*}{local checkpoint} & \url{https://huggingface.co/lmms-lab/llava-onevision-qwen2-7b-ov} \\
    \midrule
    \multirow{2}{*}{LLaVA-OneVision-Chat-7B}          & \multirow{2}{*}{local checkpoint} & \url{https://huggingface.co/lmms-lab/llava-onevision-qwen2-7b-ov-chat} \\
    \midrule
    \multirow{2}{*}{LLaVA-OneVision-SI-72B}          & \multirow{2}{*}{local checkpoint} & \url{https://huggingface.co/lmms-lab/llava-onevision-qwen2-72b-si} \\
    \midrule
    \multirow{2}{*}{LLaVA-OneVision-SFT-72B}          & \multirow{2}{*}{local checkpoint} & \url{https://huggingface.co/lmms-lab/llava-onevision-qwen2-72b-ov-sft} \\
    \midrule
    \multirow{2}{*}{LLaVA-OneVision-Chat-72B}          & \multirow{2}{*}{local checkpoint} & \url{https://huggingface.co/lmms-lab/llava-onevision-qwen2-72b-ov-chat} \\
    \midrule
    \multirow{2}{*}{InternLM-XComposer2-VL}  & \multirow{2}{*}{local checkpoint} & \url{https://huggingface.co/internlm/internlm-xcomposer2-vl-7b} \\
    \midrule
    \multirow{2}{*}{Qwen-VL-Plus}                & \multirow{2}{*}{qwen-vl-plus} & \url{https://help.aliyun.com/zh/dashscope/developer-reference/vl-plus-quick-start} \\
    \midrule
     GPT-4V       & gpt-4-vision-2023-05-15 & \url{https://platform.openai.com/} \\
    \midrule
    \multirow{2}{*}{Qwen-VL-Max}                & \multirow{2}{*}{qwen-vl-max} & \url{https://help.aliyun.com/zh/dashscope/developer-reference/vl-plus-quick-start} \\
    \midrule
    Gemini-1.5-Pro    & gemini-1.5-Pro-2023-05-15 & \url{https://ai.google.dev/} \\
    \midrule
    GPT-4o             & gpt-4o-2024-05-14 & \url{https://platform.openai.com/} \\
    \midrule
    Claude-3.5-Sonnet    & claude-3.5-sonnet-2024-05-24 & \url{https://www.anthropic.com/news/claude-3-5-sonnet} \\
    \bottomrule
    \end{tabular}
    \caption{The source of the models used in the evaluation.}
    \label{tab:model_sources}
\end{table*}
\clearpage

%% file: appendix/X_3_mainresult.tex
\section{Main Results Across 3 Question Types}
In this section, we present a detailed analysis of the experimental results across three question types: multiple-choice, single-step free-form, and multi-step free-form. Each subsection highlights the model's performance on these distinct question formats, providing insights into the effectiveness and limitations of the evaluated methods. The results are compared and discussed to showcase how different models handle varying levels of complexity and reasoning requirements.

The result of \textit{choice}, \textit{single-step} and \textit{multi-step} are shown in~\Cref{tab:choice,tab:single-step,tab:multi-step-SAR,tab:multi-step-QCR}. We observe that there is a performance gap between multiple-choice questions and free-form question types. Claude-3.5-Sonnet perform the best on multiple-choice questions with a score of 44.0\%, followed by GPT-4o and LLaVA-OneVision-Chat-72B. On single-step free-form questions, Claude still leads with an accuracy of 24.1\%, and the gap between open-source models and closed-source models has widened. For multi-step free-form questions, we calculate two metrics: Step Accuracy Rate (SAR) and Question Completeness Rate (QCR). GPT-4o achieves the highest scores in both SAR and QCR, with 32.0\% and 6.0\%, respectively. The models’ performance on QCR reflects their inability to perform complex multi-visual mathematical reasoning tasks. The multi-step part of the overall accuracy in ~\Cref{tab:main_model_performance} is calculated using the QCR metric.

\begin{table*}[p]
\centering
\resizebox{\textwidth}{!}
{%
\begin{tabular}{l|c|ccccccccccc}
\toprule
Model & Overall & AG  & Algebra & MG & Combinatorics & TG & Logic  & SG & Arithmetic & CG & DG & Statistics  \\
\midrule

\multicolumn{13}{c}{LLMs(Text-only, CoT with 2-shot)}\\
\midrule
Deepseek-Chat\cite{deepseekchat} & 27.0 & 26.3 & 32.6 & 19.4 & 31.8 & 28.5 & 9.0 & 29.9 & 27.2 & 24.8 & 28.1 & 13.2 \\
\midrule

\multicolumn{13}{c}{LLMs(Text + Image Caption, CoT with 2-shot)}\\
\midrule
Deepseek-Chat\cite{deepseekchat} & 27.5 & 27.7 & 30.1 & 18.4 & 36.8 & 30.5 & 10.9 & 32.9 & 26.8 & 26.8 & 28.7 & 17.3 \\
\midrule

\multicolumn{13}{c}{Open-source MLLMs (Text + Image, CoT with 2-shot)}\\
\midrule
Math-LLaVA-13B\cite{math-llava} & 3.6 & 2.3 & 7.9 & 3.7 & 9.0 & 2.0 & 0.0 & 1.7 & 18.1 & 5.8 & 2.3 & 5.8 \\
LLaVA-v1.5-13B\cite{llava} & 8.6 & 7.0 & 10.8 & 7.4 & 9.0 & 10.0 & 9.0 & 5.9 & 18.1 & 13.1 & 5.4 & 11.7 \\
LLaVA-v1.5-7B\cite{llava} & 18.0 & 15.7 & 18.8 & 20.3 & 9.0 & 18.0 & 13.6 & 21.4 & 0.0 & 16.7 & 25.7 & 11.7 \\
VILA-13B\cite{VILA} & 21.5 & 20.4 & 18.1 & 20.3 & 22.7 & 20.0 & 18.1 & 23.3 & 54.5 & 23.3 & 22.6 & 5.8 \\
InternLM-XComposer2.5-VL-7B\cite{InternLMXComposer-VL} & 23.0 & 21.0 & 20.2 & 24.0 & 31.8 & 27.0 & 36.3 & 16.7 & 27.2 & 24.0 & 25.0 & 5.8 \\
InternVL-Chat-8B\cite{internvl} & 25.9 & 23.9 & 31.8 & 25.9 & 31.8 & 27.5 & 27.2 & 25.1 & 36.3 & 22.6 & 21.8 & 23.5 \\
Llama-3.2-Vision-Instruct-11B\cite{llama} & 24.1 & 23.9 & 22.4 & 24.0 & 31.8 & 21.5 & 18.1 & 19.7 & 36.3 & 32.1 & 27.3 & 17.6 \\
Deepseek-VL-7B\cite{deepseekvl} & 26.0 & 26.3 & 32.6 & 19.4 & 31.8 & 28.5 & 9.0 & 29.9 & 27.2 & 24.8 & 28.1 & 29.4 \\
LLaVA-NeXT-Interleave-7B\cite{llava-next-interleave} & 26.6 & 23.9 & 23.1 & 26.8 & 31.8 & 25.5 & 18.1 & 31.7 & 9.0 & 30.6 & 26.5 & 23.5 \\
Mantis-Idefics2-8B\cite{mantis} & 26.6 & 22.8 & 27.5 & 31.4 & 27.2 & 27.0 & 22.7 & 23.9 & 9.0 & 29.9 & 28.9 & 11.7 \\
Mantis-siglip-8B\cite{mantis} & 27.8 & 29.8 & 27.5 & 28.7 & 22.7 & 27.0 & 22.7 & 26.9 & 27.2 & 21.8 & 35.1 & 17.6 \\
Qwen2VL-Instruct-7B\cite{Qwen2VL} & 27.8 & 21.0 & 28.2 & 24.0 & 31.8 & 30.0 & 22.7 & 32.3 & 27.2 & 32.1 & 26.5 & 23.5 \\
LLaVA-OneVision-SI-7B\cite{llava-onevision} & 26.3 & 23.9 & 28.2 & 21.2 & 22.7 & 26.5 & 13.6 & 27.5 & 45.4 & 23.3 & 34.3 & 13.6 \\
LLaVA-OneVision-SFT-7B\cite{llava-onevision} & 30.1 & 30.4 & 28.6 & 54.5 & 26.2 & 27.2 & 28.1 & 9.0 & 37.9 & 31.7 & 30.0 & 23.5 \\
LLaVA-OneVision-Chat-7B\cite{llava-onevision} & 31.5 & 32.7 & 29.7 & 34.2 & 22.7 & 26.0 & 4.5 & 37.7 & 54.5 & 32.8 & 32.0 & 29.4 \\
LLaVA-OneVision-SI-72B\cite{llava-onevision} & 35.8 & 34.5 & 34.0 & 38.8 & 50.0 & 36.0 & \colorbox{wkblue}{40.9} & 40.1 & 45.4 & 27.0 & 36.7 & 23.5 \\
LLaVA-OneVision-SFT-72B\cite{llava-onevision} & 37.0 & 29.8 & 42.0 & 28.7 & 36.3 & 37.5 & 31.8 & \colorbox{wkred}{47.9} & 18.1 & 32.8 & \colorbox{wkblue}{39.8} & 29.4 \\
LLaVA-OneVision-Chat-72B\cite{llava-onevision} & 38.0 & 32.1 & 42.0 & 33.3 & 54.5 & 37.5& 27.2 & {47.3} & 36.3 & 35.0 & 35.9 & 35.2 \\
\midrule

\multicolumn{13}{c}{Closed-source MLLMs (Text + Image, CoT with 2-shot)} \\
\midrule
Qwen-vl-plus\cite{qwenvl} & 27.7 & 30.4 & 25.3 & 25.0 & 18.1 & 29.0 & 31.8 & 33.5 & 45.4 & 22.6 & 24.2 & 29.4  \\
GPT-4V\cite{GPT-4V} & 32.5 & 21.0 & 34.7 & 44.4 & 36.3 & 32.5 & {36.3} & 31.7 & 45.4 & 32.1 & 33.5 & 35.2 \\
Qwen-vl-max\cite{qwenvl} & 37.3 & 35.0 & 38.4 & 29.6 & 
\colorbox{wkblue}{59.0} & 37.0 & 31.8 & 43.7 & 45.4 & \colorbox{wkblue}{37.2} & 32.8 & 23.5 \\
Gemini-1.5-Pro\cite{gemini} & 35.8 & \colorbox{wkblue}{36.2} & 36.2 & 30.5 & 31.8 & 39.0 & \colorbox{wkred}{40.9} & 40.1 & 
\colorbox{wkblue}{54.5} & 32.8 & 28.9 & 35.2 \\
GPT-4o\cite{GPT4o} & \colorbox{wkblue}{41.9} & {36.2} & \colorbox{wkblue}{46.3} & \colorbox{wkblue}{44.4} & 50.0 & \colorbox{wkblue}{40.0} & 31.8 & \colorbox{wkblue}{47.3} & 45.4 & 35.0 & \colorbox{wkred}{45.3} & \colorbox{wkred}{47.0} \\
Claude-3.5\cite{claude3} & \colorbox{wkred}{44.0} & \colorbox{wkred}{42.6} & \colorbox{wkred}{49.2} & \colorbox{wkred}{46.3} & \colorbox{wkred}{59.0} & \colorbox{wkred}{43.5} & 31.8 & 44.9 & \colorbox{wkred}{72.7} & \colorbox{wkred}{39.4} & {39.0} & \colorbox{wkblue}{41.1} \\
\midrule

\multicolumn{13}{c}{Human Performance} \\
\midrule
Human (testmini) & 80.2 & 75.2 & 71.9 & 95.2 & 85.1 & 70.4 & 80.2 & 67.5 & 85.2 & 76.7 & 69.8 & 88.4 \\

\bottomrule
\end{tabular}
}
\caption{Comparison of model performances across various mathematical subjects on the choice problem set. The \colorbox{wkred}{first} and \colorbox{wkblue}{second} highest accuracy of LMMs are marked in {red} and {blue}, respectively.}
\label{tab:choice}
\vspace{-5mm}
\end{table*}

\begin{table*}[htbp]
\centering
\resizebox{\textwidth}{!}
{%
\begin{tabular}{l|c|cccccccccc}
\toprule
Model & Overall & AG  & Algebra & MG & Combinatorics & TG & SG & Arithmetic & CG & DG \\
\midrule

\multicolumn{11}{c}{LLMs(Text-only, CoT with 2-shot)}\\
\midrule
Deepseek-Chat\cite{deepseekchat} & 3.0 & 3.1 & 2.3 & 4.0 & 5.5 & 2.5 & 3.0 & 0.0 & 3.3 & 10.0 \\
\midrule

\multicolumn{11}{c}{LLMs(Text + Image Caption, CoT with 2-shot)}\\
\midrule
Deepseek-Chat\cite{deepseekchat} & 4.3 & 5.2 & 3.4 & 5.3 & 0.0 & 1.2 & 3.0 & 20.0 & 2.9 & 20.0 \\
\midrule

\multicolumn{11}{c}{Open-source MLLMs (Text + Image, CoT with 2-shot)}\\
\midrule
Math-LLaVA-13B\cite{math-llava} & 2.5 & 0.0 & 5.8 & 6.6 & 0.0 & 3.8 & 4.5 & 20.0 & 7.2 & 0.0 \\
LLaVA-v1.5-13B\cite{llava} & 0.6 & 2.1 & 1.1 & 0.0 & 0.0 & 6.4 & 1.5 & 0.0 & 0.4 & 0.0 \\
LLaVA-v1.5-7B\cite{llava} & 0.3 & 0.0 & 2.3 & 0.0 & 11.1 & 1.2 & 1.0 & 0.0 & 2.9 & 0.0 \\
VILA-13B\cite{VILA} & 0.5 & 0.0 & 1.1 & 0.0 & 0.0 & 1.2 & 5.0 & 0.0 & 3.3 & 0.0 \\
InternLM-XComposer2.5-VL-7B\cite{InternLMXComposer-VL} & 1.0 & 1.0 & 2.3 & 0.0 & 16.6 & 6.4 & 3.5 & 0.0 & 3.8 & 10.0 \\
InternVL-Chat-8B\cite{internvl} & 0.3 & 1.0 & 4.6 & 8.0 & 5.5 & 1.2 & 3.0 & 20.0 & 2.9 & 0.0 \\
Llama-3.2-Vision-Instruct-11B\cite{llama} & 2.7 & 4.2 & 6.9 & 8.0 & 11.1 & 2.5 & 5.0 & 20.0 & 3.8 & 0.0 \\
Deepseek-VL-7B\cite{deepseekvl} & 1.5 & 0.0 & 2.3 & 0.0 & 0.0 & 0.0 & 3.0 & 0.0 & 3.3 & 10.0 \\
LLaVA-NeXT-Interleave-7B\cite{llava-next-interleave} & 2.7
& 1.0 & 4.6 & 1.3 & 0.0 & 1.2 & 3.5 & 0.0 & 5.0 & 0.0 \\
Mantis-Idefics2-8B\cite{mantis} & 2.0 & 1.0 & 4.6 & 5.3 & 0.0 & 7.7 & 3.0 & 0.0 & 8.4 & 0.0 \\
Mantis-siglip-8B\cite{mantis} & 1.1 & 2.1 & 3.4 & 5.3 & 5.5 & 5.1 & 0.0 & 0.0 & 4.6 & 0.0 \\
Qwen2VL-Instruct-7B\cite{Qwen2VL} & 2.8 & 3.1 & 4.6 & 4.0 & 0.0 & 2.5 & 4.5 & 40.0 & 6.7 & 10.0 \\
LLaVA-OneVision-SI-7B\cite{llava-onevision} & 6.7 & 5.2 & 10.4 & 5.3 & 11.1 & 6.4 & 7.5 & 20.0 & 9.3 & 10.0 \\
LLaVA-OneVision-SFT-7B\cite{llava-onevision} & 5.5 & 5.2 & 8.1 & 5.3 & 0.0 & 6.4 & 7.5 & 20.0 & 6.0 & 0.0 \\
LLaVA-OneVision-Chat-7B\cite{llava-onevision} & 4.8 & 1.0
& 8.1 & 4.0 & 5.5 & 3.8 & 6.0 & 20.0 & 6.7 & 20.0 \\
LLaVA-OneVision-SI-72B\cite{llava-onevision} & 13.1 & 9.4
& 11.6 & 12.0 & 0.0 & 6.4 & 16.6 & 20.0 & \colorbox{wkblue}{19.9} & 20.0 \\
LLaVA-OneVision-SFT-72B\cite{llava-onevision} & 13.7 & 13.6 & 13.9 & 12.0 & 5.5 & 6.4 & 20.2 & 20.0 & 15.2 & 40.0 \\
LLaVA-OneVision-Chat-72B\cite{llava-onevision} & 13.1 & 11.5 & 17.4 & 12.0 & 11.1 & 5.1 & 18.1 & 20.0 & 15.2 & 40.0 \\
\midrule

\multicolumn{11}{c}{Closed-source MLLMs (Text + Image, CoT with 2-shot)} \\
\midrule
Qwen-vl-plus\cite{qwenvl} & 12.6 & 8.4 & 13.9 & 8.0 & 16.6 & 10.3 & 16.6 & 20.0 & 11.8 & 20.0 \\
GPT-4V\cite{GPT-4V} & 16.4 & 12.6 & \colorbox{wkblue}{27.9} & 17.3 & 11.1
& 11.6 & 22.2 & \colorbox{wkblue}{40.0} & 11.8 & 40.0 \\
Qwen-vl-max\cite{qwenvl} & 15.8 & 13.6 & 21.1 & 18.6 & {11.1} & {10.3} & 20.2 & 20.0 & {15.6} & 30.0 \\
Gemini-1.5-Pro\cite{gemini} & \colorbox{wkblue}{23.4} & 16.8 & 26.7 & \colorbox{wkblue}{26.6} & 22.2 & 11.6 & 28.7 & 20.0 & 15.2 & \colorbox{wkred}{70.0} \\
GPT-4o\cite{GPT4o} & 22.5 & \colorbox{wkblue}{21.0} & {25.5} & \colorbox{wkred}{28.0} & \colorbox{wkblue}{27.7} & \colorbox{wkblue}{14.2} & \colorbox{wkred}{34.3} & \colorbox{wkred}{40.0} & 15.2 & {50.0} \\
Claude-3.5\cite{claude3} & \colorbox{wkred}{24.1} & \colorbox{wkred}{22.1} & \colorbox{wkred}{27.9} & {22.6} & \colorbox{wkred}{38.8} & \colorbox{wkred}{14.2} & \colorbox{wkblue}{32.8} & {20.0} & \colorbox{wkred}{22.4} & \colorbox{wkblue}{60.0} \\
\midrule

\multicolumn{11}{c}{Human Performance} \\
\midrule
Human (testmini) & 73.2 & 69.8 & 72.9 & 86.2 & 80.1 & 65.8 & 61.7 & 85.9 & 70.6 & 71.3 \\

\bottomrule
\end{tabular}
}
\caption{Comparison of model performances across various mathematical subjects on the single-step problem set. The \colorbox{wkred}{first} and \colorbox{wkblue}{second} highest accuracy of MLLMs are marked in {red} and {blue}, respectively.}
\label{tab:single-step}
\vspace{-5mm}
\end{table*}

\begin{table*}[htbp]
\centering
\resizebox{\textwidth}{!}
{%
\begin{tabular}{l|c|ccccccccc}
\toprule
Model & Overall & AG  & Algebra & MG & Combinatorics & TG & SG & Arithmetic & CG \\
\midrule

\multicolumn{10}{c}{LLMs(Text-only, CoT with 2-shot)}\\
\midrule
Deepseek-Chat\cite{deepseekchat} & 8.7 & 11.4 & 8.8 & 3.1 & 0.0 & 3.3 & 9.5 & 0.0 & 18.7 \\
\midrule

\multicolumn{10}{c}{LLMs(Text + Image Caption, CoT with 2-shot)}\\
\midrule
Deepseek-Chat\cite{deepseekchat} & 9.1 & 11.6 & 7.7 & 5.2 & 0.0 & 6.6 & 19.0 & 0.0 & 19.4 \\
\midrule

\multicolumn{10}{c}{Open-source MLLMs (Text + Image, CoT with 2-shot)}\\
\midrule
Math-LLaVA-13B\cite{math-llava} & 4.1 & 2.6 & 4.4 & 3.1 & 0.0 & 15.0 & 0.0 & 0.0 & 4.1 \\
LLaVA-v1.5-13B\cite{llava} & 2.5 & 1.7 & 2.2 & 2.0 & 0.0 & 0.0 & 4.7 & 0.0 & 6.9 \\
LLaVA-v1.5-7B\cite{llava} & 1.6 & 3.5 & 0.0 & 2.0 & 0.0 & 0.0 & 0.0 & 0.0 & 0.0 \\
VILA-13B\cite{VILA} & 2.0 & 0.0 & 2.2 & 0.0 & 0.0 & 3.3 & 4.7 & 0.0 & 9.0 \\
InternLM-XComposer2.5-VL-7B\cite{InternLMXComposer-VL} & 2.1 & 0.8 & 2.2 & 2.0 & 0.0 & 3.3 & 0.0 & 0.0 & 6.9 \\
InternVL-Chat-8B\cite{internvl} & 4.0 & 3.0 & 5.5 & 5.2 & 0.0 & 3.3 & 9.5 & 0.0 & 2.0 \\
Llama-3.2-Vision-Instruct-11B\cite{llama} & 5.0 & 1.7 & 0.0 & 2.0 & {50.0} & 0.0 & 19.0 & 0.0 & 18.7 \\
Deepseek-VL-7B\cite{deepseekvl} & 2.3 & 0.0 & 8.8 & 3.1 & 0.0 & 0.0 & 4.7 & 0.0 & 0.0 \\
LLaVA-NeXT-Interleave-7B\cite{llava-next-interleave} & 5.1 & 7.1 & 7.7 & 2.0 & 0.0 & 3.3 & 9.5 & 0.0 & 4.8 \\
Mantis-Idefics2-8B\cite{mantis} & 1.8 & 1.3 & 2.2 & 1.5 & 0.0 & 0.0 & 0.0 & 0.0 & 6.2 \\
Mantis-siglip-8B\cite{mantis} & 5.5 & 3.9 & 10.0 & 4.1 & 0.0 & 6.6 & 4.7 & 0.0 & 6.9 \\
Qwen2VL-Instruct-7B\cite{Qwen2VL} & 10.8 & 11.6 & 11.1 & 3.1 & 0.0 & 10.0 & 16.6 & 0.0 & 19.4 \\
LLaVA-OneVision-SI-7B\cite{llava-onevision} & 16.0 & 11.4 & 23.3 & 8.3 & 0.0 & 10.0 & 35.7 & 0.0 & 28.4 \\
LLaVA-OneVision-SFT-7B\cite{llava-onevision} & 17.8 & 14.9 & 22.2 & 16.6 & 0.0 & 18.3 & 14.2 & 0.0 & 27.7 \\
LLaVA-OneVision-Chat-7B\cite{llava-onevision} & 18.3 & 14.9 & 22.2 & 16.6 & 0.0 & 18.3 & 14.2 & 0.0 & 27.7 \\
LLaVA-OneVision-SI-72B\cite{llava-onevision} & 24.0 & 26.7 & 23.3 & \colorbox{wkblue}{25.5} & {50.0} & 15.0 & 28.5 & 0.0 & 19.4 \\
LLaVA-OneVision-SFT-72B\cite{llava-onevision} & 25.9 & 25.4 & 33.3 & 13.0 & {50.0} & 30.0 & 33.3 & 0.0 & 27.7 \\
LLaVA-OneVision-Chat-72B\cite{llava-onevision} & 26.2 & 27.1 & 30.0 & 16.6 & {50.0} & 18.3 & 36.9 & 0.0 & 27.7 \\
\midrule

\multicolumn{10}{c}{Closed-source MLLMs (Text + Image, CoT with 2-shot)} \\
\midrule
Qwen-vl-plus\cite{qwenvl} & 24.2 & 26.7 & 23.3 & 16.6 & {50.0} & 18.3 & 28.5 & 0.0 & 19.4 \\
GPT-4V\cite{GPT-4V} & 23.8 & 23.4 & 24.8 & 22.3 & 50.0 & 15.0 & \colorbox{wkred}{40.4} & \colorbox{wkred}{100.0} & 20.1 \\
Qwen-vl-max\cite{qwenvl} & 29.5 & 29.1 & 36.6 & 19.7 & 0.0 & 30.0 & 28.5 & 0.0 & 29.1 \\
Gemini-1.5-Pro\cite{gemini} & 31.7 & \colorbox{wkred}{34.4} & \colorbox{wkblue}{38.8} & 21.8 & \colorbox{wkblue}{50.0} & \colorbox{wkred}{36.6} & 30.9 & 0.0 & 24.3 \\
GPT-4o\cite{GPT4o} & \colorbox{wkred}{32.6} & 30.2 & \colorbox{wkred}{43.3} & 21.3 & \colorbox{wkred}{100.0} & 28.3 & \colorbox{wkblue}{38.0} & 0.0 & \colorbox{wkblue}{34.0} \\
Claude-3.5\cite{claude3} & \colorbox{wkblue}{32.3} & \colorbox{wkblue}{30.7} & 26.6 & \colorbox{wkred}{32.8} & 0.0 & \colorbox{wkblue}{35.0} & {38.0} & {0.0} & \colorbox{wkred}{38.8} \\
\midrule

\multicolumn{10}{c}{Human Performance} \\
\midrule
Human (testmini) & 78.5 & 75.2 & 76.4 & 83.2 & 79.1 & 72.4 & 66.5 & 79.0 & 71.7 \\

\bottomrule
\end{tabular}
}
\caption{Comparison of model performances across various mathematical subjects on the multi-step problem set(SAR). The \colorbox{wkred}{first} and \colorbox{wkblue}{second} highest accuracy of MLLMs are marked in {red} and {blue}, respectively.}
\label{tab:multi-step-SAR}
\vspace{-5mm}
\end{table*}

\begin{table*}[p]
\centering
\resizebox{\textwidth}{!}
{%
\begin{tabular}{l|c|ccccccccc}
\toprule
Model & Overall & AG  & Algebra & MG & Combinatorics & TG & SG & Arithmetic & CG \\
\midrule

\multicolumn{10}{c}{LLMs(Text-only, CoT with 2-shot)}\\
\midrule
Deepseek-Chat\cite{deepseekchat} & 0.0 & 0.0 & 0.0 & 0.0 & 0.0 & 0.0 & 0.0 & 0.0 & 0.0 \\
\midrule

\multicolumn{10}{c}{LLMs(Text + Image Caption, CoT with 2-shot)}\\
\midrule
Deepseek-Chat\cite{deepseekchat} & 1.0 & 0.0 & 0.0 & 0.0 & 0.0 & 0.0 & 14.2 & 0.0 & 0.0 \\
\midrule

\multicolumn{10}{c}{Open-source MLLMs (Text + Image, CoT with 2-shot)}\\
\midrule
Math-LLaVA-13B\cite{math-llava} & 0.0 & 0.0 & 0.0 & 0.0 & 0.0 & 0.0 & 0.0 & 0.0 & 0.0 \\
LLaVA-v1.5-13B\cite{llava} & 0.0 & 0.0 & 0.0 & 0.0 & 0.0 & 0.0 & 0.0 & 0.0 & 0.0 \\
LLaVA-v1.5-7B\cite{llava} & 0.0 & 0.0 & 0.0 & 0.0 & 0.0 & 0.0 & 0.0 & 0.0 & 0.0 \\
VILA-13B\cite{VILA} & 0.0 & 0.0 & 0.0 & 0.0 & 0.0 & 0.0 & 0.0 & 0.0 & 0.0 \\
InternLM-XComposer2.5-VL-7B\cite{InternLMXComposer-VL} & 0.0 & 0.0 & 0.0 & 0.0 & 0.0 & 0.0 & 0.0 & 0.0 & 0.0 \\
InternVL-Chat-8B\cite{internvl} & 0.0 & 0.0 & 0.0 & 0.0 & 0.0 & 0.0 & 0.0 & 0.0 & 0.0 \\
Llama-3.2-Vision-Instruct-11B\cite{llama} & 1.0 & 0.0 & 0.0 & 0.0 & 0.0 & 0.0 & 14.2 & 0.0 & 0.0 \\
Deepseek-VL-7B\cite{deepseekvl}& 0.0 & 0.0 & 0.0 & 0.0 & 0.0 & 0.0 & 0.0 & 0.0 & 0.0 \\
LLaVA-NeXT-Interleave-7B\cite{llava-next-interleave} & 0.0 & 0.0 & 0.0 & 0.0 & 0.0 & 0.0 & 0.0 & 0.0 & 0.0 \\
Mantis-Idefics2-8B\cite{mantis} & 0.0 & 0.0 & 0.0 & 0.0 & 0.0 & 0.0 & 0.0 & 0.0 & 0.0 \\
Mantis-siglip-8B\cite{mantis} & 0.0 & 0.0 & 0.0 & 0.0 & 0.0 & 0.0 & 0.0 & 0.0 & 0.0 \\
Qwen2VL-Instruct-7B\cite{Qwen2VL} & 0.0 & 0.0 & 0.0 & 0.0 & 0.0 & 0.0 & 0.0 & 0.0 & 0.0 \\
LLaVA-OneVision-SI-7B\cite{llava-onevision} & 1.0 & 0.0 & 0.0 & 0.0 & 0.0 & 0.0 & 14.2 & 0.0 & 0.0 \\
LLaVA-OneVision-SFT-7B\cite{llava-onevision} & 0.0 & 0.0 & 0.0 & 0.0 & 0.0 & 0.0 & 0.0 & 0.0 & 0.0 \\
LLaVA-OneVision-Chat-7B\cite{llava-onevision} & 0.0 & 0.0 & 0.0 & 0.0 & 0.0 & 0.0 & 0.0 & 0.0 & 0.0 \\
LLaVA-OneVision-SI-72B\cite{llava-onevision} & 1.0 & 0.0 & 0.0 & 0.0 & 0.0 & 0.0 & 14.2 & 0.0 & 0.0 \\
LLaVA-OneVision-SFT-72B\cite{llava-onevision} & 2.0 & 0.0 & 6.6 & 0.0 & 0.0 & 0.0 & 14.2 & 0.0 & 0.0 \\
LLaVA-OneVision-Chat-72B\cite{llava-onevision} & 2.0 & 0.0 & 6.6 & 0.0 & 0.0 & 0.0 & 14.2 & 0.0 & 0.0 \\
\midrule

\multicolumn{10}{c}{Closed-source MLLMs (Text + Image, CoT with 2-shot)} \\
\midrule
Qwen-vl-plus\cite{qwenvl} & 1.0 & 0.0 & 0.0 & 0.0 & 0.0 & 0.0 & 14.2 & 0.0 & 0.0 \\
GPT-4V\cite{GPT-4V} & 3.0 & 0.0 & 6.6 & 0.0 & 0.0 & 0.0 & 28.5 & 0.0 & 0.0 \\
Qwen-vl-max\cite{qwenvl} & 2.0 & 0.0 & 6.6 & 0.0 & 0.0 & 0.0 & 14.2 & 0.0 & 0.0 \\
Gemini-1.5-Pro\cite{gemini}& 5.0 & 0.0 & 6.6 & 12.5 & 0.0 & 0.0 & 28.5 & 0.0 & 0.0 \\
GPT-4o\cite{GPT4o} & 6.0 & 0.0 & 6.6 & 6.2 & 50.0 & 0.0 & 28.5 & 0.0 & 8.3 \\
Claude-3.5\cite{claude3} & 4.0 & 0.0 & 0.0 & 6.2 & 0.0 & 10.0 & 28.5 & 0.0 & 0.0 \\
\midrule

\multicolumn{10}{c}{Human Performance} \\
\midrule
Human (testmini) & 66.0 & 60.6 & 66.7 & 52.2 & 100.0 & 70.0 & 71.4 & 63.2 & 75.0 \\

\bottomrule
\end{tabular}
}
\caption{Comparison of model performances across various mathematical subjects on the multi-step problem set(QCR).}
\label{tab:multi-step-QCR}
\vspace{10mm}
\end{table*}

%% file: appendix/X_4_cot.tex
\section{Results of CoT, 2-shot on 3 Question Types}

We observe a distinct difference in how prompting strategies affect model performance across different question types as shown in~\Cref{tab:CoT and shot}. For multiple-choice questions, the addition of Chain-of-Thought (CoT) and 2-shot examples tend to decrease performance for most models. Specifically, out of the ten models we test, eight models perform best with the original prompt. This suggests that for multiple-choice questions, which typically require selecting an answer from given options, simpler prompts lead to better outcomes as they reduce potential confusion or overthinking induced by extra information.

In contrast, for free-form question types, CoT and 2-shot strategies bring about more significant performance improvements, particularly for multi-step problems. The complex nature of these questions benefits from the step-by-step reasoning facilitated by CoT and the illustrative examples provided by 2-shot prompting. Models like Claude-3.5-sonnet show a substantial increase in performance on single-step free-form questions when CoT and 2-shot examples are used, improving from 19.6\% to 25.6\%. Similarly, on multi-step free-form questions, models such as GPT-4o improved from 25.4\% to 32.6\% with the addition of these strategies.

These findings highlight that while CoT and few-shot prompting strategies may not universally enhance performance across all question types, they are particularly effective for free-form questions that require detailed reasoning and problem-solving steps. Incorporating these strategies can aid models in navigating the complexities of open-ended mathematical problems, thereby improving their overall reasoning capabilities.

\begin{table*}[h]
\small
    \centering
    \resizebox{\textwidth}{!}{%
\begin{tabular}{l|ccc|ccc|ccc}
\toprule
\textbf{Models} & \multicolumn{3}{c|}{\textbf{Multiple-Choice}} & \multicolumn{3}{c|}{\textbf{Single-Step Free-Form}} & \multicolumn{3}{c}{\textbf{Multi-Step Free-Form(SAR)}} \\
\cmidrule{2-10}
 & Original & +CoT & +2-shot \& CoT & Original & +CoT & +2-shot \& CoT & Original & +CoT & +2-shot \& CoT \\
\midrule
\textbf{Closed-source Models} \\
\midrule
Claude-3.5-sonnet~\cite{claude3} & 39.5 & 41.6\textcolor{red}{\scalebox{0.8}{(+2.1)}} & \textbf{44.0}\textcolor{red}{\scalebox{0.8}{(+2.4)}} & 19.6 & \textbf{25.3}\textcolor{red}{\scalebox{0.8}{(+5.7)}} & 24.1\textcolor{darkgreen}{\scalebox{0.8}{(-1.2)}} & 30.4 & \textbf{32.7}\textcolor{red}{\scalebox{0.8}{(+2.3)}} & 32.3\textcolor{darkgreen}{\scalebox{0.8}{(-0.4)}} \\
GPT-4o~\cite{GPT4o} & \textbf{41.9} & 36.2\textcolor{darkgreen}{\scalebox{0.8}{(-5.7)}} & 41.9\textcolor{red}{\scalebox{0.8}{(+5.7)}} & 24.1 & \textbf{28.1}\textcolor{red}{\scalebox{0.8}{(+4.0)}} & 22.5\textcolor{darkgreen}{\scalebox{0.8}{(-5.6)}} & 25.4 & 31.6\textcolor{red}{\scalebox{0.8}{(+6.2)}} & \textbf{32.6}\textcolor{red}{\scalebox{0.8}{(+1.0)}} \\
Gemini-1.5-pro~\cite{gemini} & \textbf{40.5} & 35.0\textcolor{darkgreen}{\scalebox{0.8}{(-5.5)}} & 35.8\textcolor{red}{\scalebox{0.8}{(+0.8)}} & 19.7 & 22.6\textcolor{red}{\scalebox{0.8}{(+2.9)}} & \textbf{23.4}\textcolor{red}{\scalebox{0.8}{(+0.8)}} & 31.1 & \textbf{33.9}\textcolor{red}{\scalebox{0.8}{(+2.8)}} & 31.7\textcolor{darkgreen}{\scalebox{0.8}{(-2.2)}} \\
Qwen-vl-max~\cite{qwenvl} & \textbf{41.5} & 41.3\textcolor{darkgreen}{\scalebox{0.8}{(-0.2)}} & 37.3\textcolor{darkgreen}{\scalebox{0.8}{(-4.0)}} & 15.5 & \textbf{17.1}\textcolor{red}{\scalebox{0.8}{(+1.6)}} & 15.8\textcolor{darkgreen}{\scalebox{0.8}{(-1.3)}} & 29.4 & 29.0\textcolor{darkgreen}{\scalebox{0.8}{(-0.4)}} & \textbf{29.5}\textcolor{red}{\scalebox{0.8}{(+0.5)}} \\
GPT-4V~\cite{GPT-4V} & 28.9 & \textbf{33.2}\textcolor{red}{\scalebox{0.8}{(+4.3)}} & 32.5\textcolor{darkgreen}{\scalebox{0.8}{(-0.7)}} & \textbf{19.8} & 18.2\textcolor{darkgreen}{\scalebox{0.8}{(-1.6)}} & 16.4\textcolor{darkgreen}{\scalebox{0.8}{(-1.8)}} & 20.0 & {23.2}\textcolor{red}{\scalebox{0.8}{(+3.2)}} & \textbf{23.8}\textcolor{red}{\scalebox{0.8}{(+0.6)}} \\
\midrule
\textbf{Open-source Models} \\
\midrule
LLaVA-OneVision-Chat-72B~\cite{llava-onevision} & \textbf{39.3} & 38.0\textcolor{darkgreen}{\scalebox{0.8}{(-1.3)}} & 38.0\textcolor{darkgreen}{\scalebox{0.8}{(0.0)}} & \textbf{15.8} & 14.2\textcolor{darkgreen}{\scalebox{0.8}{(-1.6)}} & 13.1\textcolor{darkgreen}{\scalebox{0.8}{(-1.1)}} & 25.2 & 24.0\textcolor{darkgreen}{\scalebox{0.8}{(-1.2)}} & \textbf{26.2}\textcolor{red}{\scalebox{0.8}{(+2.2)}} \\
LLaVA-OneVision-Chat-7B~\cite{llava-onevision} & \textbf{31.6} & 31.0\textcolor{darkgreen}{\scalebox{0.8}{(-0.6)}} & 31.5\textcolor{red}{\scalebox{0.8}{(+0.5)}} & \textbf{9.8} & 8.8\textcolor{darkgreen}{\scalebox{0.8}{(-1.0)}} & 4.8\textcolor{darkgreen}{\scalebox{0.8}{(-4.0)}} & 14.7 & \textbf{18.6}\textcolor{red}{\scalebox{0.8}{(+3.9)}} & 18.3\textcolor{darkgreen}{\scalebox{0.8}{(-0.3)}} \\
LLaVA-NeXT-Interleave-7B~\cite{llava-next-interleave} & \textbf{29.7} & 29.4\textcolor{darkgreen}{\scalebox{0.8}{(-0.3)}} & 26.6\textcolor{darkgreen}{\scalebox{0.8}{(-2.8)}} & \textbf{6.0} & 3.1\textcolor{darkgreen}{\scalebox{0.8}{(-2.9)}} & 2.7\textcolor{darkgreen}{\scalebox{0.8}{(-0.4)}} & \textbf{11.5} & 6.8\textcolor{darkgreen}{\scalebox{0.8}{(-4.7)}} & 5.1\textcolor{darkgreen}{\scalebox{0.8}{(-1.7)}} \\
Qwen2VL-Instruct-7B~\cite{Qwen2VL} & \textbf{33.6} & 28.2\textcolor{darkgreen}{\scalebox{0.8}{(-5.4)}} & 27.8\textcolor{darkgreen}{\scalebox{0.8}{(-0.4)}} & \textbf{6.2} & 5.2\textcolor{darkgreen}{\scalebox{0.8}{(-1.0)}} & 2.8\textcolor{darkgreen}{\scalebox{0.8}{(-2.4)}} & 5.5 & 10.7\textcolor{red}{\scalebox{0.8}{(+5.2)}} & \textbf{10.8}\textcolor{red}{\scalebox{0.8}{(+0.1)}} \\
Deepseek-VL-Chat-7B~\cite{deepseekvl} & \textbf{29.0} & 28.8\textcolor{darkgreen}{\scalebox{0.8}{(-0.2)}} & 26.0\textcolor{darkgreen}{\scalebox{0.8}{(-2.8)}} & \textbf{6.0} & 2.7\textcolor{darkgreen}{\scalebox{0.8}{(-3.3)}} & 1.5\textcolor{darkgreen}{\scalebox{0.8}{(-1.2)}} & \textbf{5.1} & 2.5\textcolor{darkgreen}{\scalebox{0.8}{(-2.6)}} & 2.3\textcolor{darkgreen}{\scalebox{0.8}{(-0.2)}} \\
\bottomrule
\end{tabular}
}
   \caption{Model Performance Evaluation across 3 Question Types and Configurations. The best performance in each category is \textbf{bolded}. Improvements over the previous configuration are indicated in \textcolor{red}{red} for increases and \textcolor{darkgreen}{green} for decreases.}
    \label{tab:CoT and shot}
    \vspace{-5mm}
\end{table*}

%% file: appendix/X_5_imagerelevance.tex
\section{Image Relevance}
To the best of our knowledge, this study is the first to analyze multi-image mathematical tasks from the perspective of image relevance. Beyond relevance, we also investigate the impact of image quantity on model performance. However, no clear patterns are observed, suggesting that the number of images may not be a critical factor in determining performance. Instead, relevance appears to play a more significant role, as the interdependence among images demands a model’s ability to perform cross-image understanding, posing a greater challenge to its reasoning capabilities.

\begin{table*}[h]
    \centering
    \resizebox{\textwidth}{!}{
    \begin{tabular}{l|ccccc|ccc|ccc}
        \toprule
        
        \textbf{Benchmarks}& Language & Multi-image & Avg Question length & Avg Analysis length & Image Relevance Annotation & Subject Number & Size & Source & MC & FF & MS \\
        
        \midrule

        MathVista & EN & \xmark & 77.1 & \xmark & \xmark & 7 & 1K/6K & Synthesized & \cmark & \xmark & \xmark \\
        
        GeoQA & EN & \xmark & 37.1 & 58.2 & \xmark & 1 & 0.7K & Internet & \cmark & \xmark & \xmark\\
        
        MATH-Vision & EN & \xmark & 42.4 & \xmark & \xmark & 16 & 3K & Synthesized & \cmark & \xmark & \xmark\\
        
        MMMU-math & EN & \xmark & 40.8 & \xmark & \xmark & 8 & 0.5K & Textbook & \cmark & \cmark & \xmark\\
        
        Mathverse-mv& EN & \cmark & 76.9 & \xmark & \xmark & 1 & 0.8K & Synthesized  & \cmark & \xmark & \xmark \\
        
        CMM-math & CN & \cmark & - & - & \xmark &7 & 0.7K & Internet & \cmark & \cmark & \xmark \\
        
        \midrule
        
        MV-MATH(Ours) & EN & \cmark & 80.2 & 150.9 & \cmark & 11 & 2K & Internet\&Annotated & \cmark & \cmark & \cmark\\
        \bottomrule
    \end{tabular}}
    \caption{Comparison with existing multimodal math benchmarks. MC: Multiple Choice, FF: Free-form, MS:Multi-Step.}
    \label{tab:comparison_among_benchmarks}
\end{table*}

%% file: appendix/X_6_datalabeling.tex
\section{Data Collection and Annotation}
\subsection{Data Collection}
We have crawled a large number of multimodal math test questions totaling around 380k multimodal math questions. To format all questions for use, we process them by OCR engine like Mathpix interface. Due to inherent errors in the OCR engine, we introduce manual checks to ensure the accuracy of parsing results and to verify whether the questions belong to multimodal math problems. Specifically, we use the three-stage strategy outlined in ~\Cref{sec:dataset} to complete the final filtering of the data, ensuring not only the high quality of the questions but also the high quality of the images. 

In the 1st stage, text-image alignment refers to ensuring the number of referenced images matches those returned by the Mathpix API. For instance, if the question text mentions 'as shown in figure 4' but the Mathpix returns 3 images, it is filtered out by our rule-based system. The 2nd stage applies rule-based filtering to detect missing text fields (e.g., missing answers or analysis) and categorize samples into multiple-choice and free-form subsets for further screening. The 3rd stage involves manual verification, where we filter out blurred images or images with text overlays. Our annotation team includes graduate students and a field expert.

\subsection{Annotation}
For the categorization of subjects and image relevance, we first obtain preliminary results through majority voting among three models: GPT-4o, Claude-3.5-Sonnet, and Qwen-VL-Max. For questions without a consensus from the voting, no annotations are initially assigned. Subsequently, each question is independently reviewed by two graduate students specializing in relevant fields, with each producing an individual annotation result. Finally, questions with conflicting annotations are adjudicated by domain experts to determine the final MV-MATH annotations.

%% file: appendix/X_7_comparison.tex
\section{Comparison with Existing Benchmarks}

We provide a detailed comparison between our MV-MATH and existing mathematical 
benchmarks in~\Cref{tab:comparison_among_benchmarks}.

%% file: appendix/X_8_casestudy.tex
\section{Case Study}
\begin{figure*}[p]
    \centering
\includegraphics[width=0.95\linewidth]{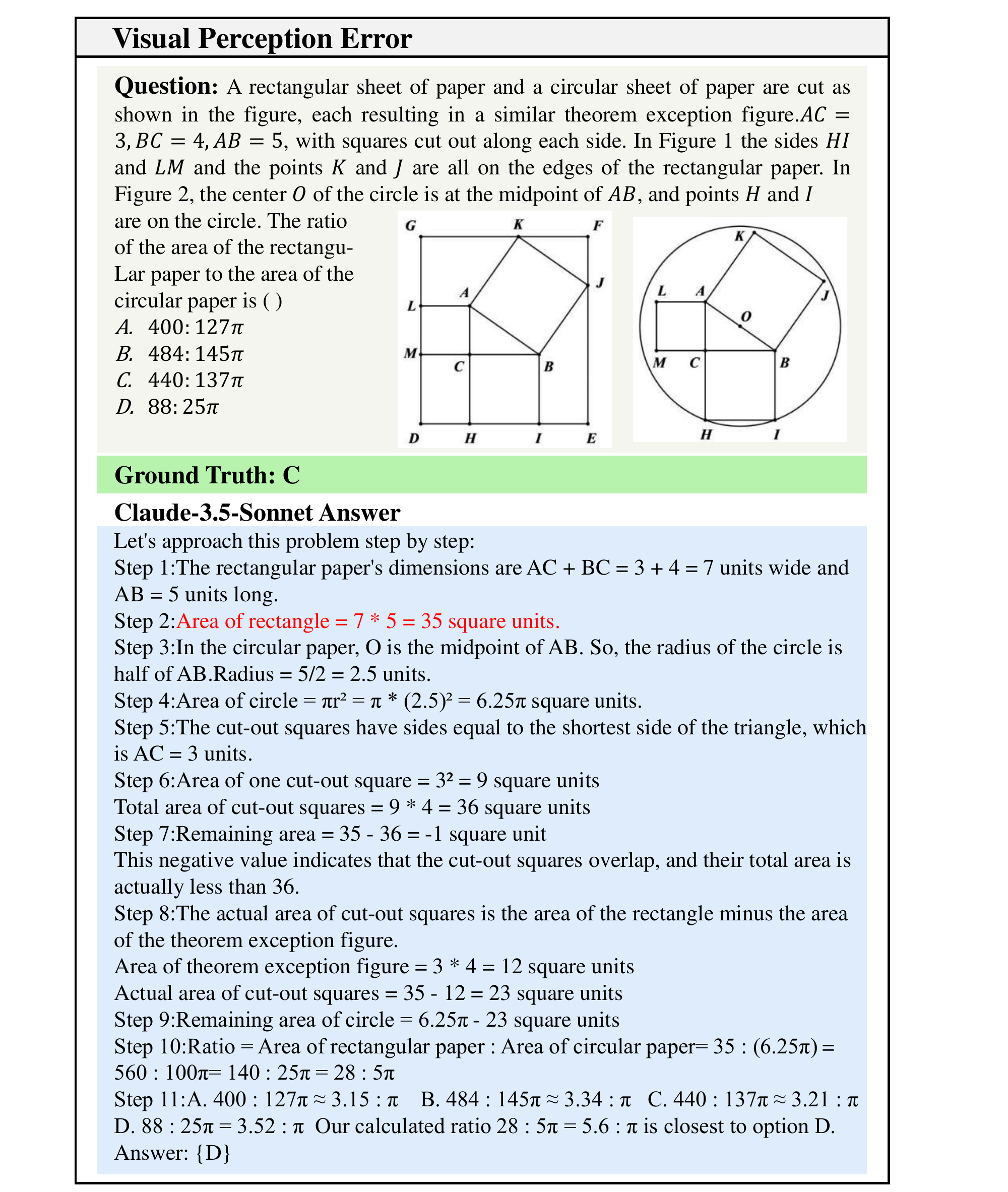}
    \caption{An example of visual perception error.}
    \label{fig: VPE1}
\end{figure*}

\begin{figure*}[p]
    \centering
\includegraphics[width=0.95\linewidth]{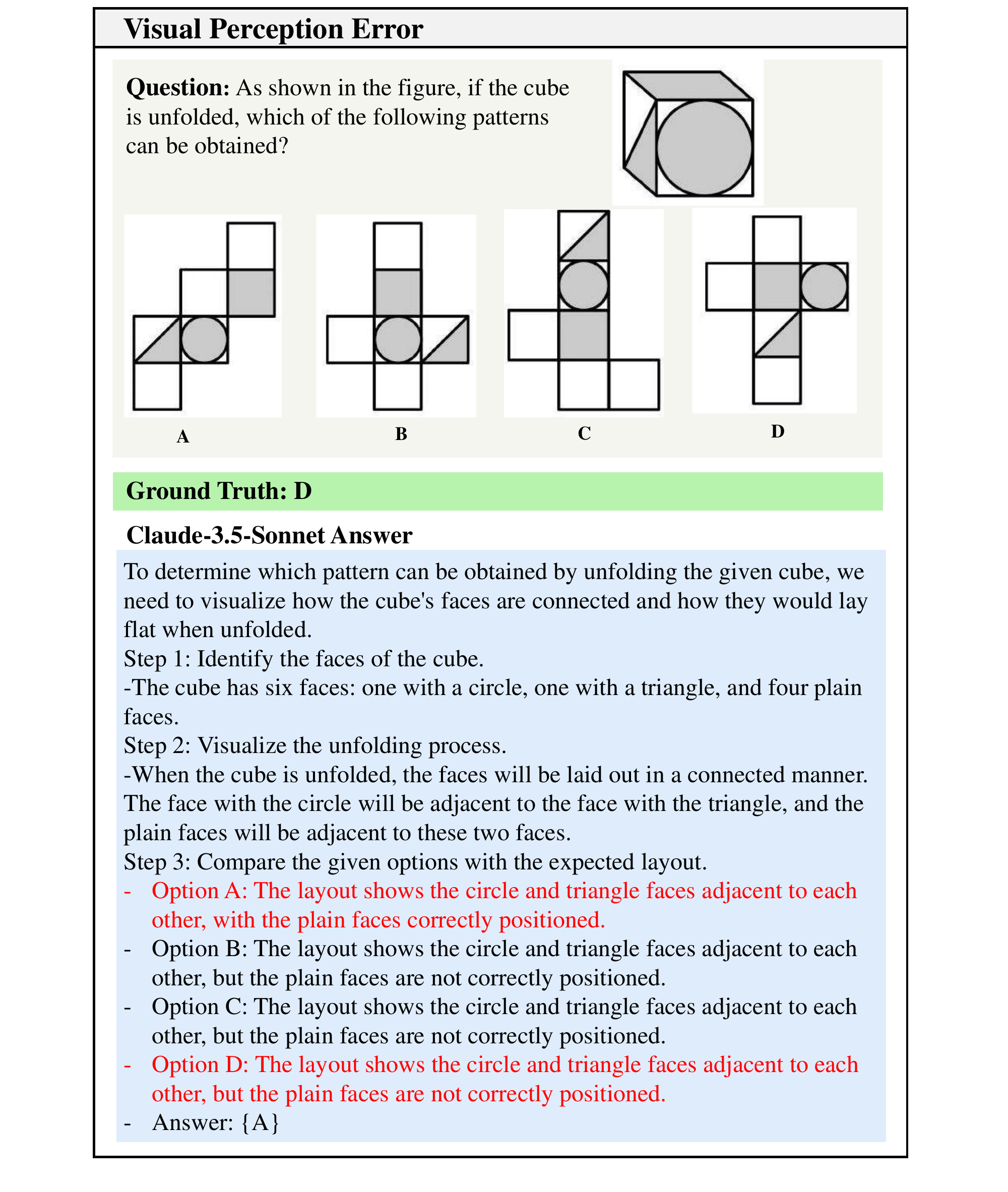}
    \caption{An example of visual perception error.}
    \label{fig: VPE2}
\end{figure*}

\begin{figure*}[!htbp]
    \centering
\includegraphics[width=0.95\linewidth]{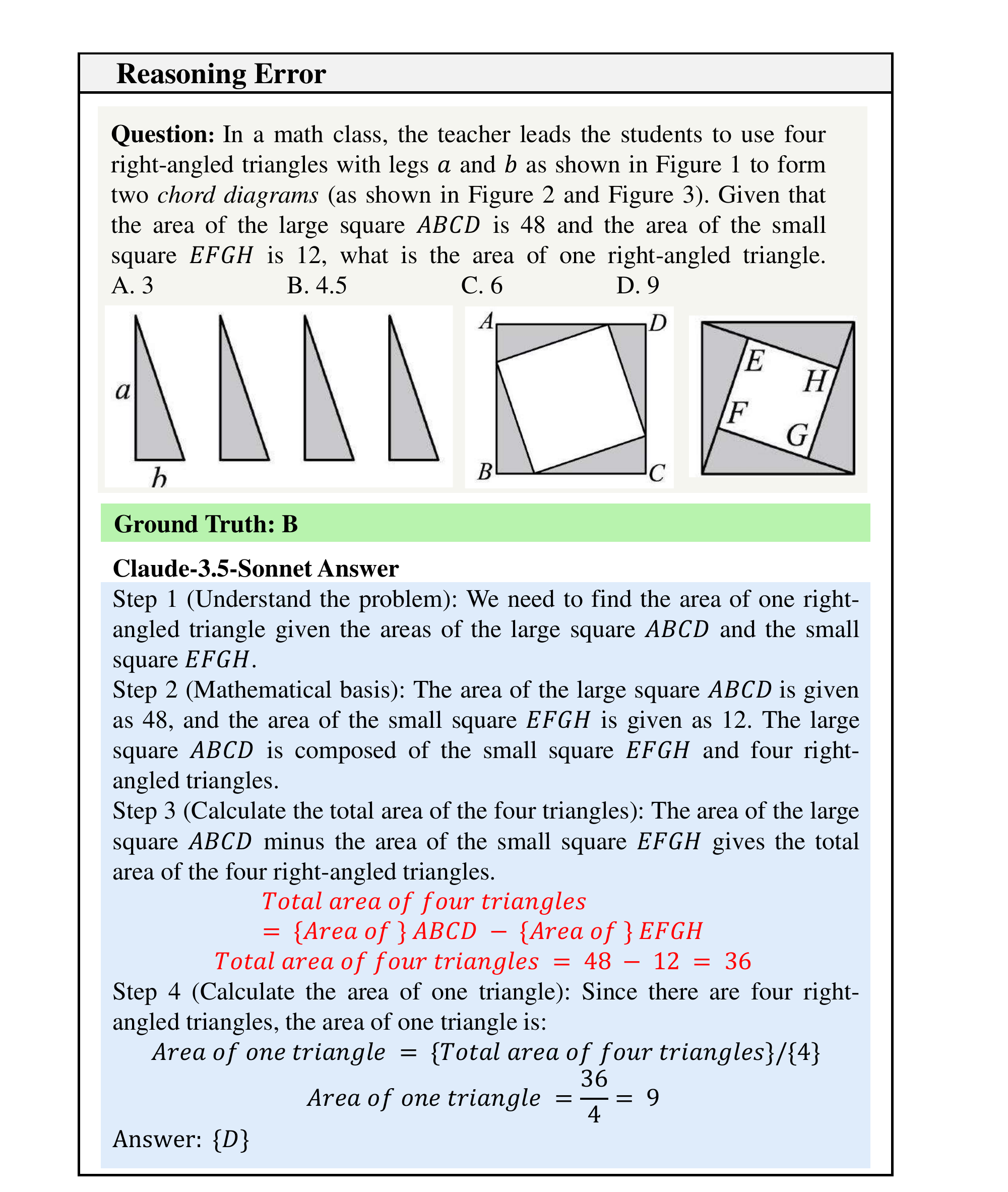}
    \caption{An example of reasoning error.}
    \label{fig: RE1}
\end{figure*}

\begin{figure*}[!htbp]
    \centering
\includegraphics[width=0.95\linewidth]{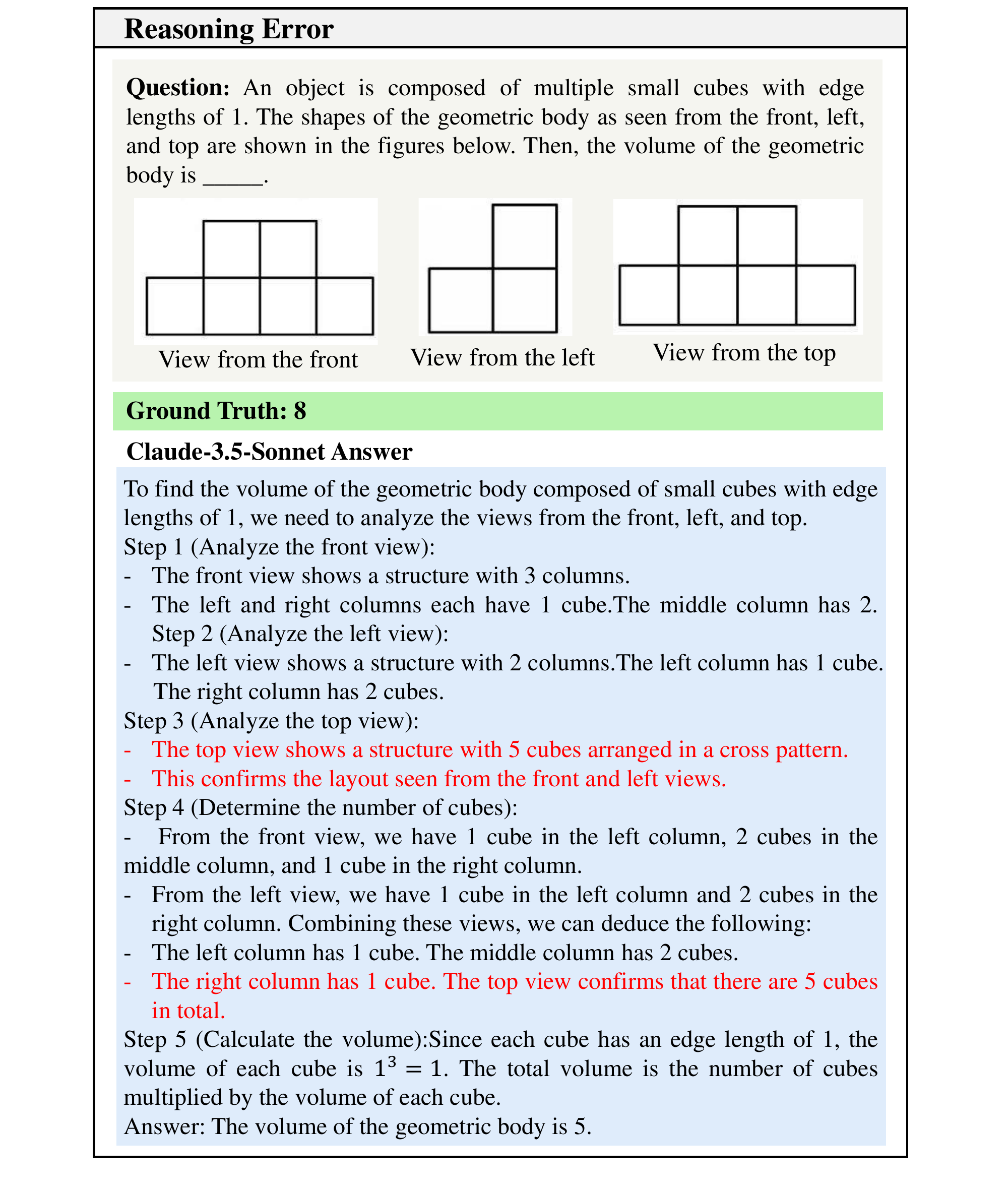}
    \caption{An example of reasoning error.}
    \label{fig: RE2}
\end{figure*}

\begin{figure*}[!htbp]
    \centering
\includegraphics[width=0.95\linewidth]{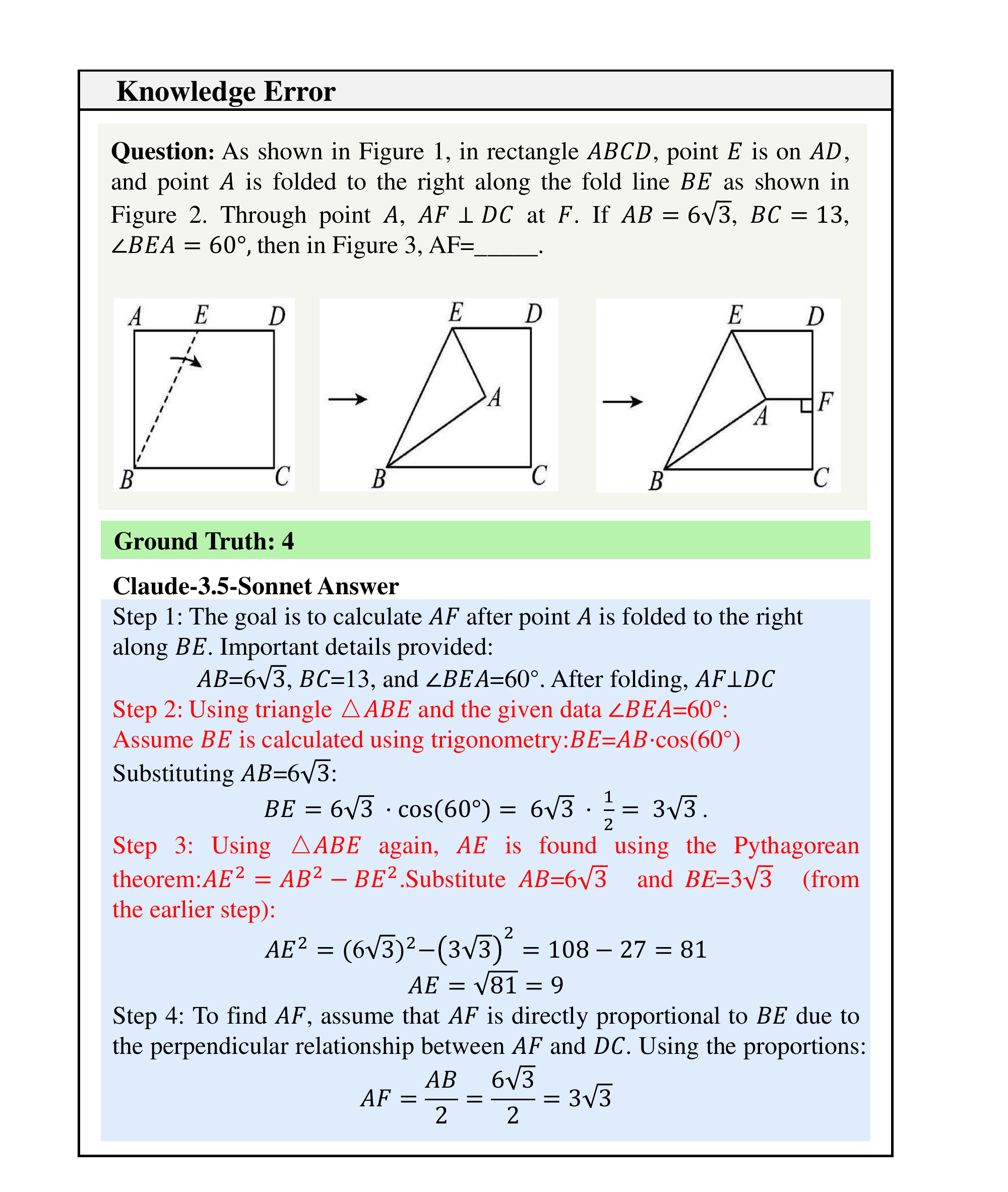}
    \caption{An example of knowledge error.}
    \label{fig: KE1}
\end{figure*}

\begin{figure*}[!htbp]
    \centering
\includegraphics[width=0.95\linewidth]{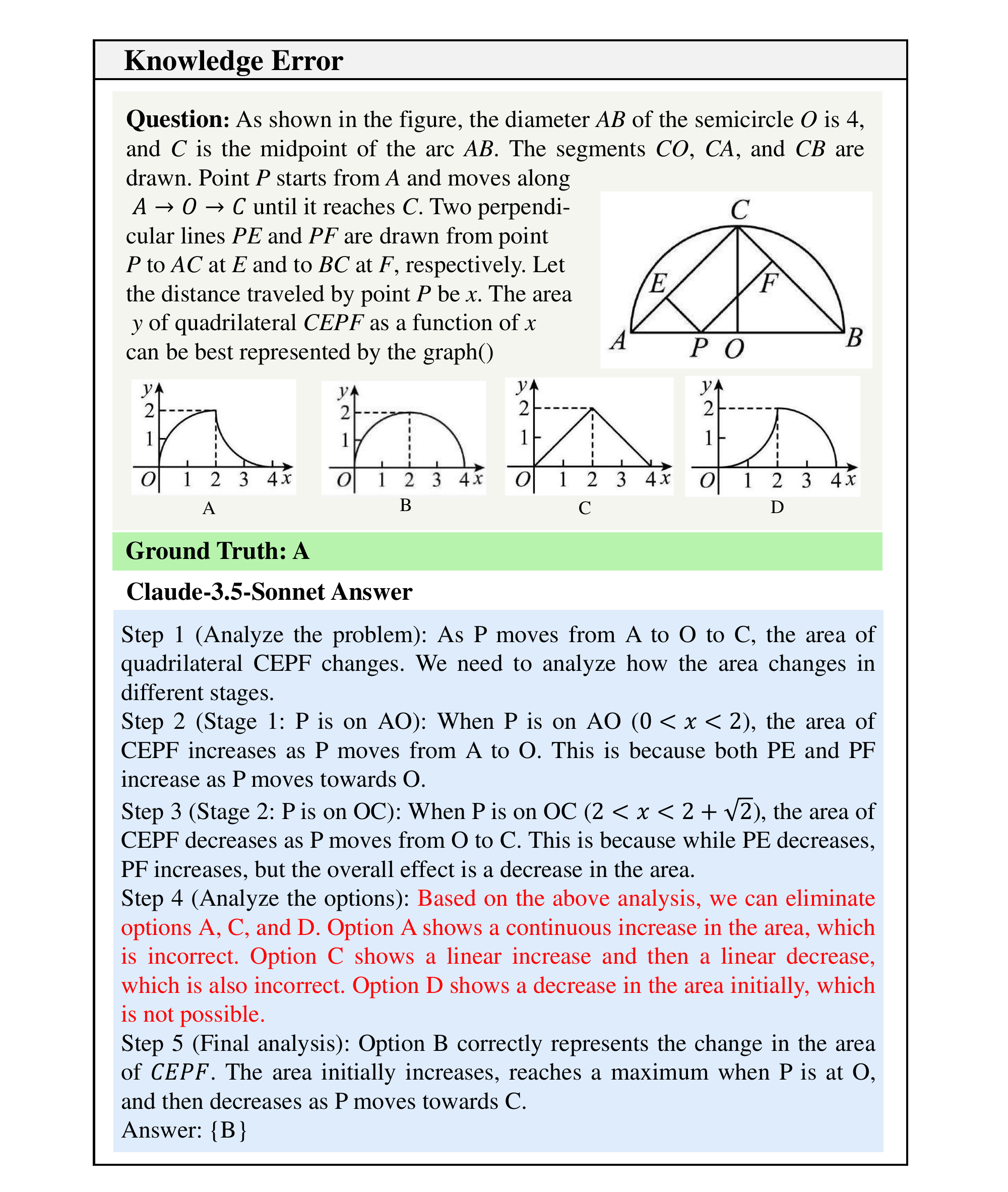}
    \caption{An example of knowledge error.}
    \label{fig: KE2}
\end{figure*}

\begin{figure*}[!htbp]
    \centering
\includegraphics[width=0.95\linewidth]{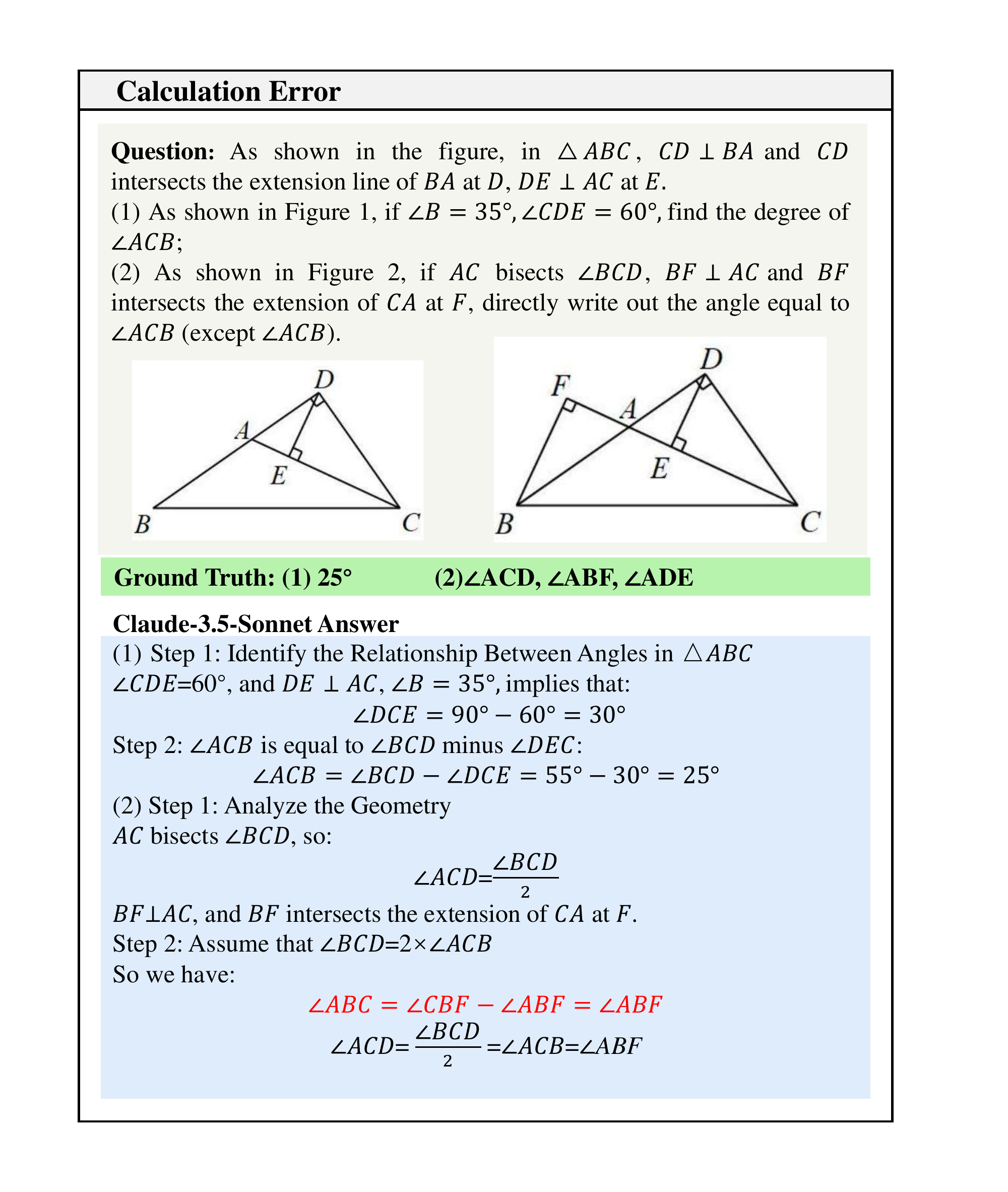}
    \caption{An example of calculation error.}
    \label{fig: CE1}
\end{figure*}

\begin{figure*}[!htbp]
    \centering
\includegraphics[width=0.95\linewidth]{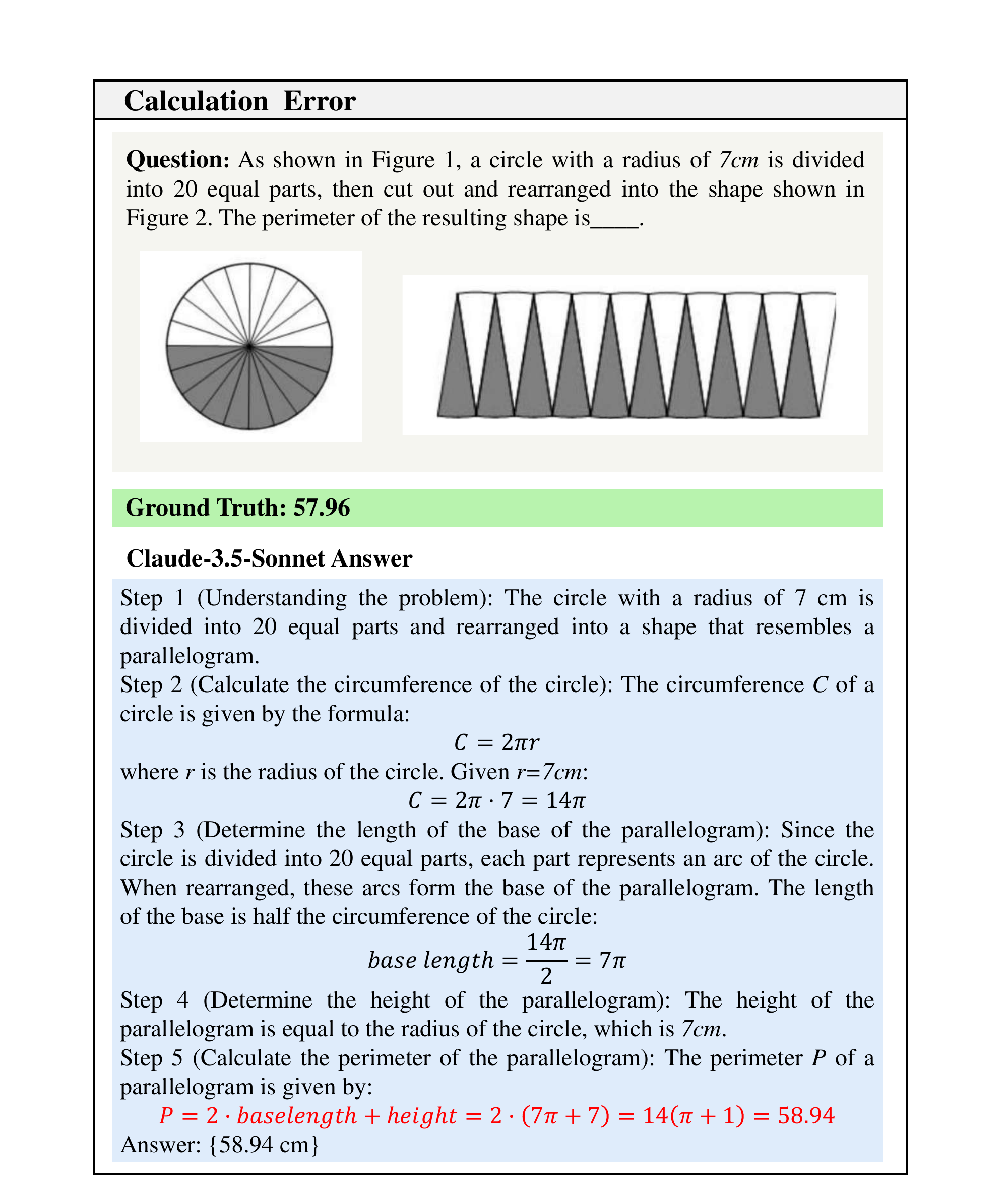}
    \caption{An example of calculation error.}
    \label{fig: CE2}
\end{figure*}

\begin{figure*}[!htbp]
    \centering
\includegraphics[width=0.95\linewidth]{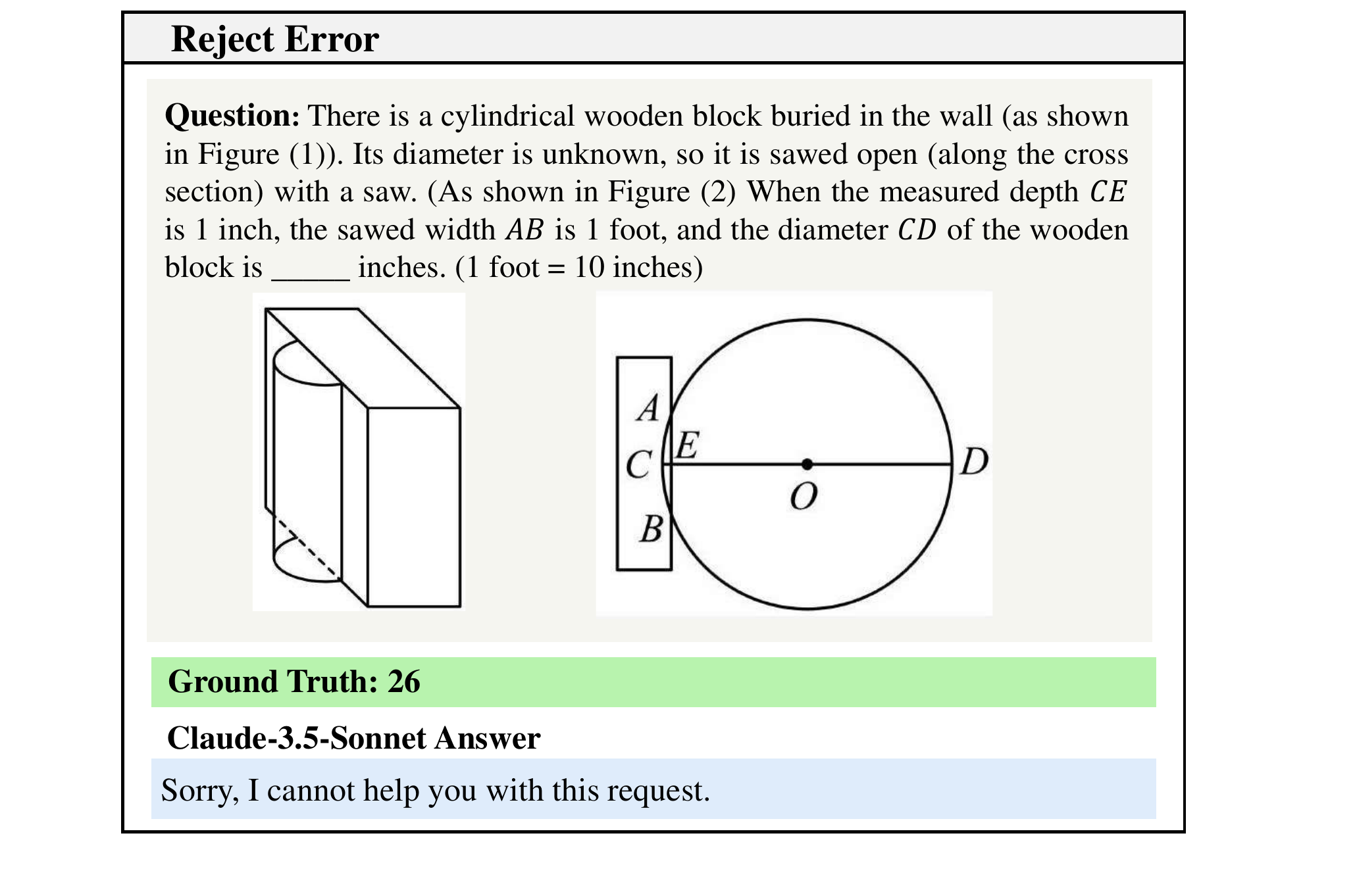}
    \vspace{-5mm}
    \caption{An example of reject error.}
    \label{fig: reject1}
\end{figure*}

\begin{figure*}[!htbp]
    \centering
\includegraphics[width=0.95\linewidth]{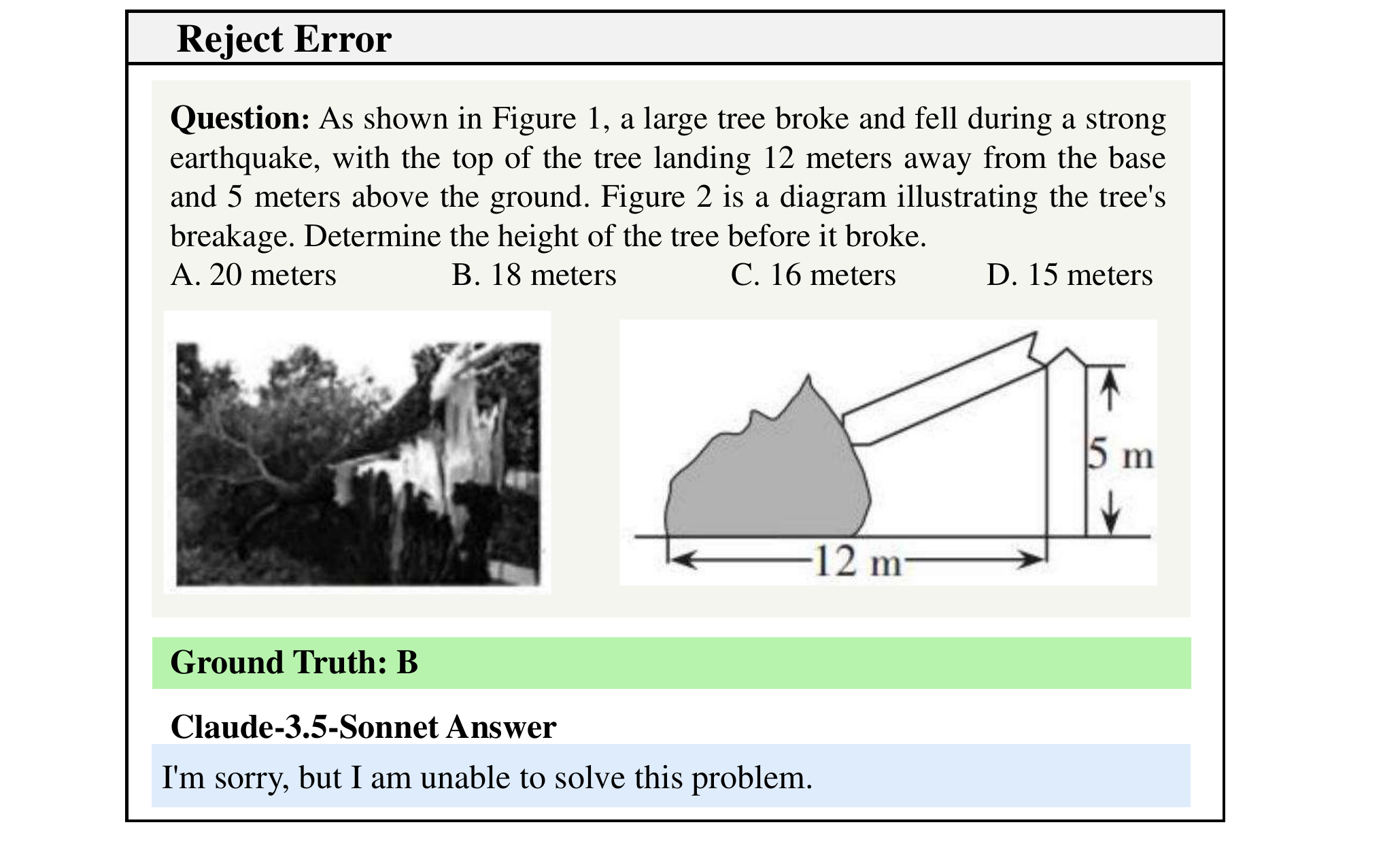}
    \caption{An example of reject error.}
    \label{fig: reject2}
\end{figure*}

In this section, we provide more detailed error examples of Claude-3.5-Sonnet. We classify Claude's errors into five categories: Visual Perception Error, Reasoning Error, Calculation Error, Knowledge Error, and Reject Error. Detailed examples can be seen in~\Cref{fig: VPE1,fig: VPE2,fig: RE1,fig: RE2,fig: KE1,fig: KE2,fig: CE1,fig: CE2,fig: reject1,fig: reject2}.

\clearpage

\clearpage

\clearpage

\clearpage

\clearpage

\clearpage

\clearpage

\clearpage

\clearpage